\pdfoutput=1
\def\arxivversion{1}
\documentclass{article}

\usepackage{preprint_style,times}

\usepackage[T1]{fontenc}
\usepackage[utf8]{inputenc}
\usepackage{microtype}
\usepackage{amsmath,amssymb,amsfonts,amsthm}
\usepackage{booktabs,array,tabularx,multirow}
\usepackage{algorithm}
\usepackage{algpseudocode}
\usepackage{graphicx,xcolor}
\usepackage{placeins}
\usepackage{hyperref}
\usepackage{url}
\hypersetup{hidelinks}
\newtheorem{proposition}{Proposition}
\usepackage{amsmath,amssymb,amsthm,xcolor,booktabs}
\newtheorem{remark}{Remark}
\newcommand{\E}{\mathbb{E}}
\newcommand{\one}{\mathbf{1}}

\newcommand{\ind}{\mathbb{I}}
\newcommand{\esw}{WG-BAcc}
\newcommand{\eswhead}{\begin{tabular}[b]{@{}c@{}}WG-\\BAcc\end{tabular}}
\newcommand{\pga}{pgAUROC}
\newcommand{\appref}[1]{Appendix~\ref{#1}}
\newcommand{\sourcecomment}[1]{}

\title{Transferring Dense Response Contrasts\\ to Distinguish Unseen Perturbations}
\author{}

\ifdefined\arxivversion
\title{\centering\normalfont\fontsize{16}{19}\selectfont
\mbox{Transferring Dense Response Contrasts to Distinguish}\\
\mbox{Unseen Perturbations}{\normalsize\textsuperscript{\ddag}}}
\let\arxivsectionfloatbarrier\FloatBarrier
\renewcommand{\FloatBarrier}{}
\AddToHook{cmd/section/before}{\arxivsectionfloatbarrier}
\author{%
\begin{minipage}[t]{\textwidth}
\centering\normalsize\bfseries
Haoran Zheng\textsuperscript{1,3}, Sheng Yin\textsuperscript{2}, Yunlong Bie\textsuperscript{2}, Qiuyi Li\textsuperscript{2},
Yuan He\textsuperscript{2}, \\[2pt]Yong Zhang\textsuperscript{3}, 
Yan Gao\textsuperscript{2}, Ziqing Deng\textsuperscript{3}, Tao Qin\textsuperscript{2}, Yuxiang Li\textsuperscript{3}, \\[2pt] Haiguang Liu\textsuperscript{2,\dag}, Qijin Yin\textsuperscript{2,\dag}, Shaoshuai Li\textsuperscript{3,\dag}\\[4pt]
\normalfont\small
\textsuperscript{1}University of the Chinese Academy of Sciences, Beijing, China\\
\textsuperscript{2}Zhongguancun Academy, Beijing, China\\
\textsuperscript{3}BGI Research, Beijing, China
\end{minipage}%
}
\makeatletter
\def\@maketitle{\vbox{\hsize\textwidth
{\centering\LARGE\sc\@title\par}
\vskip 0.18in
{\centering\@author\par}
\vskip 0.28in minus 0.04in}}
\makeatother
\hypersetup{
  pdftitle={Transferring Dense Response Contrasts to Distinguish Unseen Perturbations},
  pdfauthor={Haoran Zheng, Sheng Yin, Yunlong Bie, Qiuyi Li, Yuan He, Yong Zhang, Yan Gao, Ziqing Deng, Tao Qin, Yuxiang Li, Haiguang Liu, Qijin Yin, Shaoshuai Li},
  pdfsubject={Preprint},
  pdfkeywords={cellular perturbation prediction, Perturb-seq, response transfer}
}

\fi

\begin{document}
\maketitle
\ifdefined\arxivversion
\begingroup
\renewcommand{\thefootnote}{\dag}
\footnotetext[0]{Corresponding authors: liuhaiguang@bza.edu.cn, yinqijin@bza.edu.cn, lishaoshuai@genomics.cn}
\renewcommand{\thefootnote}{\ddag}
\footnotetext[0]{RELAY is the perturbation-prediction module (Cyto.Perturbation) of the Cyto model family for virtual cells.}
\endgroup
\lhead{\raisebox{-0.15em}{\includegraphics[height=0.95em]{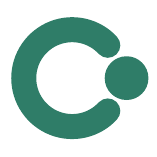}}\hspace{0.35em}Cyto.Perturbation}
\rhead{}
\fi

% !TEX root = ../main.tex
\begin{abstract}
Predicting how unseen perturbations alter gene expression is essential for prioritizing biological experiments. Through a systematic analysis of single-cell perturbation prediction, we show that post-perturbation gene expression changes combine two distinct components, namely each target gene's typical response across perturbations and the perturbation-specific effect on the same target gene. Capturing the first component alone via a training-only gene prior that assigns each target gene its majority training label already attains a differential-expression (DE) Macro-F1 of 0.8814, surpassing every language-model and expression-based predictor we evaluate. LLM mechanistic reasoning stays near chance within each target gene. Prompt ablations show that retrieval LLMs instead rely on the outcomes of related training perturbations rather than biological text, yet retrieving a few sparse binary labels leaves the dense continuous responses measured across all training pairs unused and often overrides reliable priors. Meanwhile, our error decomposition shows that direct predictive and generative models misestimate gene means or contrast amplitudes, while learned similarity-based transfer still falls well below the gene prior on DE. We introduce RELAY, a lightweight method without a language model that anchors on training gene means, transfers dense signed and magnitude contrasts over a biological knowledge kernel, and maps them to calibrated DE and direction-of-change (DIR) decisions. Across the four PerturbQA cell lines, RELAY outperforms every evaluated baseline, reaching DE and DIR Macro-F1 of 0.8941 and 0.8722 and improving balanced accuracy within each target gene by 4.65 and 4.41 percentage points over the strongest baseline. RELAY also outperforms competing predictors on an unseen cell line and best discriminates unseen chemical compounds acting on the same target gene while nearly matching the gene prior globally. Inference code and evaluation scripts are available at \url{https://github.com/Cyto-System/Cyto.Perturbation}.
\end{abstract}
% !TEX root = ../main.tex
\section{Introduction}\label{sec:intro}

\begin{figure}[t]
\centering
\includegraphics[height=2.87in]{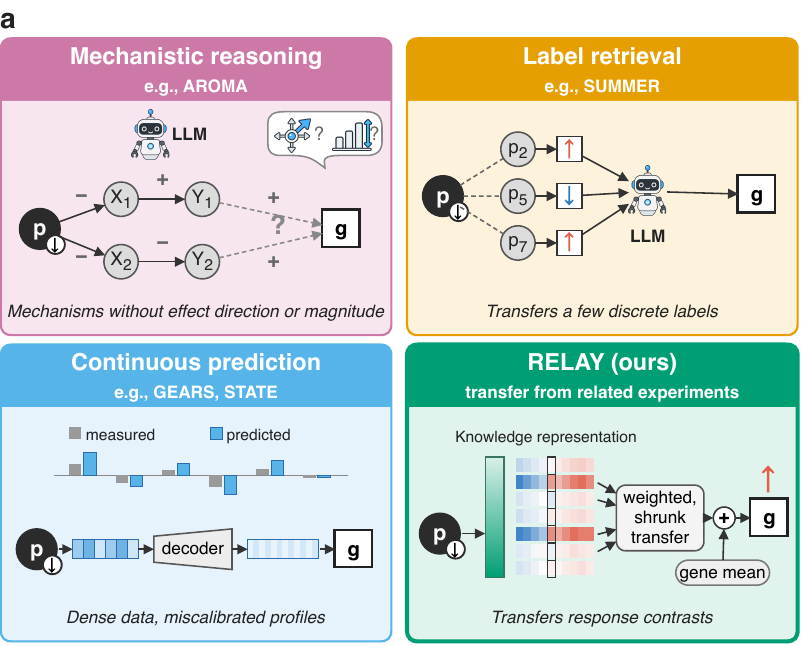}\hfill
\includegraphics[height=2.87in]{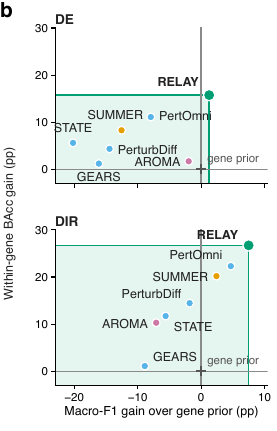}
\caption{\textbf{Prediction strategies, global accuracy, and within-gene discrimination.} \textbf{a,} Information used to predict the response of target gene $g$ to unseen perturbation $p$ across three existing paradigms and RELAY ($p_i$ denotes related training perturbations and $X_i, Y_i$ denote pathway intermediates). \textbf{b,} Global Macro-F1 gain over the gene prior (origin) versus within-gene \esw{} gain over chance in percentage points (pp) on PerturbQA DE (top) and DIR (bottom). Shading marks the region where RELAY outperforms on both metrics.}
\label{fig:teaser}
\end{figure}

Predicting how a genetic perturbation changes gene expression is essential for prioritizing functional experiments. Perturb-seq screens measure these changes by pairing pooled genetic perturbations with single-cell RNA sequencing \citep{dixit2016perturbseq,adamson2016perturbseq,replogle2022perturbseq}. Building on these screens, discrete perturbation prediction asks two fundamental questions for each perturbation and target gene, namely whether the gene is differentially expressed (DE) and, for DE-positive pairs, whether it is up- or downregulated (DIR), as benchmarked in PerturbQA \citep{wu2025perturbqa}. Solving both tasks requires capturing how each target gene typically responds across perturbations while distinguishing among perturbations that act on the same gene. Gene-level tendencies are strong enough that a training-only \emph{gene prior}, which predicts each gene's majority training label, already reaches a DE Macro-F1 of 0.8814. Because this prior assigns every perturbation of a gene the same label, however, it cannot distinguish among perturbations and scores at chance ($0.5$) within each gene \citep{yuan2026core}. We therefore evaluate two complementary objectives, namely global accuracy, measured by Macro-F1, and within-gene discrimination, measured by support-weighted within-gene balanced accuracy (\esw{}, introduced in Section~\ref{sec:evaluation}, Figure~\ref{fig:teaser}b).

None of the evaluated baselines meets both objectives on both tasks (Figure~\ref{fig:teaser}). Although recent large language model (LLM) methods report gains over expression-based predictors on PerturbQA \citep{wu2025perturbqa,wang2026aroma,wei2026vcworld}, every LLM and expression-based baseline we evaluate falls below the gene prior in DE Macro-F1. In particular, AROMA---which reasons over biological mechanisms---comes closest to the prior globally, yet remains near chance within genes because curated pathways and signed regulatory networks rarely specify effect magnitudes or unambiguous multi-step directions. By contrast, SUMMER---which retrieves related training pairs alongside gene descriptions---achieves the strongest within-gene discrimination among LLM baselines, though at a cost of 12.5 DE Macro-F1 points.

Because SUMMER supplies both biological descriptions and retrieved training labels in the same prompt, its gain could stem from either source. Deleting individual prompt fields resolves this ambiguity: removing biological descriptions changes \esw{} by at most 1.33 points, whereas removing the retrieved training outcomes reduces DIR to chance. Within-gene discrimination thus comes from the empirical outcomes of related perturbations rather than from free-form reasoning over biological text. Relying on a few retrieved discrete labels, however, leaves most measured training responses unused and often overrides reliable gene priors on DE. Continuous expression predictors instead train on measured responses across all genes. However, our error decomposition shows that direct predictive and generative models---including GEARS, STATE, and the functional-diffusion model PerturbDiff---misestimate gene-level offsets or contrast amplitudes, leaving their held-out profiles no more accurate than the training gene means; even contrastive similarity-based transfer (PertOmni), which lowers profile error and achieves the strongest baseline within-gene discrimination, still falls 7.9 DE Macro-F1 points below the gene prior.

These findings indicate that the two objectives draw on different sources of signal. Gene-level response tendencies, which the training gene means already capture, support global accuracy, whereas perturbation-specific measured responses support within-gene discrimination. Directly addressing each diagnosed failure mode, we introduce \textbf{RELAY}, a lightweight response-transfer method that requires no language model. To avoid the gene-offset errors of end-to-end profile decoders, RELAY retains the training gene means to support global accuracy. To exploit the dense training responses missed by discrete label retrieval without cancelling opposite-signed effects, it decomposes responses around those means into centered signed and magnitude contrasts. Finally, to control the amplitude overshoot of deep predictors, it uses regularized kernel ridge regression to transfer those contrasts from related training perturbations while shrinking noisy estimates toward zero. To quantify perturbation relatedness, it represents each perturbed gene by a \emph{knowledge representation}---a vector extracted from multi-source knowledge graphs that encodes curated annotations and interaction partners---and weights training perturbations by the similarity of their knowledge representations. Task-specific readouts then combine each gene's baseline tendency with the transferred contrast. For DIR, adding the gene mean shifts predictions toward each gene's typical direction while preserving the within-gene ranking of the transferred contrast. For DE, a learned readout evaluates the predicted effect on a gene against both the perturbation's overall response pattern and the gene's typical responsiveness.

Our main contributions are summarized below.
\begin{enumerate}
  \item We show why existing within-gene rankings conflate score resolution with binary decision quality, introduce support-weighted within-gene balanced accuracy (\esw{}), and diagnose why three existing strategies fall short in global accuracy or within-gene discrimination. Specifically, mechanistic reasoning lacks response-level evidence and falls back on the gene prior, label retrieval derives its within-gene signal from sparse training labels that often override reliable gene priors, and continuous expression predictors misestimate both gene-level offsets and contrast amplitudes.
  \item Motivated by these diagnostics, we introduce RELAY, a lightweight method that anchors on training gene means, transfers dense signed and magnitude contrasts through a multi-source knowledge kernel with regularized shrinkage, and pairs them with task-specific readouts.
  \item We show that, on average across the four PerturbQA cell lines, RELAY achieves the lowest held-out continuous profile error and outperforms every evaluated baseline on both global and within-gene metrics, improving \esw{} over the strongest baseline by 4.65 (DE) and 4.41 (DIR) percentage points while extending effectively to a held-out cell line and unseen chemical compounds.
\end{enumerate}

% !TEX root = ../main.tex
\section{Problem Setup and Related Work}\label{sec:background}

\subsection{Tasks and Available Information}\label{sec:task}\label{sec:information}
We study two tasks for a perturbation $p$ and a measured target gene $g \in \mathcal G$ in a given cell line. Differential expression (DE) asks whether $p$ significantly alters the expression of $g$, and direction of change (DIR) asks whether a DE-positive gene $g$ is up- or downregulated. Our main benchmark, PerturbQA, poses both tasks for CRISPRi knockdowns across four cell lines, namely K562, HepG2, Jurkat, and RPE1 \citep{wu2025perturbqa}. To test generalization beyond genetic knockdowns, we also evaluate on GeneTAK, a benchmark curated by VCWorld from the Tahoe-100M single-cell atlas that profiles small-molecule compounds across five cell lines \citep{wei2026vcworld,zhang2025tahoe}. Within each cell line, perturbations are split into disjoint training and test sets $\mathcal P_{\mathrm{tr}}$ and $\mathcal P_{\mathrm{te}}$, ensuring that every test query involves an unseen perturbation in that cell line. Beyond holding out perturbations within each cell line, we evaluate cross-cell-line transfer on PerturbQA by predicting responses in RPE1 from the training data of K562, HepG2, and Jurkat, with RPE1 contributing only unperturbed controls.

For each task $T\in\{\mathrm{DE},\mathrm{DIR}\}$, the benchmarks provide binary labels only on selected query sets $\mathcal Q_{\mathrm{tr}}^T$ and $\mathcal Q_{\mathrm{te}}^T$, leaving most perturbation--gene pairs unlabeled. The underlying Perturb-seq screens, however, record dense continuous measurements beyond these sparse labels. For every training perturbation $p \in \mathcal P_{\mathrm{tr}}$ and target gene $g \in \mathcal G$, the measured response $r(p,g)$ captures the difference in mean $\log(1+\mathrm{CP10K})$ expression between perturbed cells and matched controls. Across $\mathcal P_{\mathrm{tr}}\times\mathcal G$, these measurements form a dense training \emph{response matrix} alongside the sparse \emph{labeled support set} $\mathcal Q_{\mathrm{tr}}^T$ (\appref{sec:supervision} details dataset construction and split sizes). On PerturbQA, the source datasets also supply published effect estimates $F(p,g)$ with standard errors $\mathrm{SE}(p,g)$ that quantify per-pair measurement precision \citep{nadig2025trade,nadig2025statistics}. To relate unseen test perturbations to $\mathcal P_{\mathrm{tr}}$, predictors can draw on curated knowledge graphs that catalog functional annotations and molecular interactions for perturbed genes \citep{wu2025perturbqa} and compound targets \citep{zhang2025tahoe,chandak2023primekg}.

% !TEX root = ../main.tex
\subsection{Related Work}\label{sec:related}\label{sec:transfer-related}\label{app:related}
\paragraph{Benchmarking and baseline analyses.}
Because many target genes exhibit strong baseline response tendencies across perturbations, simple baselines can achieve deceptively high aggregate scores. On continuous expression profiles, benchmark analyses show that linear or mean predictors often match or exceed deep generative models \citep{ahlmann2025linear,wong2025controls} and zero-shot single-cell foundation models \citep{kedzierska2025zeroshot}. On discrete tasks, PerturbQA therefore evaluates perturbation rankings within each target gene \citep{wu2025perturbqa}, and CORE shows that marginal gene-level label frequencies and yes-biased LLM prompts can inflate aggregate accuracy even when within-gene discrimination remains near chance \citep{yuan2026core}. Prior benchmark studies largely document when simple baselines match complex models or how label frequencies skew aggregate scores. Going beyond these observations, we introduce a support-weighted within-gene decision metric that separates binary decision quality from score resolution and systematically diagnose why existing predictors fail across three modeling paradigms. Specifically, we quantify the coverage and sign consistency of directed regulatory paths in curated knowledge graphs, isolate the source of within-gene signal in retrieval-augmented LLMs through prompt-field deletions and identity--outcome exchanges, and decompose continuous profile error into gene-level offset, directional alignment, and amplitude components.

\paragraph{Language-model reasoning and retrieval.}
Existing approaches to perturbation question answering fall into two broad families, i.e., language-model methods tailored to discrete DE and DIR queries, and continuous profile predictors whose outputs must be converted into binary decisions. Within the first family, \emph{mechanistic reasoning} models such as AROMA and SynthPert train on synthetic reasoning traces to answer perturbation queries by reasoning over biological mechanisms \citep{wang2026aroma,phillips2026synthpert}. By contrast, \emph{label retrieval} methods draw on historical experimental outcomes from the labeled support set. Specifically, SUMMER and VCWorld retrieve related support-set outcomes alongside biological descriptions to prompt an LLM \citep{wu2025perturbqa,wei2026vcworld}, whereas CORE organizes same-gene positive and negative support outcomes into contrastive evidence for voting or LLM reasoning \citep{yuan2026core}. While mechanistic traces rely on qualitative pathway context without measured response magnitudes, in-context retrieval is constrained by sparse benchmark labels and finite prompt budgets.

\paragraph{Continuous profile prediction and response transfer.}
The second family estimates continuous expression profiles across target genes and thresholds the predicted profiles for discrete DE and DIR decisions \citep{wu2025perturbqa,wei2026vcworld}. Deep \emph{continuous prediction} models decode learned representations into post-perturbation expression profiles. Among them, GEARS combines coexpression and Gene Ontology graphs \citep{roohani2024gears}, STATE pairs perturbation representations with pretrained cell-state embeddings \citep{adduri2026state}, PerturbDiff applies functional diffusion in a reproducing kernel Hilbert space to generate perturbed single-cell distributions conditioned on perturbation embeddings \citep{yuan2026perturbdiff}, and TxPert adds multi-graph perturbation embeddings to an encoded basal state before decoding \citep{wenkel2026txpert}. Instead of decoding latent embeddings from scratch, \emph{similarity-based profile prediction} reuses measured training responses directly, either by averaging a perturbation's knowledge-graph neighbors (optionally refined by an LLM) \citep{fawkes2026kg} or by contrastively aligning multimodal text embeddings of genes, pathways, compounds, and cell lines with measured perturbation responses and averaging $k$-nearest neighbors in the aligned space, as in PertOmni \citep{long2026pertomni}. However, whether decoding full profiles end to end or averaging raw neighbor profiles directly, existing continuous predictors either misestimate baseline gene tendencies and contrast amplitudes, risk cancelling opposite-signed regulatory effects when inferring differential expression, or lack calibrated task-specific readouts for discrete decisions. Bridging these perspectives, RELAY anchors on training gene means, uses a multi-source knowledge kernel to transfer dense signed and magnitude contrasts with regularized shrinkage, and maps the transferred contrasts to calibrated DE and DIR decisions.

\paragraph{Biological knowledge graphs and regulatory networks.}
Both knowledge-grounded reasoning and response transfer rely on curated databases to relate perturbations and genes. Unsigned context resources group entities by shared biological function or physical interaction. Specifically, Gene Ontology (GO) \citep{ashburner2000go,aleksander2023go}, Reactome \citep{milacic2024reactome}, and CORUM \citep{tsitsiridis2023corum} annotate biological processes, pathways, and protein complexes, while STRING \citep{szklarczyk2023string,szklarczyk2025string}, BioPlex \citep{huttlin2021bioplex}, and PrimeKG \citep{chandak2023primekg} catalog protein--protein associations and compound--target links. By contrast, signed regulatory resources---including TRRUST \citep{han2018trrust}, CollecTRI \citep{mullerdott2023collectri}, OmniPath \citep{turei2026omnipath}, and directional edges in STRING \citep{szklarczyk2025string}---specify whether an interaction activates or inhibits its target, providing the directional constraints used in network sign-consistency analyses \citep{thiele2015sign}. Because unsigned resources cover most perturbed genes whereas signed regulatory links remain sparse and omit effect sizes, we examine in Section~\ref{sec:mechanisms} whether curated networks provide sufficient directional evidence for direct inference or serve better as similarity measures for response transfer.

% !TEX root = ../main.tex
\section{Diagnosing Existing Predictors}\label{sec:diagnostics}\label{sec:diagnostics-v2}

We begin our analysis by showing why existing evaluation metrics do not cleanly separate baseline gene tendencies and score resolution from within-gene decision quality, and we introduce a support-weighted within-gene metric that addresses both issues (Section~\ref{sec:evaluation}). Using this dual evaluation framework, we then diagnose why existing predictors fall short in global accuracy or within-gene discrimination (Figure~\ref{fig:teaser}b) by examining representative baselines across three prevailing modeling paradigms, namely mechanistic reasoning (AROMA, Section~\ref{sec:mechanisms}), label retrieval (SUMMER, Section~\ref{sec:retrieval-transfer}), and continuous prediction (GEARS, STATE, PerturbDiff, and the similarity-based transfer model PertOmni, Section~\ref{sec:response-learning}). Our analysis progresses from models that rely on curated biological relations without inference-time training outcomes, to models that retrieve sparse discrete training labels in context, and finally to predictors trained on dense continuous expression profiles.

\subsection{Global and Within-Gene Evaluation}\label{sec:evaluation}\label{sec:marginal}\label{sec:eval-v2}\label{sec:ablation}\label{sec:marginal-variation}\label{app:metrics}
Existing studies evaluate discrete perturbation predictions either by global class-wise F1 across all queries \citep{wang2026aroma} or by per-gene AUROC (\pga{}) across perturbations of the same target gene \citep{wu2025perturbqa}. Neither perspective alone captures both global accuracy and within-gene discrimination. The training-only \emph{gene prior}, which assigns each target gene its majority training label, achieves high global Macro-F1 yet gives every perturbation of a gene the same prediction, whereas within-gene metrics ignore global miscalibration across genes. We therefore pair global Macro-F1, computed over all test queries in each cell line and task, with within-gene evaluation on the mixed-label target genes $\mathcal G_{\mathrm{mix}}$ whose test queries contain both classes (shared by all evaluated models).

Within $\mathcal G_{\mathrm{mix}}$, ranking quality is not equivalent to final binary decision quality. For each gene, AUROC evaluates pairwise rankings across positive and negative queries and counts tied score pairs as $0.5$ \citep{fawcett2006roc}. When a predictor outputs binary decisions, a gene's AUROC equals its balanced accuracy $\mathrm{BAcc}_g$ \citep{brodersen2010balanced}, whereas finer-grained scores can raise \pga{} by resolving ties among queries that receive the same binary decision. Because continuous predictors output real-valued scores while LLMs return coarse categorical answers or small vote counts, comparing methods by \pga{} alone conflates decision quality with score resolution. We therefore retain \pga{} as a secondary ranking metric and evaluate within-gene decisions directly through $\mathrm{BAcc}_g$.

Averaging $\mathrm{BAcc}_g$ equally across genes (pgBAcc), however, is noisy because more than half of the mixed-label DE genes have only a single test query in one class. For these genes, flipping one prediction changes $\mathrm{BAcc}_g$ by $0.5$. Under Bernoulli trials that are independent within and across the two classes, the variance of a gene's estimated balanced accuracy is one quarter of the sum of its sensitivity and specificity variances, bounded by
\begin{equation}
 \operatorname{Var}(\widehat{\mathrm{BAcc}}_g)
 \leq \frac1{16}\left(\frac1{n_g^+}+\frac1{n_g^-}\right),
\end{equation}
where $n_g^+$ and $n_g^-$ count the positive and negative test queries for gene $g$. Weighting each gene inversely to this variance bound defines the \emph{support-weighted within-gene balanced accuracy} (\esw{}) as
\begin{equation}\label{eq:esw-v2}
 w_g=\frac{n_g^+n_g^-}{n_g^++n_g^-},\qquad
 \mathrm{WG\text{-}BAcc}=\frac{\sum_{g\in\mathcal G_{\mathrm{mix}}}w_g\,\mathrm{BAcc}_g}{\sum_{g\in\mathcal G_{\mathrm{mix}}}w_g}.
\end{equation}
Here $w_g$ equals half the harmonic mean of the two class counts and corresponds to minimax inverse-variance weighting that minimizes the worst-case variance bound in Equation~(1) across $\mathcal{G}_{\mathrm{mix}}$ under working independence across genes (\appref{app:proofs}), thereby downweighting noisy low-support genes. Specifically, the 1,601 of 2,806 mixed-label DE genes with $w_g\le1$ (most having a single query in one class) account for 76\% of the aggregate variance bound under equal weighting (pgBAcc) but only 34\% under \esw{}. Macro-F1 and \esw{} thus serve as our primary metrics for global accuracy and within-gene discrimination, reported alongside \pga{} and pgBAcc to separate ranking resolution, unweighted within-gene decision quality, and support-weighted decision reliability.

\begin{figure}[t!]
\centering
\includegraphics[width=0.88\linewidth]{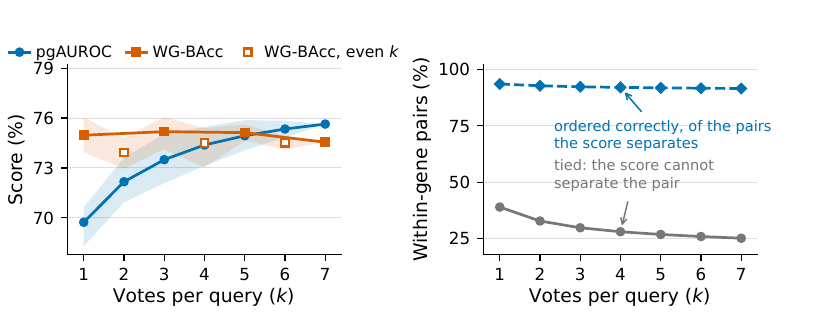}
\caption{\textbf{Ranking resolution and decision evaluation.} The left panel shows \pga{} and \esw{} across vote counts $k$, with shaded bands spanning all $\binom{7}{k}$ vote subsets. Open even-$k$ points assign tied $0.5$ scores to up (the majority class in K562 DIR, 2,530 of 3,119 queries) and are not directly comparable to odd-$k$ decisions. The right panel shows the share of tied within-gene pairs and the ordering accuracy among untied pairs.}
\label{fig:vote-resolution}
\end{figure}

\phantomsection\label{app:vote-resolution}
To verify empirically how score resolution affects \pga{} versus \esw{}, we varied SUMMER's vote count on the 3,119 K562 DIR test queries by sampling up to seven independent answers per query under fixed prompts and decoding settings (Figure~\ref{fig:vote-resolution}). This controlled analysis spans output granularities ranging from AROMA's single ternary output ($k=1$) and SUMMER's default three-vote average ($k=3$) toward finer multi-vote averages. For each $k$-vote subset, a query's score is the mean of its valid votes (mapping scores $\ge 0.5$ to up), averaged over all $\binom{7}{k}$ subsets. As $k$ increases from $1$ to $7$, \pga{} rises from 69.73\% to 75.64\%, whereas across odd vote counts \esw{} remains nearly flat between 74.54\% and 75.18\% (with full-population accuracy between 86.85\% and 87.14\%). Because a $k$-vote mean takes at most $k+1$ distinct values, adding votes raises \pga{} by reducing tied within-gene positive--negative pairs from 38.96\% to 25.14\%, even as ordering accuracy on untied pairs slips slightly from 93.48\% to 91.48\%.

\subsection{From Biological Context to Response Evidence}\label{sec:mechanisms}\label{sec:mechanism-gap}
Mechanistic reasoning traces how a perturbation alters biological processes and how these changes affect the target gene. AROMA \citep{wang2026aroma} is a representative method in this family that uses post-training to optimize LLM reasoning for perturbation prediction. Although AROMA achieves strong global performance on PerturbQA, our analysis shows that it agrees with the gene prior on 96.77\% of DE decisions. On the remaining 3.23\% where it departs from the prior, it introduces more errors than it corrects, lowering overall accuracy by 0.70 percentage points. This reliance on the prior is consistent with a mismatch between supervision and inference. AROMA is fine-tuned on PerturbReason traces generated by a teacher model that is shown the observed outcome before producing its explanation, and is then reinforced for correct predictions \citep{wang2026aroma}. At inference, however, the student model receives only the query inputs without the outcome. Because neither training stage controls for baseline gene tendencies, defaulting to each gene's majority training label offers the easiest path to high reward.

Beyond the training objective, mechanistic reasoning draws its biological facts from curated knowledge graphs, and therefore depends on whether these graphs contain the directional and quantitative evidence needed to distinguish perturbations. To test this premise, we examine both unsigned context resources (GO and CORUM) and signed regulatory databases (STRING, TRRUST, OmniPath, and CollecTRI, with resource versions and filtering protocols in \appref{app:mechanistic}). Curated relations describe shared biological context far more often than directed regulatory paths (Table~\ref{tab:coverage}). On DIR queries, GO biological-process annotations \citep{ashburner2000go} cover both genes in 94.96\% of pairs and CORUM complex membership covers both in 42.86\%, whereas signed relations at both endpoints within individual resources range from only 1.35\% (TRRUST) to 11.79\% (STRING). Enumerating simple directed paths of at most three signed edges \citep{szklarczyk2025string,han2018trrust,turei2026omnipath} connects the perturbed and target genes in just 17.99\% of DIR queries and 10.59\% of DE queries. Requiring the terminal edge to directly regulate gene expression reduces coverage further to 15.87\% and 7.73\%. Moreover, curated relations do not encode effect magnitudes, even though DE depends on whether an expression change crosses a statistical detection threshold.

\begin{table}[H]
\vspace{-4pt}
\centering\small
\caption{\textbf{Coverage of query-specific directional evidence.} All rows combine STRING, TRRUST, OmniPath signaling, and CollecTRI (\appref{app:mechanistic}). Percentages are averaged with equal weight across cell lines, and counts are pooled across cell lines (116,041 DE and 16,208 DIR queries in total).}
\label{tab:coverage}
\setlength{\tabcolsep}{5pt}
\begin{tabularx}{\linewidth}{@{}Xrrr@{}}
\toprule
Evidence condition & DE (\%) & DIR (\%) & Pairs (DE / DIR)\\
\midrule
Local directed relations at both endpoints & 26.27 & 35.91 & 30,721 / 5,895\\
Directed connection, at most three edges & 10.59 & 17.99 & 12,315 / 2,957\\
Local relations, including gene expression regulation & 18.57 & 31.45 & 21,738 / 5,183\\
Short connection ending in expression regulation & 7.73 & 15.87 & 9,002 / 2,604\\
\bottomrule
\end{tabularx}
\vspace{-6pt}
\end{table}

Even on the queries connected by signed paths, static network topology alone cannot reliably determine the direction of change. Transcriptional outcomes depend on cell-line-specific cofactors and relative pathway strengths that are absent from static interaction databases. When we propagate edge signs along short directed paths ending in transcriptional regulation (via $-\prod_{e\in\mathcal P}\varsigma_e$ under CRISPRi knockdown, following the protocol in \appref{app:sign}), competing paths between the same perturbed and target genes yield conflicting terminal signs on two thirds of the covered DIR queries. Whether we require all paths to agree (\emph{consistency}) or take a majority vote across distinct node sequences (\emph{majority}), path-sign propagation resolves fewer than 10\% of all DIR queries and performs near chance in balanced accuracy (Table~\ref{tab:path-sign-aggregation}).

\begin{table}[H]
\vspace{-4pt}
\centering\small
\caption{\textbf{Coverage and accuracy of path-sign aggregation on DIR.} Coverage and BAcc are computed on queries resolved by path signs before fallback. Mean BAcc weights cell lines equally, whereas pooled BAcc combines these queries across cell lines.}
\label{tab:path-sign-aggregation}
\setlength{\tabcolsep}{6pt}
\begin{tabularx}{\linewidth}{@{}Xrr@{}}
\toprule
Measure & Consistency & Majority\\
\midrule
Queries resolved by path signs & 660 & 1,618\\
Coverage of all 16,208 DIR queries (\%) & 4.07 & 9.98\\
\addlinespace[2pt]
Mean BAcc across cell lines (\%) & 48.89 & 53.69\\
Pooled BAcc (\%) & 45.12 & 49.90\\
\bottomrule
\end{tabularx}
\vspace{-6pt}
\end{table}

\begin{figure}[H]
\vspace{-8pt}
\centering
\includegraphics[width=0.69\linewidth]{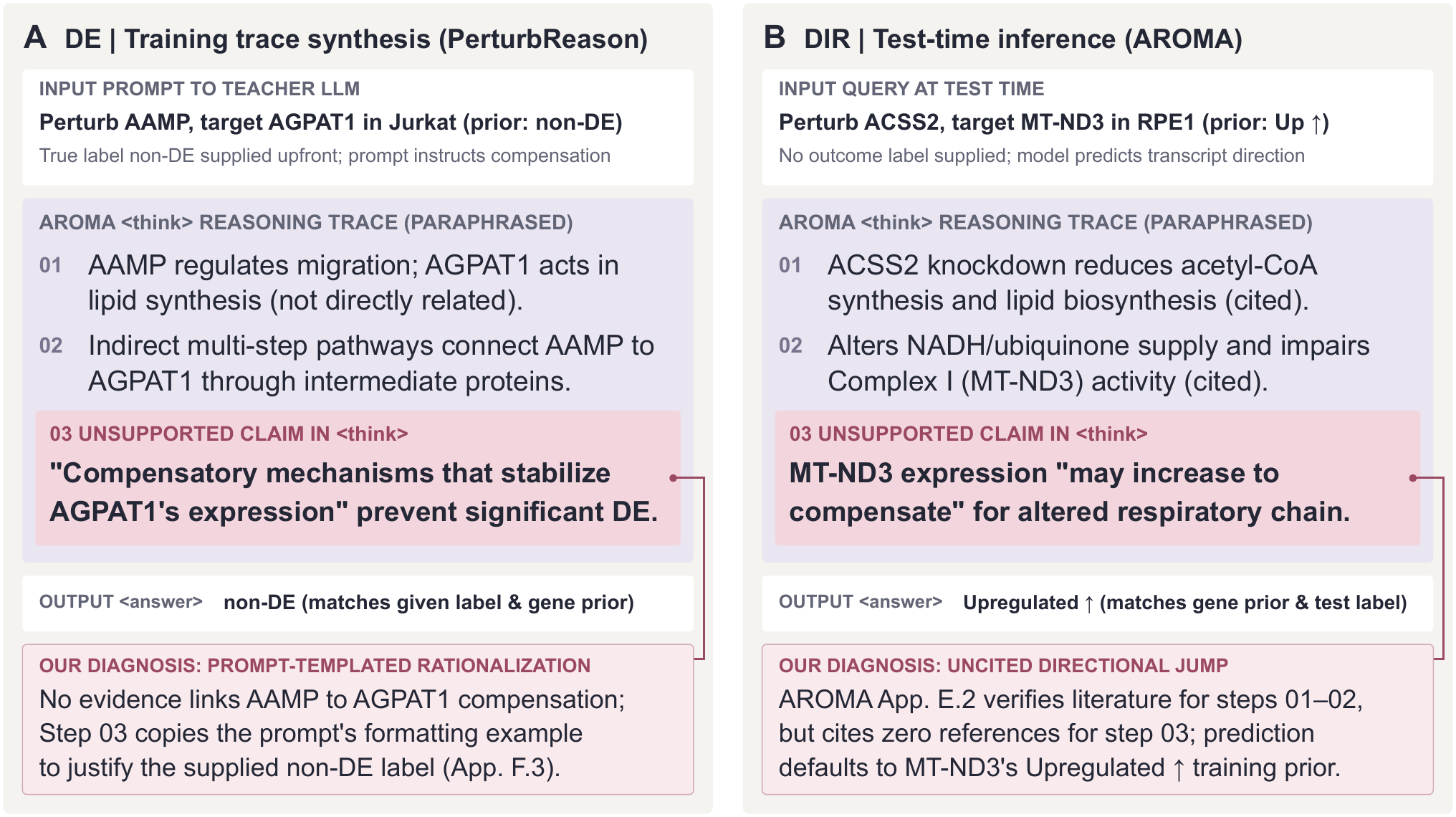}
\vspace{-6pt}
\caption{\textbf{Unsupported compensation steps in PerturbReason synthesis and AROMA inference.} Each panel separates the input prompt (top white card), the generated \texttt{<think>} trace (middle lavender card, produced by the PerturbReason synthesis teacher in A and by AROMA at test time in B, with the unsupported step in rose), the final \texttt{<answer>} tag, and our diagnosis (bottom callout). \textbf{(A)} In PerturbReason synthesis, the teacher receives the true non-DE label upfront and follows the prompt's compensation template \citep[App.~F.3]{wang2026aroma}. \textbf{(B)} At test time, AROMA cites metabolic links but makes an uncited compensation jump matching MT-ND3's training prior \citep[App.~E.2]{wang2026aroma}.}
\label{fig:decisive-response-claims}
\vspace{-8pt}
\end{figure}

\phantomsection\label{app:cases}%
Because curated networks lack threshold-relevant magnitudes and unambiguous directional paths, AROMA's explanations bridge the gap to the final answer with unverified compensation claims (Figure~\ref{fig:decisive-response-claims}). First, in the Jurkat training-synthesis example pairing AAMP knockdown with target gene AGPAT1 (Figure~\ref{fig:decisive-response-claims}A, \citealp[App.~F.3 and App.~G]{wang2026aroma}), AAMP regulates cell migration whereas AGPAT1 catalyzes phosphatidic acid synthesis. As AROMA's retrieved summaries note, the two genes have no direct regulatory link in the supplied descriptions, and neither those summaries nor the source-traced passages in Appendix~G of \citet{wang2026aroma} establish a compensatory link from AAMP knockdown to AGPAT1. Instead, the claim that ``compensatory mechanisms that stabilize AGPAT1's expression'' prevent DE originates from the synthesis prompt, which supplies the true non-DE label upfront, instructs the teacher to invoke ``buffering, compensation, or feedback effects'' when intermediate evidence is missing, and includes ``compensatory mechanisms in jurkat could stabilize AGPAT1's expression'' in its format example \citep[App.~F.3]{wang2026aroma}. Second, in the RPE1 test-time example pairing ACSS2 knockdown with Complex~I subunit MT-ND3 (Figure~\ref{fig:decisive-response-claims}B, \citealp[App.~E.2]{wang2026aroma}), AROMA's sentence-level verification cites studies for the first two steps---showing that ACSS2 loss reduces acetyl-CoA synthesis, alters NADH/ubiquinone supply, and impairs Complex~I activity---but cites zero references for the final claim that MT-ND3 ``may increase to compensate.''

Crucially, in the PerturbQA training split, the gene prior for AGPAT1 in Jurkat is already non-DE (with mean absolute training response $\nu=0.043$), and the gene prior for MT-ND3 in RPE1 is already upregulated (with mean signed training response $\mu=+0.033$). Consistent with AROMA's 96.77\% agreement with the gene prior, both outputs match the target gene's training prior while filling in the missing mechanistic link with an unsupported compensation hypothesis. Distinguishing perturbations within a gene thus requires empirical perturbation responses rather than qualitative biological context alone.

\subsection{Outcome Retrieval and Its Coverage Limit}\label{sec:retrieval-transfer}\label{sec:reference-signal}
Label retrieval answers a query using the observed outcomes of related training pairs. SUMMER, a representative method, supplies an LLM with the biological descriptions and observed outcomes of 11--12 retrieved training pairs per query \citep{wu2025perturbqa}. This strategy achieves the strongest within-gene discrimination among LLM baselines, yet falls 12.5 DE Macro-F1 points below the gene prior.

\paragraph{Reference-field deletion.}
To isolate the source of this within-gene signal, we delete individual reference fields from the prompt while holding the query, the retrieved reference identities, and their ordering fixed (Table~\ref{tab:summer-deletion}, with decoding settings in \appref{app:inputs}). Every condition with references retains the query and reference gene names, the cell line, and the structured reasoning questions. Specifically, \emph{Full reference} includes both biological descriptions and observed outcomes, \emph{No descriptions} removes the description prose, \emph{No outcomes} removes the observed-outcome lines, and \emph{Query only} removes all retrieved instances and descriptions. Because the retrieval index is held fixed, comparing \emph{No descriptions} against \emph{Full reference} isolates the contribution of biological prose at the reasoning stage rather than its role in selecting neighbors. Removing descriptions changes \esw{} by at most 1.33 points, whereas removing outcomes reduces DIR to the query-only level ($-19.36$ points) and eliminates more than half of the DE gain ($-4.71$ points in Table~\ref{tab:summer-deletion}).

\begin{table}[H]
\vspace{-4pt}
\centering
\caption{\textbf{SUMMER\textquoteright s within-gene signal comes from retrieved outcomes.} Deletion runs, means across four cell lines. $\Delta$\esw{} is relative to Full with 95\% paired perturbation-cluster intervals, in percentage points. Macro-F1 uses full populations. DE deletion runs use single-sample decoding (\appref{app:inputs}), and query-only answers non-DE for every query.}
\label{tab:ablation}\label{tab:summer-deletion}
\small
\setlength{\tabcolsep}{3.1pt}
\begin{tabular}{@{}llrcr@{}}
\toprule
Task & Input & \esw{} & $\Delta$\esw{} [95\% CI] & Macro-F1\\
\midrule
DE & Full reference & 0.5849 & --- & 0.7895\\
 & No descriptions & 0.5860 & $+0.11\;[-0.65,\,1.06]$ & 0.7865\\
 & No outcomes & 0.5378 & $-4.71\;[-5.71,\,-3.87]$ & 0.4590\\
 & Query only & 0.5000 & $-8.49\;[-9.49,\,-7.76]$ & 0.4626\\
\midrule
DIR & Full reference & 0.7019 & --- & 0.8215\\
 & No descriptions & 0.6886 & $-1.33\;[-2.41,\,-0.27]$ & 0.8215\\
 & No outcomes & 0.5083 & $-19.36\;[-22.08,\,-17.82]$ & 0.4218\\
 & Query only & 0.5061 & $-19.59\;[-22.45,\,-17.95]$ & 0.4442\\
\bottomrule
\end{tabular}
\vspace{-6pt}
\end{table}

\phantomsection\label{app:binding}%
\paragraph{Identity--outcome exchange.}
A remaining question is whether SUMMER uses the specific pairing between each retrieved reference and its outcome, or merely responds to the aggregate proportion of positive and negative labels in the prompt. To test this, we conduct an identity--outcome exchange experiment on the 4,121 Jurkat DIR test queries. Within this set, 1,332 queries contain retrieved references with opposite labels that can be swapped while matching graph-neighbor status and repetition counts, whereas the remaining 2,789 queries stay fixed across conditions so that all 4,121 queries are evaluated together. On the swappable pairs, we compare four conditions, namely leaving references \emph{unchanged}, exchanging only their \emph{outcomes}, exchanging only their gene \emph{identities}, and exchanging \emph{both} simultaneously.

Swapping only outcomes or only identities breaks the correspondence between each retrieved perturbation and its empirical result while preserving the exact label counts in the prompt. By contrast, swapping both restores every original reference--outcome pair and changes only its presentation order. Across all 4,121 queries, the unchanged baseline reaches a \esw{} of 0.6703 under single-sample decoding (\appref{app:inputs}). Exchanging only outcomes or only identities drops \esw{} to 0.6221 ($-4.83$ percentage points, 95\% CI $[-6.82, -3.03]$) and 0.6348 ($-3.56$ points), whereas exchanging both nearly matches the baseline at 0.6649 ($-0.55$ points, 95\% CI $[-1.56, 0.61]$). Biological knowledge thus determines which related training perturbations to consult and how to weight them, while their empirical outcomes carry the within-gene signal.

Sparse label availability, however, severely restricts how much evidence in-context retrieval can access. Across the four PerturbQA cell lines, the median test gene has only 12--30 labeled training perturbations for DE and 4--7 for DIR, even though the underlying Perturb-seq screens provide measured continuous responses across all 1,086--1,596 training perturbations. Relying on a small retrieved set also causes SUMMER to override reliable gene priors. On the 42,153 DE test queries whose target gene has at least 20 identical training labels, the gene prior is 99.57\% accurate, compared with only 88.85\% for SUMMER.

Simply packing more retrieved examples into the prompt does not resolve this bottleneck. SUMMER caps retrieval at 15 examples per query to fit context limits \citep{wu2025perturbqa}. While deterministic voting continues to improve when expanded to the full labeled support set, LLM-based reasoning must restrict retrieval to a moderate, prompt-feasible budget \citep{yuan2026core}. Because LLMs have finite context and attention capacity, adding weakly related examples yields diminishing returns and increases distraction \citep{liu2024lost,shi2023large}. A single prompt therefore cannot aggregate evidence across the full training set, and in SUMMER this limited sample improves within-gene discrimination at the expense of global DE accuracy.

\subsection{Continuous Response Predictors}\label{sec:response-learning}
Continuous response profiles can in principle support both global accuracy and within-gene discrimination. On the PerturbQA test queries, the sign of the ground-truth measured response $r(p,g)$ reproduces every DIR label, $|r(p,g)|$ ranks DE queries within genes at a \pga{} of 0.987, and applying a single global threshold to $|r(p,g)|$ reaches a DE Macro-F1 of 0.946 and \esw{} of 0.924. In practice, however, direct predictive and generative baselines---GEARS \citep{roohani2024gears}, STATE \citep{adduri2026state}, and PerturbDiff \citep{yuan2026perturbdiff}---all fall 1.8--20.2 Macro-F1 points below the gene prior across tasks. Evaluating their predicted profiles across all 1,264 test perturbations (Table~\ref{tab:profile-comparison}, protocol in \appref{app:continuous-diagnosis}) reveals that their held-out profiles achieve worse MSE than simply predicting the training gene means. In these direct models, estimation errors outweigh the recovered perturbation-specific signal, whereas response-transfer methods---PertOmni \citep{long2026pertomni} and RELAY---convert measured training responses into a net MSE reduction over the gene means, with RELAY achieving the lowest MSE and MAE and the highest per-perturbation and per-gene correlations before task-specific readout calibration.

\begin{table}[H]
\vspace{-4pt}
\centering\small
\caption{\textbf{Signed response-profile prediction on test perturbations.} Means weight cell lines equally and cover 1,264 test perturbations. Errors are averaged over perturbations and genes against measured batch-matched $\log(1+\mathrm{CP10K})$ responses, with MSE scaled by $10^{3}$ and MAE by $10^{2}$. Pearson $r$ is computed per perturbation across genes and per gene across perturbations, excluding constant vectors (dashes mark undefined values). Gene means predict each gene's mean training response. GEARS, STATE, PerturbDiff, and PertOmni are reproduced on the official splits (\appref{app:baseline-data}).}
\label{tab:profile-comparison}
\setlength{\tabcolsep}{5pt}
\begin{tabular}{@{}lrrrr@{}}
\toprule
Predictor & MSE $\downarrow$ & MAE $\downarrow$ & Per-perturbation $r$ $\uparrow$ & Per-gene $r$ $\uparrow$\\
\midrule
RELAY & \textbf{4.849} & \textbf{4.565} & \textbf{.5165} & \textbf{.3141}\\
PertOmni & 5.160 & 4.664 & .4924 & .2781\\
Gene means & 6.154 & 4.962 & .4190 & ---\\
PerturbDiff & 6.322 & 5.010 & .4064 & .1904\\
STATE & 6.504 & 5.041 & .3335 & .1811\\
GEARS & 6.851 & 5.422 & .3019 & .0535\\
Zero response & 7.905 & 5.352 & --- & ---\\
\bottomrule
\end{tabular}
\vspace{-6pt}
\end{table}

Aggregate MSE alone does not reveal which part of the predicted profile fails. Gene means govern global accuracy, whereas contrasts around them govern within-gene discrimination, and a predicted contrast can err in either directional alignment or amplitude. To separate these factors, let $m(g)$ and $\widehat m(g)$ denote the means of the measured and predicted responses over test perturbations, and write the test-centered responses as $r(p,g)=m(g)+\delta(p,g)$ and $\widehat r(p,g)=\widehat m(g)+\widehat\delta(p,g)$. Because $\sum_p(\widehat\delta-\delta)=0$ for every gene, the cross term between gene-mean error and contrast error vanishes. With overlines averaging over test perturbations and genes, let $\sigma_\delta^2=\overline{\delta^2}$ and $\sigma_{\widehat\delta}^2=\overline{\widehat\delta^2}$, with pooled contrast correlation $\rho=\overline{\widehat\delta\,\delta}/(\sigma_{\widehat\delta}\sigma_\delta)$ and amplitude ratio $\eta=\sigma_{\widehat\delta}/\sigma_\delta$ (taking $\eta=\rho=0$ when $\sigma_{\widehat\delta}=0$). Expanding the contrast error as $\overline{(\widehat\delta-\delta)^2}=\sigma_\delta^2(1+\eta^2-2\rho\eta)=\sigma_\delta^2[(1-\rho^2)+(\eta-\rho)^2]$ yields the exact decomposition
\begin{equation}\label{eq:mse-split}
 \mathrm{MSE}=\overline{(\widehat m-m)^2}+\sigma_\delta^2(1-\rho^2)+\sigma_\delta^2(\eta-\rho)^2.
\end{equation}
The three terms represent gene-level offset error, unexplained contrast error $\sigma_\delta^2(1-\rho^2)$ (the residual alignment error after optimal linear rescaling), and amplitude error $\sigma_\delta^2(\eta-\rho)^2$, which vanishes only when $\eta=\rho$. At a fixed gene component, predicted contrasts reduce MSE below zero-contrast error by $\sigma_\delta^2\eta(2\rho-\eta)$, strictly improving profile accuracy if and only if $0<\eta<2\rho$.

\begin{table}[H]
\vspace{-4pt}
\centering\small
\caption{\textbf{Decomposition of held-out profile error.} Means weight cell lines equally on test perturbations, with errors scaled by $10^{3}$. Slope denotes the regression of the predictor's gene component on the training gene means across genes, and brackets report 95\% perturbation-bootstrap intervals. Slope is ideal at 1 and $\eta$ at $\rho$. Each term of Equation~(3) is computed within a cell line and then averaged, so the reported amplitude error is the mean of $\sigma_\delta^2(\eta-\rho)^2$ across cell lines and need not equal the same expression evaluated at the averaged $\eta$ and $\rho$.}
\label{tab:continuous-split}
\setlength{\tabcolsep}{2.2pt}
\begin{tabular}{@{}lrrrrrrr@{}}
\toprule
Predictor & MSE $\downarrow$ & Gene-level $\downarrow$ & Amplitude $\downarrow$ & Unexplained $\downarrow$ & Slope & $\rho$ $\uparrow$ & $\eta$\\
\midrule
Zero response & 7.905 & 1.784 & 0 & 6.120 & 0 & --- & 0\\
Gene means & 6.154 & 0.033 & 0 & 6.120 & 1 & --- & 0\\
GEARS & 6.851 & 0.570 & 0.341 & 5.940 & .829 [.811,.848] & .119 [.106,.133] & .362\\
STATE & 6.504 & 0.430 & 0.434 & 5.640 & .569 [.533,.605] & .257 [.231,.281] & .479\\
PerturbDiff & 6.322 & 0.138 & 0.586 & 5.599 & .946 [.901,.994] & .289 [.262,.314] & .594\\
PertOmni & 5.160 & 0.032 & 0.062 & 5.066 & 1.078 [1.054,1.104] & .415 [.393,.436] & .481\\
RELAY & \textbf{4.849} & \textbf{0.023} & 0.053 & \textbf{4.772} & 1.053 [1.027,1.078] & \textbf{.470} [.450,.489] & .512\\
\bottomrule
\end{tabular}
\vspace{-6pt}
\end{table}

Table~\ref{tab:continuous-split} shows that while GEARS, STATE, and PerturbDiff reduce unexplained contrast error relative to the training gene means, their gene-level and amplitude errors more than offset this gain. First, GEARS and STATE substantially misestimate gene-level offsets: regressing each predictor's gene component on the training gene means yields an average slope of 0.57 for STATE (0.21--0.81 across cell lines), while GEARS deviates from unit slope in both directions (0.64--0.74 in three cell lines and 1.27 in RPE1). PerturbDiff tracks training gene means much more closely (slope 0.95; gene-level error 0.138) and achieves higher contrast alignment than STATE ($\rho=0.289$ vs.\ $0.257$), yet its diffusion samples inflate contrast amplitude to $\eta=0.594\approx 2.06\rho$, producing the largest amplitude error of any baseline ($0.586\times 10^{-3}$); on validation perturbations, increasing PerturbDiff's classifier-free guidance scale from 0.0 to 1.5 monotonically enlarges $\eta$ and worsens MSE in every cell line. Contrast amplitudes similarly exceed the MSE-optimal ratio $\eta/\rho=1$ for STATE ($1.72$ on average) and GEARS ($1.6$--$7.9$ in HepG2, Jurkat, and RPE1, whereas GEARS's K562 contrasts are uncorrelated with the measured ones, $\rho=0.001$). Consistent with this amplitude overshoot, VCWorld finds on GeneTAK that STATE and scVI \citep{lopez2018scvi} call two to five times too many DE genes \citep{wei2026vcworld}. By contrast, PertOmni averages measured training responses of learned nearest neighbors rather than decoding expression directly; this nearly eliminates gene-level error ($0.032$) and controls contrast amplitude ($\eta/\rho=1.16$, amplitude error $0.062$, $\rho=0.415$, $\eta=0.481$), lowering profile MSE below training gene means and surpassing the gene prior on DIR Macro-F1, yet directly thresholding its predicted response magnitude still leaves DE Macro-F1 7.9 points below the gene prior. RELAY---a ridge regression with training gene means as its default---attains $\eta/\rho=1.09$ while further reducing unexplained contrast error from 5.066 to 4.772.

Replacing each direct predictor's gene component with the training gene means ($\mu+\widehat\delta$) isolates the downstream impact of these errors. Restoring $\mu$ lowers MSE from 6.504 to 6.107 for STATE, from 6.851 to 6.314 for GEARS, and from 6.322 to 6.219 for PerturbDiff (validation-centered), compared with 6.154 for the training gene means (all $\times10^{-3}$), and under the diagnostic sign threshold ($\widehat r > 0$; \appref{app:continuous-diagnosis}) raises global DIR Macro-F1 from 0.733 to 0.754 for STATE and from 0.675 to 0.709 for GEARS. Adding a per-gene constant or rescaling $\widehat\delta$ by a positive factor leaves signed within-gene rankings unchanged, and empirically within-gene decision accuracy does not improve either (\appref{app:continuous-diagnosis}). Combining gene-mean restoration with an oracle rescaling of $\widehat\delta$ to its MSE-optimal amplitude ($\eta=\rho$) further reduces MSE to 5.67 for STATE, 5.97 for GEARS, and 5.63 for PerturbDiff, compared with 4.81 for RELAY under the same rescaling (all $\times10^{-3}$). Even after correcting both gene offsets and contrast amplitudes, a substantial alignment gap remains, as unexplained contrast error accounts for 52\% of STATE's, 58\% of GEARS's, and 56\% of PerturbDiff's excess MSE relative to RELAY in Table~\ref{tab:continuous-split} (with amplitude error contributing another 36\% for PerturbDiff).

Together, the diagnostics across Sections~\ref{sec:mechanisms}--\ref{sec:response-learning} address our central question---how to recover perturbation-specific variation while preserving reliable gene-level tendencies---by motivating four design choices:
\begin{itemize}\setlength{\itemsep}{1pt}\setlength{\parskip}{0pt}\setlength{\parsep}{0pt}
 \item \textbf{Mean anchoring.} Because gene-level offsets $\overline{(\widehat m-m)^2}$ degrade global accuracy, predictions should explicitly retain training gene means rather than re-estimating them.
 \item \textbf{Shrinkage-controlled contrast transfer.} Because discrete labels cover only about 1\% of pairs whereas deep predictors inflate residual amplitudes ($\eta>\rho$), perturbation-specific changes should be transferred from dense measured responses as regularized contrasts around those means.
 \item \textbf{Sign-invariant magnitude view.} Because related perturbations can regulate a target gene in opposite directions, DE transfer needs an absolute-deviation view alongside signed contrasts so opposite-signed effects do not cancel.
 \item \textbf{Task-specific calibrated readouts.} Because thresholding raw responses trades away reliable gene priors, discrete DE and DIR decisions must integrate baseline gene tendencies, profile-wide response context, and measurement uncertainty.
\end{itemize}

% !TEX root = ../main.tex
\section{RELAY: Response Transfer for DE and DIR Prediction}\label{sec:method}\label{sec:diag-implications}

\begin{figure}[p]
\centering
\includegraphics[height=0.875\textheight,keepaspectratio]{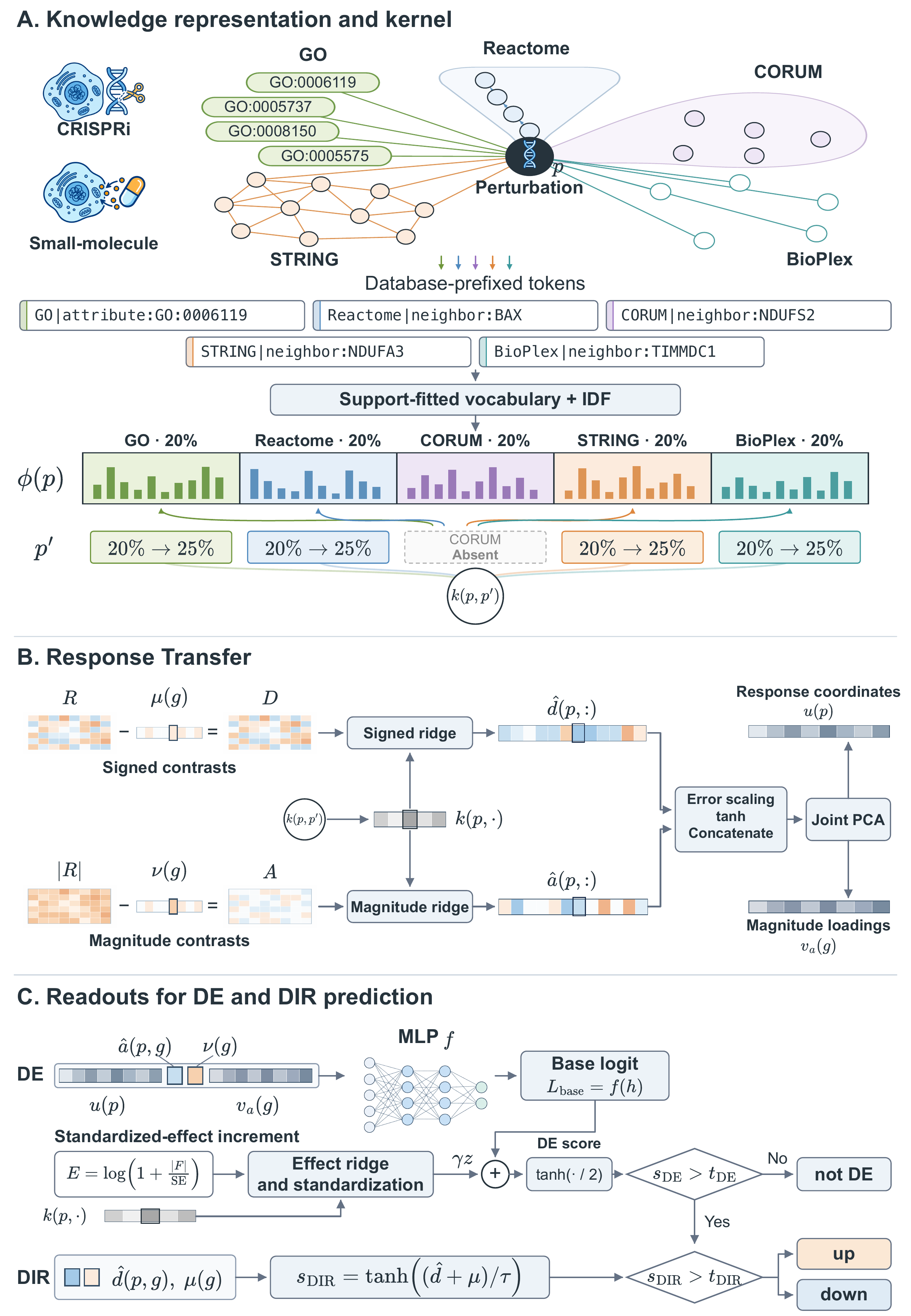}
\caption{\textbf{RELAY's response-transfer pipeline.} \textbf{A,} Knowledge representations define a knowledge kernel $k(p,p')$. \textbf{B,} Kernel ridge regressions use the query's similarities $k(p,\cdot)$ to transfer signed and magnitude contrasts, which a joint PCA summarizes for the DE readout. \textbf{C,} Readouts for DE and DIR prediction combine gene-level tendencies with the transferred contrasts. Source percentages are normalization weights rather than learned weights, and edges denote curated annotations and interaction partners.}
\label{fig:perturbfactor-overview}
\end{figure}

Building on these four design choices, we design \textbf{RELAY} to decouple baseline gene tendencies from perturbation-specific variation and map both to calibrated decisions without a language model (Figure~\ref{fig:perturbfactor-overview}). Unlike similarity-based profile predictors that average raw neighbor profiles \citep{fawkes2026kg,long2026pertomni} or retrieval methods that vote over sparse discrete labels \citep{yuan2026core}, RELAY anchors on empirical training gene means to support global accuracy, transfers dense signed and magnitude contrasts across related perturbations with regularized shrinkage, and converts the transferred responses into discrete DE and DIR predictions through task-specific readouts.

For an unseen query $(p,g)$, RELAY operates in three steps corresponding to these choices. First, it decomposes the training response matrix around per-gene means into centered signed and magnitude contrasts so opposite-signed effects do not cancel (Section~\ref{sec:method-decomposition}). Second, it encodes $p$ as a multi-source knowledge representation and estimates transfer weights across all training perturbations via kernel ridge regression over the resulting knowledge kernel, de-correlating redundant neighbors and shrinking noisy contrasts toward zero to control amplitude error (Section~\ref{sec:method-shrinkage}). Third, task-specific readouts re-anchor the transferred contrasts on gene-level tendencies: the DIR readout combines the signed contrast with the gene mean, whereas the DE readout evaluates the local magnitude contrast within transcriptome-wide PCA context, optionally refined by transferred measurement-precision statistics when available (Section~\ref{sec:method-readout}). Adapting source aggregation or target pooling extends RELAY to held-out cell lines and unseen compounds (Section~\ref{sec:method-beyond}).

\subsection{Response Contrasts}\label{sec:method-decomposition}
Our diagnostic decomposition in Section~\ref{sec:response-learning} showed that end-to-end profile predictors lose global accuracy through gene-offset errors $\overline{(\widehat m-m)^2}$. Meanwhile, because the knowledge kernel measures shared pathway or complex membership rather than regulatory sign, training perturbations that are functionally related to $p$ can include both activators and repressors that shift the same target gene $g$ in opposite directions. Summing only signed responses across such training neighbors can cancel out large effects of opposite sign, whereas transferring response magnitudes $|r(p',g)|$ directly captures whether perturbing that pathway strongly affects $g$. For a training subset $\mathcal S\subseteq\mathcal P_{\mathrm{tr}}$ of $n=|\mathcal S|$ perturbations, equal to $\mathcal P_{\mathrm{tr}}$ at inference or an inner cross-fitting fold, let $R\in\mathbb R^{n\times|\mathcal G|}$ denote the response matrix with entries $r(p,g)$ ($p\in\mathcal S, g\in\mathcal G$, Section~\ref{sec:information}) and elementwise absolute value $|R|$. To retain baseline tendencies and avoid sign cancellation during DE transfer, RELAY centers each gene column of $R$ and $|R|$, defining the signed and magnitude contrasts
\begin{equation}\label{eq:pf-decomposition}
 d(p,g)=r(p,g)-\mu(g),\quad a(p,g)=|r(p,g)|-\nu(g),
\end{equation}
where $\mu(g)=n^{-1}\sum_{p\in\mathcal S}r(p,g)$ and $\nu(g)=n^{-1}\sum_{p\in\mathcal S}|r(p,g)|$ are the support column means of $R$ and $|R|$. Stacking these contrasts across $\mathcal S\times\mathcal G$ yields matrices $D,A\in\mathbb R^{n\times|\mathcal G|}$ (Figure~\ref{fig:perturbfactor-overview}B) for transfer to unseen perturbations.

\paragraph{Local response-transfer model.}
To formalize why transferring $A$ alongside $D$ is necessary when biological kernels capture functional similarity without regulatory sign (Section~3.2), we examine how \emph{any} fixed linear transfer rule behaves when functionally related perturbations agree with the query only partially in sign. Fix a target gene $g$ and an unseen query $p$ whose true contrast around the gene baseline $\mu_0(g)\in\mathbb R$ is $\zeta_p\theta_{p,g}$, with direction $\zeta_p\in\{-1,+1\}$ and magnitude $\theta_{p,g}>0$. Let $\mathcal N(p)\subseteq\mathcal S$ denote the $m$ support perturbations functionally related to $p$, and write the support responses as
\[
r(p',g)=\mu_0(g)+x_{p'}+\varepsilon_{p'},\qquad
x_{p'}=\begin{cases} s_{p'}\,\zeta_p\,\theta_{p',g}, & p'\in\mathcal N(p),\\ 0, & p'\in\mathcal S\setminus\mathcal N(p).\end{cases}
\]
Here $\theta_{p',g}=\theta_{p,g}+\xi_{p'}$ allows neighbor-to-neighbor effect heterogeneity with $\E[\xi_{p'}]=0$ and $\mathrm{Var}(\xi_{p'})=\sigma_\xi^2$; the sign factor $s_{p'}\in\{-1,+1\}$ satisfies $\Pr(s_{p'}=+1)=q\in[0,1]$, defining the \emph{net sign alignment} $\E[s_{p'}]=2q-1$; $\varepsilon_{p'}$ is zero-mean measurement noise with $\mathrm{Var}(\varepsilon_{p'})=\sigma_{\varepsilon,p'}^2<\infty$; and $\{s_{p'},\xi_{p'},\varepsilon_{p'}\}_{p'\in\mathcal S}$ are mutually independent given $\zeta_p$. Given the sample-centered contrasts $d,a$ of Equation~\eqref{eq:pf-decomposition} and arbitrary fixed transfer weights $w\in\mathbb R^{n}$ (possibly negative and not summing to one, including the kernel ridge weights of Section~\ref{sec:method-shrinkage}), the transferred contrasts $\hat d(p,g)=\sum_{p'\in\mathcal S}w_{p'}d(p',g)$ and $\hat a(p,g)=\sum_{p'\in\mathcal S}w_{p'}a(p',g)$ aggregate $\mathcal N(p)$ relative to the mean weight $\bar w=n^{-1}\one_n^\top w$ through the \emph{effective neighborhood weight}
\[
W_{\mathrm{eff}}(p)=\sum_{p'\in\mathcal N(p)}\bigl(w_{p'}-\bar w\bigr).
\]

\begin{proposition}[Sign attenuation vs.\ magnitude preservation under unsigned kernels]
Under the local response-transfer model above (proof in \appref{app:proofs}):
\begin{enumerate}
\item \textbf{(Signed transfer scales with net sign alignment.)} Without symmetry or distributional assumptions on $\varepsilon_{p'}$,
\[
\E\bigl[\hat d(p,g)\mid\zeta_p\bigr]=\underbrace{(2q-1)}_{\substack{\text{net sign}\\\text{alignment}}}\,W_{\mathrm{eff}}(p)\,\zeta_p\,\theta_{p,g},
\]
\[
\mathrm{Var}\bigl(\hat d(p,g)\mid\zeta_p\bigr)=
\underbrace{\bigl(4q(1-q)\theta_{p,g}^2+\sigma_\xi^2\bigr)\!\!\sum_{p'\in\mathcal N(p)}\!\!(w_{p'}-\bar w)^2}_{\text{sign-mixing and effect heterogeneity}}
+\underbrace{\sum_{p'\in\mathcal S}\sigma_{\varepsilon,p'}^2(w_{p'}-\bar w)^2}_{\text{measurement noise}} .
\]
Thus $\hat d(p,g)$ preserves the query direction $\zeta_p$ in expectation whenever $(2q-1)W_{\mathrm{eff}}(p)>0$, whereas balanced sign disagreement ($q=1/2$) eliminates the expected signed signal for every fixed $w$.

\item \textbf{(Magnitude transfer preserves positive signal.)} Suppose $(s_{p'},\xi_{p'})$ are identically distributed across $p'\in\mathcal N(p)$ and $\varepsilon_{p'}$ are identically distributed across $p'\in\mathcal S$ with a continuous symmetric distribution around zero ($\varepsilon_{p'}\stackrel{d}{=}-\varepsilon_{p'}$, CDF $F_\varepsilon$) supported on all of $\mathbb R$ (for Part~2 alone, finite variances may be relaxed to $\E|\xi_{p'}|+\E|\varepsilon_{p'}|<\infty$). Define the folded-noise expectation $\psi(t)=\E|t+\varepsilon_{p'}|$ and the related-neighbor expected magnitude $G=\E\bigl[\psi\bigl(\mu_0(g)+s_{p'}\zeta_p\theta_{p',g}\bigr)\mid\zeta_p\bigr]$. Then
\begin{align*}
\E\bigl[\hat a(p,g)\mid\zeta_p\bigr]&=W_{\mathrm{eff}}(p)\,\underbrace{\bigl(G-\psi(\mu_0(g))\bigr)}_{\substack{\text{excess expected}\\\text{magnitude}}},\\
G&\ge q\,\psi\bigl(\mu_0(g)+\zeta_p\theta_{p,g}\bigr)+(1-q)\,\psi\bigl(\mu_0(g)-\zeta_p\theta_{p,g}\bigr).
\end{align*}
Whenever $W_{\mathrm{eff}}(p)>0$, $\E[\hat a(p,g)\mid\zeta_p]>0$ holds across two full-dimensional regimes:
\begin{enumerate}
\item \emph{Baseline-aligned or balanced regime}, $(2q-1)\,\zeta_p\,\mu_0(g)\ge 0$ (subsuming $\mu_0(g)=0$ for all $q\in[0,1]$ and $q=1/2$ for all $\mu_0(g)\in\mathbb R$):
\begin{align*}
G&\;\ge\;\tfrac12\bigl[\psi(\mu_0(g)+\theta_{p,g})+\psi(\mu_0(g)-\theta_{p,g})\bigr]\;>\;\psi(\mu_0(g)),\\
G&\;\ge\;\max\bigl(\theta_{p,g},\ \psi(\mu_0(g))\bigr),
\end{align*}
with $G-\psi(0)\ge\psi(\theta_{p,g})-\psi(0)=\int_0^{\theta_{p,g}}(2F_\varepsilon(t)-1)\,dt>0$ at $\mu_0(g)=0$.
\item \emph{Baseline-dominating effect regime}, $\theta_{p,g}>2|\mu_0(g)|$ for every $q\in[0,1]$ and $\zeta_p\in\{-1,+1\}$:
\[
G\;\ge\;\psi\bigl(\theta_{p,g}-|\mu_0(g)|\bigr)\;>\;\psi(\mu_0(g)).
\]
\end{enumerate}
\end{enumerate}
\end{proposition}

Proposition~1 separates the roles of the two views whenever $W_{\mathrm{eff}}(p)>0$. When related perturbations tend to share the query's sign ($q>1/2$), the signed contrast preserves the query direction $\zeta_p$ for DIR, but partial sign disagreement shrinks its expected amplitude by $|2q-1|$. By contrast, the magnitude contrast remains strictly positive in expectation throughout both the baseline-aligned and baseline-dominating regimes---including at $q=1/2$, where the signed contrast carries zero expected signal---preserving DE sensitivity under unsigned biological kernels. When $\varepsilon_{p'}$ has a continuous density at zero with $f_\varepsilon(0)>0$ (for example $\psi(0)=\sigma_\varepsilon\sqrt{2/\pi}$ and $f_\varepsilon(0)=(\sigma_\varepsilon\sqrt{2\pi})^{-1}$ under $\mathcal N(0,\sigma_\varepsilon^2)$), the zero-baseline excess magnitude scales across signal regimes as
\[
\psi(\theta_{p,g})-\psi(0)=\begin{cases}
f_\varepsilon(0)\,\theta_{p,g}^2+o(\theta_{p,g}^2), & \theta_{p,g}\to 0^+,\\[3pt]
\theta_{p,g}-\E|\varepsilon_{p'}|+o(1), & \theta_{p,g}\to\infty,
\end{cases}
\]
rising only quadratically below the noise level and linearly once the effect exceeds it, so the magnitude view does not rescue effects that sit below the measurement noise.

PerturbQA lies squarely between the two extremes of sign concordance. DIR, which is scored from $\hat d$ alone, reaches $.8611$ pgAUROC (Table~6), so related perturbations agree with the query in sign more often than not. Replacing $|\hat d+\mu|$ with $\hat a+\nu$ nevertheless raises DE pgAUROC by $3.99$ points (Table~7a), which shows that this agreement is far from complete (Section~5.3).

\begin{remark}[From arbitrary weights to kernel ridge transfer]
Proposition~1 applies to any weight vector with $W_{\mathrm{eff}}(p)>0$. When $1\le m<n$, convex weights supported on $\mathcal N(p)$ give $W_{\mathrm{eff}}(p)=1-m/n>0$, and the kernel ridge weights $w_\alpha(p)=(K+\alpha I_n)^{-1}k_p$ of Section~\ref{sec:method-shrinkage} ($\alpha>0$) on an idealized block kernel---$k(p',p'')=k_0\in(0,1]$ for distinct $p',p''\in\mathcal N(p)\cup\{p\}$, $k(p',p')=1$, and zero elsewhere---yield
\[
W_{\mathrm{eff}}(p)=\Bigl(1-\frac{m}{n}\Bigr)\frac{k_0\,m}{1+\alpha+k_0(m-1)}>0 .
\]
On general non-block kernels $K\succeq 0$, this structural condition can be replaced directly by cross-moments between support and query contrasts (observed with noise independent of the query). Let $c_{a,p}=\E[a(p,g)\,a_g]$ denote the magnitude cross-moment between query $p$ and support vector $a_g\in\mathbb R^n$, with $\hat a_\alpha(p,g)=w_\alpha(p)^\top a_g$. Whenever raw magnitude alignment $k_p^\top c_{a,p}>0$ holds,
\[
\E\bigl[\hat a_\alpha(p,g)\,a(p,g)\bigr]=k_p^\top(K+\alpha I_n)^{-1}c_{a,p}>0
\]
for every $\alpha>0$ under proportional population alignment $c_{a,p}=\sigma_a^2k_p$ ($\sigma_a^2>0$, $k_p\neq 0$), and for every penalty above an explicit finite threshold $\alpha_0$ without proportionality, including under empirical support centering (explicit $\alpha_0$ derived in \appref{app:proofs}), independently of sign concordance. Conversely, if sign disagreement scales the population-centered signed cross-moment $\E[\delta(p,g)\,y_g]=\chi\,\sigma_\delta^2k_p$ by $\chi\in[-1,1]$, then
\[
\E\bigl[\hat d_\alpha(p,g)\,\delta(p,g)\bigr]=\chi\,\sigma_\delta^2\,k_p^\top(K+\alpha I_n)^{-1}k_p ,
\]
which vanishes as $\chi\to0$. The separation between the two views therefore carries over directly to RELAY's kernel ridge weights.
\end{remark}

\subsection{Response Transfer over a Knowledge Kernel}\label{sec:method-encoding}\label{sec:method-shrinkage}
Transferring contrasts to an unseen perturbation requires quantifying how closely it relates to each training perturbation. Because existing knowledge graphs rarely specify a transcriptional response directly (Section~\ref{sec:mechanisms}) yet reliably identify functionally related perturbations (Section~\ref{sec:retrieval-transfer}), biological knowledge enters RELAY strictly as a similarity measure rather than as a heuristic neighbor average. Specifically, a \emph{knowledge representation} $\phi(p)$ encodes the annotations and interaction partners of perturbed gene $p$ across GO \citep{ashburner2000go,aleksander2023go}, Reactome \citep{milacic2024reactome}, CORUM \citep{tsitsiridis2023corum}, STRING \citep{szklarczyk2023string}, and BioPlex \citep{huttlin2021bioplex} as database-prefixed tokens (e.g., \texttt{STRING|neighbor:NDUFA3}, where the prefix keeps vocabularies disjoint across databases).

Computing vocabularies and inverse document frequency (IDF) weights \citep{sparckjones1972idf} on the support $\mathcal S$ only, we $L_2$-normalize the IDF-weighted token vector within each source $j$ and then normalize across the active sources of $p$. Let $\kappa_j(p,p')$ denote the resulting cosine similarity within source $j$, and let $\ind_j(p)\in\{0,1\}$ indicate whether $p$ retains at least one token in source $j$, with active source count $h(p)=\sum_j\ind_j(p)$. The inner product $k(p,p')=\phi(p)^\top\phi(p')$ then decomposes across sources as
\begin{equation}\label{eq:abl-kernel}
 k(p,p')=\frac{\sum_j\ind_j(p)\ind_j(p')\kappa_j(p,p')}{\sqrt{h(p)h(p')}}\quad\text{if }h(p)h(p')>0,\quad\text{and }0\text{ otherwise},
\end{equation}
so each nonzero representation $\phi(p)$ has unit length while pairwise similarity is normalized by the active source counts of both perturbations rather than their intersection alone. As a consequence, the similarity between two perturbations with different annotation coverage is bounded by $\sqrt{\min(h(p),h(p'))/\max(h(p),h(p'))}$ (for example $1/\sqrt5$ between a perturbation annotated in one source and one annotated in all five), shrinking similarities for sparsely annotated queries relative to richly annotated support perturbations so their transferred contrasts stay closer to zero and their predictions stay closer to the gene means.

Given these similarities, RELAY transfers contrasts from the entire support $\mathcal S$ while controlling prediction amplitude via kernel ridge regression \citep{hoerl1970ridge,saunders1998krr}. Let $X\in\mathbb R^{n\times V}$ stack the support representations $\phi(p')$ as rows ($V$ being the support vocabulary size), with Gram matrix $K=XX^\top$ and query similarity vector $k_p=(k(p,p'))_{p'\in\mathcal S}\in\mathbb R^n$. For each contrast matrix $B\in\{D,A\}$, ridge regression with a shared cell-line penalty $\alpha>0$ predicts
\begin{equation}\label{eq:pf-ridge}
\begin{split}
 \widehat B(p,:)&=\phi(p)^\top W_B^*=k_p^\top(K+\alpha I_n)^{-1}B,\\
 W_B^*&=\arg\min_W\ \|XW-B\|_F^2+\alpha\|W\|_F^2.
\end{split}
\end{equation}
Both views thus share the transfer coefficients $w(p)=(K+\alpha I_n)^{-1}k_p$, so the predicted contrast rows $\widehat d(p,:)=w(p)^\top D$ and $\widehat a(p,:)=w(p)^\top A$ combine the rows of $D$ and $A$. Unlike unweighted neighbor averaging, the inverse Gram factor $(K+\alpha I_n)^{-1}$ accounts for redundancy among correlated training perturbations, allowing weights to take negative values and not sum to one. At the same time, the penalty $\alpha$ shrinks weak predictions toward zero, controlling the amplitude overshoot diagnosed in Section~\ref{sec:response-learning}. In the limit of an unannotated query ($k_p=0$), the predicted contrasts vanish and the response falls back cleanly to the training gene means.

\paragraph{Working covariance model and background similarity.}
Beyond this fallback property, the inverse Gram operator $M_\alpha=(K+\alpha I_n)^{-1}$ in Equation~\eqref{eq:pf-ridge} provides two structural advantages over plain neighbor averaging: its penalty governs on which side of the calibrated amplitude $\eta=\rho$ the pooled prediction falls (Equation~\eqref{eq:mse-split}), and an isotropic background similarity across unrelated perturbations avoids the signal dilution of neighbor averaging on centered contrasts. To formalize both properties across a target gene set $\mathcal G$ and an arbitrary support $\mathcal S$ of $n$ perturbations (including any unannotated perturbations with $h(p')=0$, where $K_{p',:}=0$ and $(k_p)_{p'}=0$ by Equation~\eqref{eq:abl-kernel}), let $d_g\in\mathbb R^n$ and $\delta(p,g)$ denote zero-mean support and query contrasts around the population gene mean (distinct from the test-centered diagnostic $\delta$ of Section~\ref{sec:response-learning}; \appref{app:proofs}), observed with zero-mean noise $\varepsilon_g$ as $y_g=d_g+\varepsilon_g$. Conditionally on $k_p$, suppose $\E[\delta(p,g)^2]=\sigma_{\delta,g}^2$, $\E[d_{p',g}d_{p'',g}]=\sigma_{\delta,g}^2k(p',p'')$ and $\E[\delta(p,g)d_{p',g}]=\sigma_{\delta,g}^2k(p,p')$ on annotated support coordinates ($h(p'),h(p'')>0$), $\E[\varepsilon_g\varepsilon_g^\top]=\sigma_{\varepsilon,g}^2I_n$, $\E[d_g\varepsilon_g^\top]=0$, and $\E[\varepsilon_g\delta(p,g)]=0$ (allowing arbitrary correlation across $g\in\mathcal G$). Write $\hat d_\alpha(p,g)=k_p^\top M_\alpha y_g$ and define the \emph{pooled noise-to-signal ratio}
\[
\alpha^\ast=\frac{\bar\sigma_\varepsilon^2}{\bar\sigma_\delta^2},
\qquad
\bar\sigma_\delta^2=\frac{1}{\lvert\mathcal G\rvert}\sum_{g\in\mathcal G}\sigma_{\delta,g}^2>0,\quad
\bar\sigma_\varepsilon^2=\frac{1}{\lvert\mathcal G\rvert}\sum_{g\in\mathcal G}\sigma_{\varepsilon,g}^2>0 .
\]
To model non-specific similarity between unrelated perturbations (such as broad text-embedding baselines in Section~\ref{sec:main-hypothesis}), let $\tilde K=(1-b)K+b\one_n\one_n^\top$ and $\tilde k_p=(1-b)k_p+b\one_n$ mix $K$ with an all-ones background of level $b\in(0,1)$, and set $A_\alpha=(1-b)K+\alpha I_n$.

\begin{proposition}[Amplitude calibration and background similarity under anchored KRR]
Under the setup above (proof in \appref{app:proofs}):
\begin{enumerate}
\item \textbf{(Pooled amplitude calibration.)} Under the working covariance model, for every $\alpha>0$ with $\E_p[k_p^\top M_\alpha k_p]>0$, the population-pooled amplitude ratio $\eta(\alpha)$ and correlation $\rho(\alpha)$ across $\mathcal G$ satisfy
\[
\frac{\eta(\alpha)}{\rho(\alpha)}
\;=\;1+\underbrace{(\alpha^\ast-\alpha)}_{\substack{\text{under- vs.\ over-}\\\text{shrinkage}}}\,
\underbrace{\frac{\E_p\bigl[k_p^\top M_\alpha^{2}k_p\bigr]}{\E_p\bigl[k_p^\top M_\alpha k_p\bigr]}}_{>0} .
\]
Consequently, $\eta(\alpha)>\rho(\alpha)$ if and only if $\alpha<\alpha^\ast$. At $\alpha=\alpha^\ast$, $\eta(\alpha^\ast)=\rho(\alpha^\ast)=\sqrt{\E_p[k_p^\top M_{\alpha^\ast}k_p]}$, the pooled amplitude error $\bar\sigma_\delta^2(\eta-\rho)^2$ of Equation~\eqref{eq:mse-split} vanishes, and expected contrast MSE falls below the gene-mean baseline by $\bar\sigma_\delta^2\rho(\alpha^\ast)^2$.

\item \textbf{(Resilience to isotropic background similarity.)} For any $K\succeq 0$, $b\in(0,1)$, and $y\in\mathbb R^{n}$,
\begin{align*}
\tilde k_p^\top(\tilde K+\alpha I_n)^{-1}y
&\;=\;
\underbrace{k_p^\top\Bigl(K+\tfrac{\alpha}{1-b}I_n\Bigr)^{-1}y}_{\text{selective transfer (inflated penalty)}}
\;+\;
\underbrace{\Delta_{\mathrm{irr}}(p,y)}_{\text{irregularity residual}},\\
\Delta_{\mathrm{irr}}(p,y)&=\frac{b\,\bigl(1-(1-b)\,k_p^\top A_\alpha^{-1}\one_n\bigr)}{1+b\,\one_n^\top A_\alpha^{-1}\one_n}\,\one_n^\top A_\alpha^{-1}y .
\end{align*}
For every centered contrast vector $y$ ($\one_n^\top y=0$, including every column of $D$ and $A$):
\begin{enumerate}
\item \emph{Regular kernels:} If $K\one_n=\bar\kappa\one_n$ ($\bar\kappa=n^{-1}\one_n^\top K\one_n$), then $\Delta_{\mathrm{irr}}(p,y)=0$ identically, so background similarity acts solely by inflating $\alpha$ to $\alpha/(1-b)$.
\item \emph{Arbitrary PSD kernels:} The centered inner product $\one_n^\top A_\alpha^{-1}y=v_{\alpha,\perp}^\top y$ is controlled via the \emph{kernel degree dispersion} $\lVert K\one_n-\bar\kappa\one_n\rVert_2$, since $v_{\alpha,\perp}=(I_n-n^{-1}\one_n\one_n^\top)A_\alpha^{-1}\one_n$ satisfies
\[
\lVert v_{\alpha,\perp}\rVert_2\;\le\;\frac{1-b}{\alpha\bigl(\alpha+(1-b)\bar\kappa\bigr)}\,\lVert K\one_n-\bar\kappa\one_n\rVert_2
\]
(with prefactor bounded in magnitude by $\sqrt{b/(1+b\,\one_n^\top A_\alpha^{-1}\one_n)}$ whenever the joint kernel on $\mathcal S\cup\{p\}$ is positive semidefinite with $k(p,p)\le 1$).
\item \emph{Contrast with neighbor averaging:} For kernel-weighted averaging of centered $y$ when $k_p^\top\one_n>0$ (and $\hat d_{\mathrm{NW}}(p,g)=0$ if $k_p=0$),
\[
\hat d_{\mathrm{NW}}(p,g)=\frac{\tilde k_p^\top y}{\tilde k_p^\top\one_n}=(1-\lambda_p)\,\frac{k_p^\top y}{k_p^\top\one_n},
\qquad \lambda_p=\frac{nb}{(1-b)k_p^\top\one_n+nb}\in(0,1),
\]
For fixed $b\in(0,1)$, the attenuation factor $1-\lambda_p$ tends to zero as $n\to\infty$ whenever $k_p^\top\one_n=o(n)$; it is $O(n^{-1})$ if $k_p^\top\one_n=O(1)$.
\end{enumerate}
\end{enumerate}
\end{proposition}

\begin{remark}[Misspecified kernel and the selected penalty]
Proposition~2 models $K$ as the exact contrast covariance on annotated perturbations, whereas empirical alignment on PerturbQA is partial ($0.172$, Table~8). Under arbitrary conditional second moments $\Sigma_{y,g}=\E[y_gy_g^\top\mid k_p]$ and cross-moments $c_{p,g}=\E[\delta(p,g)y_g\mid k_p]$ with finite $\bar\sigma_\delta^2>0$ and pooled averages $\overline{\Sigma}_y=\lvert\mathcal G\rvert^{-1}\sum_{g\in\mathcal G}\Sigma_{y,g}$ and $\bar c_p=\lvert\mathcal G\rvert^{-1}\sum_{g\in\mathcal G}c_{p,g}$, the pooled amplitude-to-correlation ratio and its large-penalty asymptotic whenever positive raw alignment $\E_p[k_p^\top\bar c_p]>0$ holds (which by Cauchy--Schwarz also implies $\E_p[k_p^\top\overline{\Sigma}_yk_p]>0$) take the form
\[
\frac{\eta(\alpha)}{\rho(\alpha)}
=\frac{\E_p\bigl[k_p^\top M_\alpha\overline{\Sigma}_yM_\alpha k_p\bigr]}{\E_p\bigl[k_p^\top M_\alpha\bar c_p\bigr]}
\;=\;\frac{1}{\alpha}\,\frac{\E_p\bigl[k_p^\top\overline{\Sigma}_yk_p\bigr]}{\E_p\bigl[k_p^\top\bar c_p\bigr]}+o(\alpha^{-1})
\qquad (\alpha\to\infty).
\]
Positive alignment between $k_p$ and $\bar c_p$ alone therefore guarantees $0<\eta(\alpha)<2\rho(\alpha)$---and hence strict contrast MSE reduction below the gene-mean baseline in Equation~\eqref{eq:mse-split}---for all sufficiently large finite penalties without requiring small-$\alpha$ overshoot; if in addition $\eta>\rho$ at some smaller penalty $\alpha_0$ with $\E_p[k_p^\top M_\alpha\bar c_p]>0$ for $\alpha\ge\alpha_0$, continuity yields a finite crossing penalty where $\eta=\rho$ (without claiming uniqueness). In practice, RELAY selects $\alpha$ on development Macro-F1 rather than setting $\alpha=\alpha^\ast$ (Table~C.2), yielding a test ratio $\eta/\rho=1.09$ (Table~5) just on the under-regularized side of the crossing.
\end{remark}

\subsection{Readouts for DE and DIR Prediction}\label{sec:method-readout}\label{sec:method-geometry}
\paragraph{Direction.}
Because directly thresholding transferred responses ignores strong gene-level priors and degrades global decision accuracy (Section~\ref{sec:response-learning}), task-specific readouts combine each gene's baseline tendency with the transferred contrast to produce calibrated binary decisions. Because DIR depends only on the sign of the predicted response on DE-positive pairs, its readout adds the support gene mean $\mu(g)$ back to the transferred signed contrast $\widehat d(p,g)$ and normalizes by the gene's out-of-fold prediction-error scale $\tau(g)>0$ as
\begin{equation}\label{eq:pf-task-scores}
 s_{\mathrm{DIR}}(p,g)=\tanh\!\left(\frac{\widehat d(p,g)+\mu(g)}{\tau(g)}\right),
\end{equation}
predicting upregulation when $s_{\mathrm{DIR}}(p,g)>t_{\mathrm{DIR}}$ for a single calibrated threshold $t_{\mathrm{DIR}}$. Adding $\mu(g)$ anchors each gene on its typical direction across perturbations, so queries with weak kernel support ($\widehat d(p,g)\approx 0$) revert smoothly toward the gene's baseline directional tendency. Within a gene, $\tau(g)>0$, the shift $\mu(g)$, and the strictly increasing $\tanh$ make $s_{\mathrm{DIR}}(p,g)$ a strictly increasing function of $\hat d(p,g)$, so Equation~\eqref{eq:pf-task-scores} preserves within-gene ranking and changes only where a shared threshold cuts each gene.

Across genes, the standardized score $(\hat d+\mu)/\tau$ admits a direct location-scale interpretation: under $r(p,g)\mid\hat d(p,g)=\mu(g)+\hat d(p,g)+\tau(g)\varepsilon_0$ for any continuous zero-mean symmetric residual $\varepsilon_0$ supported on $\mathbb R$ with CDF $F_0$ (including Gaussian $F_0=\Phi$, or logistic where $2F_0(u)-1=\tanh(u)$ matches Equation~\eqref{eq:pf-task-scores}),
\[
\Pr\bigl(r(p,g)>0\,\big|\,\hat d(p,g)\bigr)=F_0\!\left(\frac{\hat d(p,g)+\mu(g)}{\tau(g)}\right),
\]
so a shared threshold on $s_{\mathrm{DIR}}$ cuts this directional probability at a uniform level across all genes. Because DIR is evaluated conditional on DE ($|r|>c$ for fixed $c\ge 0$), conditioning further yields
\[
\Pr\bigl(r>0\,\big|\,|r|>c,\hat d\bigr)=\frac{\bar F_0\bigl((c-\hat d-\mu)/\tau\bigr)}{\bar F_0\bigl((c-\hat d-\mu)/\tau\bigr)+F_0\bigl((-c-\hat d-\mu)/\tau\bigr)},\qquad \bar F_0=1-F_0,
\]
which remains strictly increasing in $\hat d$ within every gene for any such $F_0$ while depending on both $c/\tau(g)$ and $(\hat d+\mu)/\tau$. Equation~\eqref{eq:pf-task-scores} thus serves as a tractable cross-gene calibration score rather than an exact Bayes posterior, and we calibrate $t_{\mathrm{DIR}}$ empirically (ablations in Table~10).

\paragraph{Differential expression.}
Unlike DIR, calling DE requires evaluating whether the local effect magnitude exceeds background variation relative to the perturbation's transcriptome-wide impact, the gene's typical responsiveness, and measurement uncertainty. The DE readout therefore combines a learned base logit $L_{\mathrm{base}}(p,g)$ with a standardized-effect increment. To contextualize the local magnitude contrast $\widehat a(p,g)$ within global response patterns, we apply a joint PCA to the predicted signed and magnitude contrasts (each scaled by its prediction error and $\tanh$-transformed). This projection yields perturbation coordinates $u(p)$ in a shared component space alongside gene loadings $v_a(g)$ on the magnitude components. A lightweight MLP $f$ then maps the local contrast, the gene's baseline magnitude $\nu(g)$, and these global coordinates to the base logit via
\begin{equation}\label{eq:pf-de-features}
 h(p,g)=\bigl[u(p),\ \widehat a(p,g),\ \nu(g),\ v_a(g)\bigr],\qquad L_{\mathrm{base}}(p,g)=f\bigl(h(p,g)\bigr).
\end{equation}
Because expression contrasts alone do not reflect the statistical precision of each measurement, a parallel kernel ridge regression transfers the standardized effect $E(p,g)=\log(1+|F(p,g)|/\mathrm{SE}(p,g))$ from the training summary statistics using the same kernel weights $w(p)$. Standardizing the resulting prediction per gene yields $z(p,g)$, which shifts the base logit according to
\begin{equation}\label{eq:pf-de-amplitude}
 s_{\mathrm{DE}}(p,g)=\tanh\!\left(\frac{L_{\mathrm{base}}(p,g)+\gamma z(p,g)}{2}\right),
\end{equation}
where $\gamma$ controls the increment strength and DE is called when $s_{\mathrm{DE}}(p,g)>t_{\mathrm{DE}}$.

\paragraph{Fitting and calibration.}\label{sec:method-fitting}
Because the readouts must interpret responses transferred to unseen perturbations, fitting them on in-sample ridge predictions would underestimate transfer error and distort the scale parameters. We therefore fit all readout transformations on out-of-fold (OOF) development predictions. Specifically, we partition the training perturbations into disjoint groups and predict each held-out group from ridge regressions fitted on the remaining support, rebuilding the vocabulary, IDF weights, and support means $\mu,\nu$ within each fit (pseudocode in Algorithm~\ref{alg:perturbfactor} and settings in Table~\ref{tab:pf-training}, \appref{app:pf-implementation}). These OOF predictions determine the per-gene error scales $\tau(g)$, the joint PCA basis, the neural readout $f$ (trained with gene- and class-balanced binary cross-entropy), and the base increment slope $\beta$ in $\gamma=\lambda\beta$ (fitted with support-weighted binary cross-entropy). Calibration on pooled OOF development predictions directly reflects our dual evaluation objectives: we select the ridge penalty $\alpha$ and decision thresholds $t_{\mathrm{DE}},t_{\mathrm{DIR}}$ for global accuracy (Macro-F1), and calibrate $\lambda\in[0,1]$ to balance \esw{} and Macro-F1 (raising test DE \esw{} by $3.46$ points at a $1.04$-point cost in DE Macro-F1; Section~\ref{sec:main-hypothesis}). At test time, the upstream ridge regressions are refitted on all training perturbations $\mathcal P_{\mathrm{tr}}$, and scoring an unseen perturbation requires only its knowledge representation $\phi(p)$.

\subsection{Held-Out Cell Lines and Unseen Compounds}\label{sec:method-beyond}\label{sec:method-compounds}
\paragraph{Cross-cell-line generalization.}\label{sec:method-transfer}
In the held-out cell-line setting (protocol details in \appref{app:pf-transfer}), a target cell line $c^*$ contributes only unperturbed control cells, while $C$ source cell lines provide full training datasets. To relate the target cell line to the sources without perturbation data, we represent each cell line by its batch-averaged unperturbed control profile. Projecting these control profiles into a shared low-dimensional space via singular value decomposition (SVD) fitted on the source controls yields cell-line coordinates $b_c$. The cosine similarity between each source $c$ and the target $c^*$ in this control space defines normalized transfer weights
\begin{equation}\label{eq:cell-transfer-weights}
 \omega_c = \frac{1+\cos(b_{c^*},b_c)}{\sum_{c'=1}^C\bigl(1+\cos(b_{c^*},b_{c'})\bigr)}.
\end{equation}
For a query perturbation $p$, each source cell line $c$ independently predicts signed contrasts $\widehat d_c(p,g)$, magnitude contrasts $\widehat a_c(p,g)$, and standardized effects $\widehat E_c(p,g)$ from its own training support around its gene means $\mu_c(g),\nu_c(g)$. We then aggregate the source gene means, predicted contrasts, and standardized-effect predictions across source cell lines in response units using $\omega_c$ (renormalized over the sources that measure gene $g$), before applying error scaling and joint PCA to the combined profiles.

Because no perturbation responses are observed in $c^*$, we fit the readout transformations and calibrate thresholds via leave-one-cell-line-out cross-fitting across the sources. Each source cell line serves in turn as a pseudo-target cell line predicted from the remaining sources using Equation~\eqref{eq:cell-transfer-weights}, with every held-out perturbation excluded from all contributing source regressions so that the readouts learn to interpret cross-cell-line transferred profiles.

\paragraph{Target-pooled knowledge kernel for unseen compounds.}
Unlike CRISPRi knockdowns that target a single gene, small-molecule compounds typically act through one or more target proteins and lack per-gene perturbation identifiers (\appref{app:pf-transfer}). For each training compound, we construct its continuous response profile $r(p,g)$ by averaging log-normalized expression across treated cells and doses and subtracting plate-matched vehicle (DMSO) controls. We then decompose $r(p,g)$ into support gene means $\mu(g),\nu(g)$ and compound contrasts $d(p,g),a(p,g)$ via Equation~\eqref{eq:pf-decomposition}.

To construct a biological knowledge kernel over compounds without a language model or molecular-structure encoder, we bridge each compound $p$ to the gene-level knowledge graphs through its annotated protein targets, taking the union of drug-target proteins recorded in Tahoe-100M \citep{zhang2025tahoe} and PrimeKG drug--protein relations \citep{chandak2023primekg}. We pool the database-prefixed knowledge tokens of all target proteins of compound $p$ across GO, Reactome, CORUM, STRING, and BioPlex into a single token set, and apply the same support-computed IDF weighting, within-source $L_2$ normalization, and cross-source normalization as in Section~\ref{sec:method-encoding}. Through Equation~\eqref{eq:abl-kernel}, the inner product $k(p,p')=\phi(p)^\top\phi(p')$ defines a compound-level knowledge kernel that measures how strongly two compounds share target pathways, protein complexes, and interaction neighborhoods, while compounds without annotated targets receive $k_p=0$ and fall back to the training gene means. Kernel ridge regression over this compound kernel transfers signed and magnitude contrasts from all training compounds to unseen compounds, which are then scored by the analytic DIR readout and the PCA-based DE readout $L_{\mathrm{base}}(p,g)$ (omitting the standardized-effect increment when matched standard-error tables are unavailable).

% !TEX root = ../main.tex
\section{Experiments}\label{sec:experiments}

We evaluate whether RELAY improves both global accuracy and within-gene discrimination across PerturbQA cell lines (Section~\ref{sec:main-response-results}), analyze the contributions and trade-offs of its components (Section~\ref{sec:main-hypothesis}), and test whether its gains extend to a held-out cell line and unseen chemical compounds (Sections~\ref{sec:main-ood}--\ref{sec:main-drug}).

\subsection{Experimental Setup}\label{sec:experiment-setup}
On PerturbQA, we evaluate unseen genetic perturbations within each of the four cell lines---K562, HepG2, Jurkat, and RPE1---spanning 116,041 DE and 16,208 DIR test queries \citep{wu2025perturbqa} (per-cell-line split statistics in \appref{app:data-sources}). Using the evaluation metrics of Section~\ref{sec:evaluation}, all methods with query-level predictions are scored on identical test rows. Standalone DIR is evaluated on ground-truth DE-positive pairs, whereas joint three-class prediction calls non-DE when $s_{\mathrm{DE}}\le t_{\mathrm{DE}}$ and assigns up or down via $s_{\mathrm{DIR}}$ otherwise (Algorithm~\ref{alg:perturbfactor}). We quantify uncertainty using 2,000 paired bootstrap replicates \citep{efron1994bootstrap} that resample mixed-label genes for within-gene metrics and query perturbations for global metrics (protocols in \appref{app:main-uncertainty}).

Besides the training-only \emph{gene prior}---which predicts the positive class when a gene's positive training fraction satisfies $\pi(g)>1/2$---we compare RELAY against representative baselines from the three diagnosed families (reproduction settings in Table~\ref{tab:baseline-reproduction}, \appref{app:baseline-data}), namely \textbf{AROMA} \citep{wang2026aroma} for mechanistic reasoning, \textbf{SUMMER} \citep{wu2025perturbqa} and \textbf{CORE-Voting} \citep{yuan2026core} for label retrieval, and \textbf{GEARS} \citep{roohani2024gears}, \textbf{STATE} \citep{adduri2026state}, \textbf{PerturbDiff} \citep{yuan2026perturbdiff}, and \textbf{PertOmni} \citep{long2026pertomni}---all trained on the official cell-line splits without cross-dataset pretraining leakage---for continuous prediction. We also evaluate two controlled representation and transfer variants, \emph{RELAY with a GenePT kernel} \citep{chen2025genept} and \emph{RELAY with PertOmni kNN}, that replace only the perturbation representation or response-transfer operator while keeping RELAY's readouts fixed. Exact ties ($\pi(g)=1/2$) and unanswered baseline queries fall back to the majority label of the pooled cell-line training rate $\bar\pi$. For the four continuous baselines, binary decision thresholds are calibrated on held-out training perturbations (a 10\% validation split for GEARS, STATE, and PerturbDiff, and pooled four-fold out-of-fold development predictions for PertOmni) to maximize Macro-F1 (further thresholding details in \appref{app:baseline-details}). Because TxPert \citep{wenkel2026txpert} and the graph-neighbor system of \citet{fawkes2026kg} rely on proprietary perturbation-derived graphs, neither can be run directly on our public splits. We therefore evaluate public-graph analogues of direct neighbor response averaging \citep{fawkes2026kg} within our controlled operator ablations (Section~\ref{sec:main-hypothesis}).

% !TEX root = ../main.tex
\subsection{Main Results}\label{sec:main-response-results}\label{sec:main-utility}
Across the four PerturbQA cell lines, RELAY outperforms every evaluated baseline on both primary metrics in both DE and DIR (Table~\ref{tab:main-comparison}, with per-cell-line breakdowns and paired bootstrap intervals in \appref{app:main-cell-results}). All paired improvements are statistically significant, with 95\% bootstrap confidence intervals excluding zero. Specifically, RELAY surpasses the gene prior in global Macro-F1 by 1.27 percentage points on DE (0.8941 vs.\ 0.8814) and 7.49 points on DIR (0.8722 vs.\ 0.7973), while improving \esw{} over the strongest within-gene baseline (PertOmni) by 4.65 points on DE and 4.41 points on DIR (and over the strongest LLM baseline, SUMMER, by 7.48 and 6.53 points). Within-gene discrimination therefore need not come at the expense of global accuracy.

Across baseline families, Table~\ref{tab:main-comparison} directly reflects the failure modes diagnosed in Section~\ref{sec:diagnostics}. Mechanistic reasoning (AROMA) comes closest to the gene prior in global DE Macro-F1 among full-coverage baselines ($.8618$) yet remains near chance within genes ($.5169$ \esw{}). Conversely, label retrieval (SUMMER) achieves the strongest within-gene discrimination among LLM baselines at a $12.54$-point cost in DE Macro-F1 ($.7560$), and even reinforcement-learned multi-round voting (CORE-Voting, $\approx .8859$ DE Macro-F1 and $.7025$ \pga{}) remains $7.07$ \pga{} points below RELAY ($.7732$). Direct continuous expression predictors (GEARS, STATE, and PerturbDiff) fall below the gene prior in global Macro-F1 on both tasks due to gene-offset and amplitude errors (Section~\ref{sec:response-learning}), although STATE and PerturbDiff recover substantial within-gene ranking on DIR ($.6584$ and $.6788$ \pga{}). Notably, STATE nearly matches SUMMER in DIR \pga{} ($.6584$ vs.\ $.6637$) while trailing by $8.48$ points in \esw{} ($.6171$ vs.\ $.7019$), confirming on the full benchmark that real-valued scores can raise \pga{} by breaking ties even when binary decisions are miscalibrated (Section~\ref{sec:evaluation}). Similarity-based response transfer (PertOmni) substantially outperforms all direct neural models and LLM baselines within genes ($.6109$ DE and $.7231$ DIR \esw{}) and exceeds the gene prior on DIR Macro-F1 ($.8441$), confirming the benefit of reusing measured training responses. Yet directly thresholding its predicted response magnitude still leaves DE Macro-F1 $7.93$ points below the gene prior ($.8021$ vs.\ $.8814$), and RELAY outperforms PertOmni on every metric across both tasks ($+9.20$ DE and $+2.81$ DIR Macro-F1, $+4.65$ DE and $+4.41$ DIR \esw{}, and $+12.52$ DE and $+5.91$ DIR \pga{} points).

\begin{table}[H]
\centering\small
\caption{\textbf{Main comparison on PerturbQA.} Means weight cell lines equally, baselines are grouped by paradigm, and bold marks the best value per column. Unanswered LLM queries receive the pooled training-rate fallback, and AROMA gives a usable direction for 69.5\% of DIR queries (\appref{app:main-protocol}). $^*$CORE-Voting \citep{yuan2026core} is a label-retrieval method listed with its reported DE values (Table~\ref{tab:baseline-reproduction}).}
\label{tab:main-comparison}
\setlength{\tabcolsep}{3.5pt}
\begin{tabular}{@{}lrrrrrr@{}}
\toprule
 & \multicolumn{3}{c}{DE} & \multicolumn{3}{c}{DIR} \\
\cmidrule(lr){2-4}\cmidrule(l){5-7}
Method & Macro-F1 & WG-BAcc & pgAUROC & Macro-F1 & WG-BAcc & pgAUROC \\
\midrule
Gene prior & .8814 & .5000 & .5000 & .7973 & .5000 & .5000 \\
AROMA (reasoning) & .8618 & .5169 & .5152 & .7265 & .6029 & .5526 \\
SUMMER (retrieval) & .7560 & .5827 & .5891 & .8215 & .7019 & .6637 \\
GEARS (continuous) & .7202 & .5119 & .5143 & .7085 & .5109 & .5462 \\
STATE (continuous) & .6795 & .5559 & .5452 & .7414 & .6171 & .6584 \\
PerturbDiff (continuous) & .7370 & .5433 & .5336 & .7792 & .6445 & .6788 \\
PertOmni (continuous) & .8021 & .6109 & .6479 & .8441 & .7231 & .8020 \\
CORE-Voting (retrieval)$^*$ & $\approx .8859$ & --- & .7025 & --- & --- & --- \\
\midrule
\textbf{RELAY (ours)} & \textbf{.8941} & \textbf{.6574} & \textbf{.7732} & \textbf{.8722} & \textbf{.7672} & \textbf{.8611} \\
\bottomrule
\end{tabular}
\end{table}

Comparing the two tasks also explains why RELAY's global gain over the gene prior is much larger on DIR ($+7.49$ Macro-F1 and $+26.72$ \esw{} points) than on DE ($+1.27$ Macro-F1 and $+15.74$ \esw{} points). Whereas DE spans many consistently non-responsive genes that give the gene prior a $.8814$ floor, DIR is restricted to DE-positive pairs whose regulation flips across pathways and whose per-gene training labels are far sparser ($68{,}342$ vs.\ $498{,}438$ training pairs), allowing dense signed-contrast transfer to improve both global and within-gene accuracy simultaneously. A parallel gradient appears across individual cell lines (\appref{app:main-cell-results}). Global gains are largest where the gene prior is weakest (K562, $+4.32$ DE and $+15.43$ DIR Macro-F1, reaching $.9375$ DIR \pga{}), while within-gene gains remain substantial even where the prior is highest (RPE1, $+14.82$ DE and $+19.26$ DIR \esw{} points).

On DE, where the gene prior is already strong, RELAY achieves this discrimination by selectively revising only 5.09\% of the gene prior's decisions---correcting 2.730\% that the prior gets wrong while overturning 2.357\% that it gets right, for a net accuracy gain of 0.373 percentage points. When weighted by the within-gene support weights of \esw{}, corrections outweigh overturned decisions by 22.60\% to 6.86\%, yielding a 15.74-point \esw{} gain over chance (ranging from 12.62 to 19.37 points across cell lines, with all 95\% bootstrap intervals above zero). At the same time, on the 42,153 high-certainty DE queries whose gene has at least 20 identical training labels, RELAY retains 99.39\% accuracy compared with the prior's 99.57\% (versus 88.85\% for SUMMER in Section~\ref{sec:retrieval-transfer}).

\subsection{Ablation Study: Separating Evidence Gains from Method Design}\label{sec:main-hypothesis}\label{sec:main-readouts}
Because RELAY uses dense training responses and auxiliary $(F,\mathrm{SE})$ statistics alongside its knowledge kernel and task-specific readouts, a key question is how much of its improvement comes from richer training-side evidence versus RELAY's representation, transfer operator, and readout design. We disentangle these factors across the four PerturbQA cell lines in three stages, namely expanding support from sparse labels to dense continuous responses (Table~\ref{tab:buildup-ablation}a), substituting the perturbation representation and comparing kernel ridge regression against neighbor-averaging operators under matched inputs and readouts (Table~\ref{tab:buildup-ablation}c,d and Table~\ref{tab:encoding-geometry}), and building up RELAY's decomposition, task-specific readouts, and auxiliary $(F,\mathrm{SE})$ statistics under fixed dense responses (Table~\ref{tab:buildup-ablation}b and Tables~\ref{tab:component-fourarm}--\ref{tab:dir-term-ablation}).

\begin{table}[!ht]
\centering\small
\caption{\textbf{Separating training evidence from method design across RELAY's pipeline.} Means weight cell lines equally. Block~(a) evaluates direct scores without a learned readout (\appref{app:ablations}). Block~(b) builds up RELAY sequentially on the same dense training responses, comparing RELAY without $(F,\mathrm{SE})$ against the full model. Block~(c) substitutes GenePT embeddings for the knowledge representation or PertOmni's learned \texttt{kNN-text} neighbor average for RELAY's kernel ridge transfer under the identical readout pipeline. Block~(d) compares response-transfer operators, where the first three rows share the exact same knowledge kernel, training responses, $(F,\mathrm{SE})$, and retrained readouts (same-batch fit of Table~\ref{tab:component-fourarm}), whereas the last row averages measured responses of public STRING and GO neighbors directly.}
\label{tab:buildup-ablation}\label{tab:buildup-main}\label{tab:transfer-what-full}\label{tab:representation-substitution}\label{tab:response-operator}
\setlength{\tabcolsep}{1.4pt}
\begin{tabular}{@{}lrrrrrrrr@{}}
\toprule
 & \multicolumn{4}{c}{DE} & \multicolumn{3}{c}{DIR} & Three-class\\
\cmidrule(lr){2-5}\cmidrule(lr){6-8}\cmidrule(l){9-9}
Configuration & Macro-F1 & pgBAcc & \eswhead{} & pgAUROC & Macro-F1 & \eswhead{} & pgAUROC & Macro-F1\\
\midrule
\multicolumn{9}{@{}l}{\emph{(a) What is transferred (direct scores, no learned readout)}}\\
Gene prior & .8814 & .5000 & .5000 & .5000 & .7973 & .5000 & .5000 & ---\\
Labels ($L_g$) & .8987 & --- & .5598 & .6553 & .8510 & .6788 & .7622 & ---\\
Contrasts ($L_g$) & .8505 & --- & .5511 & .6282 & .8508 & .6603 & .7443 & ---\\
Binarized responses ($\mathcal P_{\mathrm{tr}}$) & .8148 & --- & .6354 & .6837 & .8454 & .7749 & .8221 & ---\\
Binarized, RELAY pen.\ ($\mathcal P_{\mathrm{tr}}$) & .8344 & --- & .6362 & .7005 & .8585 & .7757 & .8359 & ---\\
Contrasts $|\widehat d+\mu|$ ($\mathcal P_{\mathrm{tr}}$) & .8173 & --- & .6323 & .6687 & .8659 & .7669 & .8611 & ---\\
Magnitude $\widehat a+\nu$ ($\mathcal P_{\mathrm{tr}}$) & .8544 & --- & .6389 & .7086 & --- & --- & --- & ---\\
\midrule
\multicolumn{9}{@{}l}{\emph{(b) Component buildup and auxiliary statistics (dense responses $\mathcal P_{\mathrm{tr}}$)}}\\
Direct response readout & .8133 & .6188 & .6341 & .6670 & .8649 & .7709 & .8590 & .7427\\
$+$ learned DE readout & .8990 & .5988 & .6068 & .7370 & .8648 & .7703 & .8590 & .8077\\
$+$ centering on gene means & .9016 & .6083 & .6152 & .7465 & .8722 & .7672 & .8611 & .8148\\
$+$ magnitude (\textbf{w/o $F,\mathrm{SE}$}) & \textbf{.9045} & .6156 & .6228 & .7672 & \textbf{.8722} & \textbf{.7672} & \textbf{.8611} & \textbf{.8185}\\
$+$ std.\ effect (\textbf{Full RELAY}) & .8941 & \textbf{.6475} & \textbf{.6574} & \textbf{.7732} & \textbf{.8722} & \textbf{.7672} & \textbf{.8611} & .8115\\
\midrule
\multicolumn{9}{@{}l}{\emph{(c) Perturbation representation and transfer (matched readout pipeline)}}\\
GenePT in RELAY & .8847 & --- & .6106 & .7094 & .8292 & .7047 & .7735 & .7794\\
PertOmni kNN in RELAY & .8850 & --- & .6278 & .7275 & .8417 & .7241 & .8020 & .7845\\
\midrule
\multicolumn{9}{@{}l}{\emph{(d) Transfer operator (matched dense responses)}}\\
Kernel ridge reg.\ (RELAY) & \textbf{.8953} & --- & .6593 & \textbf{.7729} & \textbf{.8722} & \textbf{.7672} & .8611 & \textbf{.8131}\\
Weighted neighbor avg. & .8914 & --- & \textbf{.6599} & .7656 & .8637 & .7623 & \textbf{.8667} & .8043\\
Equal-weight neighbor avg. & .8899 & --- & .6508 & .7574 & .8477 & .7495 & .8534 & .7952\\
Public graph-neighbor avg. & .8215 & --- & .6264 & .6622 & .8500 & .7500 & .8392 & .7439\\
\bottomrule
\end{tabular}
\end{table}

\paragraph{Transferred quantity and support coverage.}
Block~(a) of Table~\ref{tab:buildup-ablation} compares transferring discrete labels $y$ or continuous contrasts $d=r-\mu$ restricted to each gene's labeled training perturbations $L_g$ against transferring binarized or continuous responses across all training perturbations $\mathcal P_{\mathrm{tr}}$ using direct scores that rank DE by $|\widehat r|$ and DIR by $\widehat r$ without a learned readout (protocol details in \appref{app:abl-what}). Because DIR labels are defined only on DE-positive pairs, a quarter of test genes have at most one or two labeled training perturbations in $L_g$. Expanding transfer from the sparse labeled subset $L_g$ to measured responses across all training perturbations $\mathcal P_{\mathrm{tr}}$ raises \esw{} by 8.81 points on DIR and 7.25 points on DE. Across $\mathcal P_{\mathrm{tr}}$, transferring binarized responses ($\mathrm{sign}(r)$ for DIR and $\mathbb{I}[|r|>t]$ for DE) achieves \esw{} comparable to continuous contrasts ($+0.80$ DIR and $+0.31$ DE points). However, binarization loses 3.91 DIR \pga{} points and 2.05 DIR Macro-F1 points because 61--76\% of training pairs have $|r|<0.05$ with weakly determined signs. Dense coverage across all training perturbations therefore drives the within-gene decision gain, while continuous contrast values preserve ranking resolution and global accuracy. Consistent with Proposition~1, transferring the sign-invariant magnitude view $\hat{a} + \nu$ across $\mathcal{P}_{\mathrm{tr}}$ further raises DE pgAUROC by 3.99 points over $|\hat{d} + \mu|$ (0.7086 vs.\ 0.6687) and DE Macro-F1 by 3.71 points by preventing opposite-signed pathway neighbors from cancelling.

\paragraph{Knowledge representation and transfer operator under matched evidence.}
Holding the training evidence and readout pipeline fixed isolates the contribution of RELAY's representation and transfer operator. First, replacing the multi-source knowledge representation $\phi(p)$ with $L_2$-normalized GenePT model-3 text embeddings \citep{chen2025genept} increases development-fold contrast MSE at the frozen penalty by 14.6\% (signed) and 12.5\% (magnitude, Table~\ref{tab:source-ablation}), lowering test \esw{} by 4.68 DE and 6.25 DIR points and three-class Macro-F1 by 3.21 points (Table~\ref{tab:buildup-ablation}c). Likewise, a second matched control (\emph{PertOmni kNN in RELAY}) keeps RELAY's centering, PCA context, DE readout, standardized-effect increment, and DIR readout fixed while substituting PertOmni's learned \texttt{kNN-text} neighbor-averaged responses (both signed and magnitude views, out-of-fold on development perturbations and on test perturbations) for RELAY's knowledge-kernel ridge regressions (\appref{app:operator-details}). Pairing PertOmni with RELAY's readouts raises its DE Macro-F1 from $.8021$ to $.8850$ and DE \esw{} from $.6109$ to $.6278$, confirming the readout's role in global calibration; yet full RELAY still significantly outperforms this matched control across every metric (DE: $+0.91$ $[+0.59,+1.25]$ Macro-F1, $+2.96$ $[+2.26,+3.71]$ \esw{}, $+4.57$ $[+3.46,+5.62]$ \pga{}; DIR: $+3.05$ $[+2.14,+4.01]$ Macro-F1, $+4.31$ $[+3.21,+5.36]$ \esw{}, $+5.91$ $[+4.02,+7.89]$ \pga{}; three-class Macro-F1 $+2.70$ points; 2,000 paired bootstrap replicates). Table~\ref{tab:encoding-geometry} explains why the knowledge kernel outperforms text-embedding representations. Across pairs of training perturbations, the knowledge kernel aligns much more strongly with empirical response-contrast correlation (Pearson alignment $0.172$ vs.\ $0.053$ for GenePT and $0.078$ for GEARS's GO Jaccard similarity) while maintaining low background similarity ($0.024$ vs.\ $0.427$ for GenePT) and selective effective support ($37.9\%$ vs.\ $97.5\%$). Under each representation's best grid penalty, sharpening GenePT similarities by their elementwise fourth power suppresses background similarity ($0.041$) and partially narrows the contrast-error gap ($+6.4\%$ vs.\ $+10.4\%$). Under Proposition~2 an isotropic background would cost nothing once the penalty is retuned, so the gap that remains at each representation's best penalty indicates that the background similarity of GenePT is not isotropic and that the fourth power also reweights the informative part of the kernel. By contrast, merging tokens across knowledge databases (\emph{pooled vocab.}) slightly degrades alignment and error ($+1.1\%$), and leave-one-source-out ablations (Appendix~D.3) confirm that all five curated databases contribute complementarily.

Second, Block~(d) of Table~\ref{tab:buildup-ablation} compares transfer operators given the exact same dense training responses. A direct public-graph neighbor average over STRING and GO \citep{fawkes2026kg} already accesses all training responses $\mathcal P_{\mathrm{tr}}$ but reaches only $.8215$ DE Macro-F1, $.6264$ DE \esw{}, and $.7439$ three-class Macro-F1---trailing RELAY without $(F,\mathrm{SE})$ by $8.30$ DE Macro-F1 and $7.46$ three-class Macro-F1 points. Even when we equip top-$k$ neighbor averaging with RELAY's exact knowledge kernel, $(F,\mathrm{SE})$ statistics, and retrained readouts, kernel ridge regression outperforms both equal-weight neighbor averaging ($+0.54$ DE, $+2.45$ DIR, and $+1.79$ three-class Macro-F1 points, along with $+0.85$ DE and $+1.77$ DIR \esw{} points) and similarity-weighted neighbor averaging ($+0.39$ DE, $+0.85$ DIR, and $+0.88$ three-class Macro-F1 points, along with $+0.73$ DE \pga{} and $+0.49$ DIR \esw{} points and comparable DE \esw{} at $.6593$ vs.\ $.6599$) by de-correlating redundant neighbors through $(K+\alpha I_n)^{-1}$, which keeps the selective signal under background similarity and lets the penalty $\alpha$ control the contrast amplitude (Proposition~2).

\begin{table}[!ht]
\centering\small
\caption{\textbf{Encoding geometry and contrast prediction.} Alignment denotes the Pearson correlation, over pairs of training perturbations, between kernel similarity and response-contrast similarity (mean across four cell lines). Background is the mean off-diagonal similarity. Kernel spread is the median effective support of the raw similarity vector $k_p$, $(\sum_{p'}k)^2/\sum_{p'}k^2$, as a share of the training support rather than the ridge regression weights. Error and $\rho$ are evaluated via anchored ridge regression on the development folds at the best penalty on the grid, where $\Delta$ is relative to the knowledge representation at its best penalty (4.789$\times10^{-3}$). GenePT uses the model-3 text embeddings of the substitution experiment, and the fourth power is applied elementwise.}
\label{tab:encoding-geometry}
\setlength{\tabcolsep}{2.5pt}
\begin{tabular}{@{}lrrrrr@{}}
\toprule
Representation & Alignment $\uparrow$ & Background & Kernel spread (\%) & Error $\Delta$ $\downarrow$ & $\rho$ $\uparrow$\\
\midrule
Knowledge representation (RELAY) & \textbf{.172} & .024 & 37.9 & --- & \textbf{.430}\\
Knowledge representation (pooled vocab.) & .153 & .027 & 37.3 & $+1.1\%$ & .420\\
\midrule
STRING only & .141 & .032 & 35.4 & $+3.6\%$ & .400\\
GO only & .132 & .036 & 47.4 & $+5.1\%$ & .388\\
GO Jaccard (GEARS similarity) & .078 & .137 & 86.3 & $+4.5\%$ & .387\\
\midrule
GenePT & .053 & .427 & 97.5 & $+10.4\%$ & .324\\
GenePT, fourth power & .102 & .041 & 69.2 & $+6.4\%$ & .370\\
\bottomrule
\end{tabular}
\end{table}

\paragraph{Task-specific readouts and the role of auxiliary $(F,\mathrm{SE})$ statistics.}
Finally, Block~(b) of Table~\ref{tab:buildup-ablation} and the matched component ablations in Tables~\ref{tab:component-fourarm}--\ref{tab:dir-term-ablation} isolate the contribution of RELAY's decomposition and readouts at fixed dense training responses, as well as the exact benefit and cost of the auxiliary statistics $(F,\mathrm{SE})$. Starting from direct thresholding of transferred responses ($.8133$ DE Macro-F1, $.6341$ DE \esw{}), introducing the learned PCA-contextualized DE readout raises global Macro-F1 by 8.57 points (to $.8990$), centering on gene means raises it to $.9016$, and adding the magnitude contrast view $a(p,g)$ yields \textbf{RELAY without $(F,\mathrm{SE})$}. Using only the expression matrix $R$ and knowledge kernel, RELAY without $(F,\mathrm{SE})$ already reaches $.9045$ DE Macro-F1 ($+2.31$ points above the gene prior and $+4.27$ above AROMA), $.6228$ DE \esw{} ($+1.19$ points above PertOmni, $+4.01$ above SUMMER, and $+6.69$ above STATE), $.7672$ DE \pga{} ($+6.47$ points above CORE-Voting and $+11.93$ above PertOmni), and $.8185$ three-class Macro-F1, while achieving the full model's exact DIR performance ($.8722$ Macro-F1, $.7672$ \esw{}, $.8611$ \pga{}, as DIR never uses $F$ or $\mathrm{SE}$). Thus, RELAY's joint advantage on both metrics does not depend on $(F,\mathrm{SE})$.

Adding the standardized-effect increment in \textbf{Full RELAY} introduces a targeted trade-off on DE. By incorporating per-pair measurement precision, it boosts within-gene DE discrimination by $+3.46$ \esw{} points (to $.6574$), $+3.19$ pgBAcc points (to $.6475$), and $+0.60$ \pga{} points (to $.7732$), at a modest cost of $1.04$ global DE Macro-F1 points (from $.9045$ to $.8941$, still $+1.27$ points above the gene prior). Under matched retraining (Table~\ref{tab:component-fourarm}), zeroing the PCA context ($u(p)$ and $v_a(g)$) lowers DE Macro-F1 by 3.74 points and \esw{} by 2.96 points, and removing contrasts altogether (\emph{Mean training effect}) collapses within-gene discrimination to chance. On DIR (Table~\ref{tab:dir-term-ablation}), adding the gene mean $\mu$ to the transferred contrast $\widehat d$ leaves within-gene rankings unchanged ($\text{pgAUROC} = .8611$) while raising global Macro-F1 by 10.72 points (from $.7650$ to $.8722$, with \esw{} shifting from $.7929$ to $.7672$), because $\mu$ anchors weakly transferred queries to each gene's baseline directional tendency. Normalizing by $\tau(g)$ under $\tanh$ further sharpens global calibration.

\begin{table}[!ht]
\centering\small
\caption{\textbf{DE component ablation under matched retraining} (cell lines weighted equally). The first column gives the same-batch full model, and the other columns give paired differences from it in percentage points (\emph{Joint simplification} removes mean centering, the magnitude view, and the $(F,\mathrm{SE})$ increment). Brackets are 95\% perturbation-cluster bootstrap intervals (2,000 replicates, seed 17, resampling shared across cell lines).}
\label{tab:component-fourarm}\label{tab:component-fourarm-delta}
\setlength{\tabcolsep}{3pt}
\begin{tabular}{@{}lrrrr@{}}
\toprule
Metric & \begin{tabular}[b]{@{}c@{}}Full\\(same batch)\end{tabular} & \begin{tabular}[b]{@{}c@{}}$-$ PCA\\context\end{tabular} & \begin{tabular}[b]{@{}c@{}}Joint\\simplification\end{tabular} & \begin{tabular}[b]{@{}c@{}}Mean training\\effect\end{tabular}\\
\midrule
DE Macro-F1 & .8953 & $-3.74$ [$-4.11,-3.38$] & $+0.19$ [$-0.11,+0.49$] & $-5.89$ [$-6.40,-5.38$]\\
DE pgAUROC & .7729 & $-5.67$ [$-6.47,-4.34$] & $-2.39$ [$-3.60,-1.34$] & $-27.29$ [$-29.24,-26.43$]\\
DE \esw{} & .6593 & $-2.96$ [$-3.65,-2.21$] & $-3.88$ [$-4.68,-3.21$] & $-15.94$ [$-17.16,-14.85$]\\
\bottomrule
\end{tabular}
\end{table}

\begin{table}[!ht]
\centering\small
\caption{\textbf{DIR readout terms} (cell lines weighted equally). $\mu$ is constant within a gene, so its within-gene metrics equal $0.5$ by construction. Dashes mark metrics not recorded for these variants.}
\label{tab:dir-term-ablation}
\setlength{\tabcolsep}{5pt}
\begin{tabular}{@{}lrrrr@{}}
\toprule
Readout terms & Macro-F1 & pgBAcc & \esw{} & pgAUROC\\
\midrule
$\mu$ only (gene mean) & .7645 & .5000 & .5000 & .5000\\
$\widehat d$ only (query contrast) & .7650 & --- & \textbf{.7929} & .8611\\
$\widehat d+\mu$ without $\tau$ & .8706 & --- & .7680 & .8611\\
$\tanh((\widehat d+\mu)/\tau)$ (full DIR readout) & \textbf{.8722} & .7159 & .7672 & .8611\\
\bottomrule
\end{tabular}
\end{table}

\subsection{Generalization to a Held-Out Cell Line}\label{sec:main-ood}\label{sec:main-rpe1-transfer}
To evaluate zero-shot transfer to a cellular context with no observed perturbation responses, we transfer from K562, HepG2, and Jurkat to held-out RPE1, which contributes only unperturbed control cells (Section~\ref{sec:method-beyond}). Following AROMA's three-class evaluation protocol \citep{wang2026aroma}, we compare RELAY against AROMA and a \emph{source gene prior} that assigns each gene its majority training label pooled across the three source cell lines (with the pooled source training rate as tiebreaker and fallback, Table~\ref{tab:ood}). RELAY achieves the highest three-class Macro-F1 ($.7597$ vs.\ $.73$ for AROMA and $.6843$ for the source gene prior). This gain is driven by substantial improvements on the upregulated ($.6243$ vs.\ $.57$) and downregulated ($.6980$ vs.\ $.66$) classes, while maintaining strong within-gene discrimination on both binary tasks ($.5933$ DE and $.7333$ DIR \esw{}). Because 1,511 of the 37,942 RPE1 test queries involve perturbation--gene pairs that also appear in at least one source cell line's DE training label split (\appref{app:historical-exposure}), we further evaluate on the 36,431 label-non-overlapping queries (\appref{app:rpe1-transfer-results}). On this subset, RELAY remains robust ($.8191$ DE and $.8264$ DIR Macro-F1, along with $.5807$ DE and $.6576$ DIR \esw{}) and continues to outperform an exact-pair source prior in three-class Macro-F1 ($.7176$ vs.\ $.6552$).

\begin{table}[!ht]
\centering\small
\caption{\textbf{Out-of-distribution transfer to held-out RPE1.} Three-class F1 (ND, non-DE) and \esw{} of the binary tasks, where AROMA reports only three-class values \citep[Table~2]{wang2026aroma}.}
\label{tab:ood}
\setlength{\tabcolsep}{4.5pt}
\begin{tabular}{@{}lrrrrrr@{}}
\toprule
 & \multicolumn{4}{c}{Three-class} & \multicolumn{2}{c}{\esw{}} \\
\cmidrule(lr){2-5}\cmidrule(l){6-7}
Method & ND & Up & Down & Macro-F1 & DE & DIR \\
\midrule
Source gene prior & .9590 & .4845 & .6094 & .6843 & .5000 & .5000 \\
AROMA (reported) & .96 & .57 & .66 & .73 & --- & --- \\
\textbf{RELAY (ours)} & .9566 & \textbf{.6243} & \textbf{.6980} & \textbf{.7597} & \textbf{.5933} & \textbf{.7333} \\
\bottomrule
\end{tabular}
\end{table}

\subsection{Generalization to Unseen Compounds}\label{sec:main-drug}
To evaluate whether RELAY extends beyond genetic knockdowns to small-molecule perturbations, we test on the five Tahoe-100M cell lines of GeneTAK \citep{wei2026vcworld,zhang2025tahoe}. Following PerturbQA's 75/25 perturbation-disjoint ratio, we split the 379 normalized drugs shared across all five cell lines into 284 training drugs and 95 held-out test drugs. Across the five cell lines, this split yields 166,202 DE and 85,832 DIR held-out test queries (split counts and protocols in \appref{app:genetak-data}). Besides the within-cell-line gene prior, we compare RELAY with the LLM reasoning model \textbf{VCWorld} \citep{wei2026vcworld} (evaluated with Gemma4-26B-A4B-it using retrieved training-drug labels and the released prompt templates) and the continuous predictors \textbf{STATE} and \textbf{PertOmni} \citep{long2026pertomni}, which are retrained from scratch on the 284 training drugs using Morgan fingerprint representations (Table~\ref{tab:drug-comparison}, with protocol details and per-cell-line results in \appref{app:baseline-data} and \appref{app:genetak-results}).

\begin{table}[!ht]
\centering\small
\caption{\textbf{Unseen compounds on GeneTAK.} Means across cell lines, weighted equally, over 95 held-out drugs (results by cell line in \appref{app:genetak-results}). Unanswered VCWorld queries receive the pooled training-rate fallback.}
\label{tab:drug-comparison}
\setlength{\tabcolsep}{4.5pt}
\begin{tabular}{@{}lrrrrrr@{}}
\toprule
 & \multicolumn{3}{c}{DE} & \multicolumn{3}{c}{DIR} \\
\cmidrule(lr){2-4}\cmidrule(l){5-7}
Method & Macro-F1 & WG-BAcc & pgAUROC & Macro-F1 & WG-BAcc & pgAUROC \\
\midrule
Gene prior & \textbf{.9196} & .5000 & .5000 & \textbf{.7894} & .5000 & .5000 \\
STATE & .8298 & .5113 & .5254 & .6475 & .5337 & .5561 \\
PertOmni & .8671 & .5013 & .5422 & .7790 & .5049 & .5335 \\
VCWorld & .6988 & .5282 & .5297 & .5949 & .5179 & .5140 \\
\textbf{RELAY (ours)} & .9179 & \textbf{.5463} & \textbf{.6138} & .7828 & \textbf{.5574} & \textbf{.6284} \\
\bottomrule
\end{tabular}
\end{table}

Using the target-pooled knowledge kernel (Section~\ref{sec:method-beyond}), RELAY nearly matches the gene prior's global Macro-F1 on both tasks ($.9179$ vs.\ $.9196$ on DE and $.7828$ vs.\ $.7894$ on DIR) while substantially outperforming STATE, PertOmni, and VCWorld in global accuracy and achieving the strongest within-gene discrimination across all metrics (whereas PertOmni retains higher global Macro-F1 than STATE and VCWorld, its within-gene balanced accuracy remains near chance at $.5013$ on DE and $.5049$ on DIR). Its within-gene gains over the gene prior have 95\% bootstrap confidence intervals entirely above zero on both tasks ($+4.63$ DE and $+5.74$ DIR \esw{} points, $+5.71$ DE $[1.33, 9.07]$ and $+6.37$ DIR $[2.74, 9.66]$ unweighted pgBAcc points, and $+11.38$ DE and $+12.84$ DIR \pga{} points). These within-gene gains are smaller than on CRISPRi knockdowns under two constraints documented in \appref{app:pf-transfer}, although we do not quantify their respective shares. First, compounds without an annotated target protein in Tahoe-100M or PrimeKG have $\phi(p)=0$, so every query on them receives the training gene means and contributes no within-gene signal. Second, GeneTAK provides no per-pair effect standard errors $(F,\mathrm{SE})$, so the standardized-effect increment is switched off ($\gamma=0$). Even under incomplete target coverage and without auxiliary standard-error statistics, transferring target-pooled response contrasts around training gene means consistently improves within-gene discrimination without sacrificing global calibration as competing baselines do.

% !TEX root = ../main.tex

% !TEX root = ../main.tex
\section{Conclusion}\label{sec:conclusion}\label{sec:limitations}
Predicting cellular perturbation effects requires both the global accuracy of a gene prior and the ability to distinguish perturbations acting on the same target gene. None of the evaluated baselines meets both objectives on both tasks. Our diagnosis traces the within-gene signal to measured outcomes, yet discrete benchmark labels cover only a small fraction of measured perturbation--gene pairs, whereas direct continuous predictors misestimate gene offsets or contrast amplitudes and even similarity-based response transfer leaves DE globally miscalibrated without task-specific readouts. RELAY therefore transfers dense measured response contrasts through a multi-source knowledge kernel and anchors its decisions on gene-level tendencies. On average across the four PerturbQA cell lines, RELAY outperforms every evaluated baseline on both global and within-gene metrics, and it maintains a consistent advantage over competing predictors when transferred to a held-out cell line and to unseen chemical compounds.

Two limitations point to directions for future work. First, RELAY relates perturbations solely through curated functional annotations and molecular interactions, so unannotated genes or compounds without known targets receive no contrast transfer and fall back to the training gene means. Incorporating sequence-based protein representations or chemical-structure kernels could extend contrast transfer to poorly characterized perturbations. Second, although our ablations show that RELAY without $(F,\mathrm{SE})$ already surpasses all baselines and that its gains over continuous predictors hold under identical dense training responses, its comparisons with label-only language-model baselines remain system-level (Table~\ref{tab:supervision-sources}). Supplying dense response summaries or effect statistics directly to retrieval-augmented language models would help isolate architectural trade-offs from differences in training supervision.

% !TEX root = ../main.tex
\section*{Code Availability}
Inference code, reproduction workflows, and the evaluation suite (including \esw{}, paired bootstrap confidence intervals, and continuous profile error decomposition) are available at \url{https://github.com/Cyto-System/Cyto.Perturbation}. Dataset construction, supervision details, fitting procedures, and baseline evaluation protocols are documented in \appref{app:data-sources}, Algorithm~\ref{alg:perturbfactor}, Table~\ref{tab:pf-training}, and \appref{app:baseline-data}.

\clearpage
% !TEX root = ../main.tex
\bibliographystyle{preprint_style}
\bibliography{references}

\begin{thebibliography}{43}
\providecommand{\natexlab}[1]{#1}
\providecommand{\url}[1]{\texttt{#1}}
\expandafter\ifx\csname urlstyle\endcsname\relax
  \providecommand{\doi}[1]{doi: #1}\else
  \providecommand{\doi}{doi: \begingroup \urlstyle{rm}\Url}\fi

\bibitem[Adamson et~al.(2016)Adamson, Norman, Jost, et~al.]{adamson2016perturbseq}
Britt Adamson, Thomas~M. Norman, Marco Jost, et~al.
\newblock {A multiplexed single-cell {CRISPR} screening platform enables systematic dissection of the unfolded protein response}.
\newblock \emph{Cell}, 167\penalty0 (7):\penalty0 1867--1882.e21, 2016.
\newblock \doi{10.1016/j.cell.2016.11.048}.

\bibitem[Adduri et~al.(2026)Adduri, Gautam, Bevilacqua, et~al.]{adduri2026state}
Abhinav~K. Adduri, Dhruv Gautam, Beatrice Bevilacqua, et~al.
\newblock {Predicting cellular responses to perturbation across diverse contexts with {State}}.
\newblock \emph{Cell}, 189\penalty0 (19):\penalty0 5914--5931.e20, 2026.
\newblock \doi{10.1016/j.cell.2026.07.052}.

\bibitem[Ahlmann-Eltze et~al.(2025)Ahlmann-Eltze, Huber, and Anders]{ahlmann2025linear}
Constantin Ahlmann-Eltze, Wolfgang Huber, and Simon Anders.
\newblock {Deep-learning-based gene perturbation effect prediction does not yet outperform simple linear baselines}.
\newblock \emph{Nature Methods}, 22:\penalty0 1657--1661, 2025.
\newblock \doi{10.1038/s41592-025-02772-6}.

\bibitem[Aleksander et~al.(2023)Aleksander, Balhoff, Carbon, et~al.]{aleksander2023go}
Suzi~A. Aleksander, James Balhoff, Seth Carbon, et~al.
\newblock {The {Gene Ontology} knowledgebase in 2023}.
\newblock \emph{Genetics}, 224\penalty0 (1):\penalty0 iyad031, 2023.
\newblock \doi{10.1093/genetics/iyad031}.
\newblock The Gene Ontology Consortium.

\bibitem[Ashburner et~al.(2000)Ashburner, Ball, Blake, et~al.]{ashburner2000go}
Michael Ashburner, Catherine~A. Ball, Judith~A. Blake, et~al.
\newblock {{Gene Ontology}: tool for the unification of biology}.
\newblock \emph{Nature Genetics}, 25\penalty0 (1):\penalty0 25--29, 2000.
\newblock \doi{10.1038/75556}.

\bibitem[Brodersen et~al.(2010)Brodersen, Ong, Stephan, and Buhmann]{brodersen2010balanced}
Kay~Henning Brodersen, Cheng~Soon Ong, Klaas~Enno Stephan, and Joachim~M. Buhmann.
\newblock {The balanced accuracy and its posterior distribution}.
\newblock In \emph{Proceedings of the 20th International Conference on Pattern Recognition}, pp.\  3121--3124, 2010.
\newblock \doi{10.1109/ICPR.2010.764}.

\bibitem[Chandak et~al.(2023)Chandak, Huang, and Zitnik]{chandak2023primekg}
Payal Chandak, Kexin Huang, and Marinka Zitnik.
\newblock {Building a knowledge graph to enable precision medicine}.
\newblock \emph{Scientific Data}, 10:\penalty0 67, 2023.
\newblock \doi{10.1038/s41597-023-01960-3}.

\bibitem[Chen \& Zou(2025)Chen and Zou]{chen2025genept}
Yiqun~T. Chen and James Zou.
\newblock {Simple and effective embedding model for single-cell biology built from {ChatGPT}}.
\newblock \emph{Nature Biomedical Engineering}, 9:\penalty0 483--493, 2025.
\newblock \doi{10.1038/s41551-024-01284-6}.

\bibitem[Dixit et~al.(2016)Dixit, Parnas, Li, et~al.]{dixit2016perturbseq}
Atray Dixit, Oren Parnas, Biyu Li, et~al.
\newblock {Perturb-seq: dissecting molecular circuits with scalable single-cell {RNA} profiling of pooled genetic screens}.
\newblock \emph{Cell}, 167\penalty0 (7):\penalty0 1853--1866.e17, 2016.
\newblock \doi{10.1016/j.cell.2016.11.038}.

\bibitem[Efron \& Tibshirani(1994)Efron and Tibshirani]{efron1994bootstrap}
Bradley Efron and Robert~J. Tibshirani.
\newblock \emph{{An Introduction to the Bootstrap}}.
\newblock Chapman and Hall/CRC, 1994.
\newblock \doi{10.1201/9780429246593}.

\bibitem[Fawcett(2006)]{fawcett2006roc}
Tom Fawcett.
\newblock {An introduction to {ROC} analysis}.
\newblock \emph{Pattern Recognition Letters}, 27\penalty0 (8):\penalty0 861--874, 2006.
\newblock \doi{10.1016/j.patrec.2005.10.010}.

\bibitem[Fawkes et~al.(2026)Fawkes, Hodgson, and Hartford]{fawkes2026kg}
Jake Fawkes, Liam Hodgson, and Jason Hartford.
\newblock {Knowledge graphs and reasoning {LLM}s for finding simple yet effective transcriptomic perturbation predictors}.
\newblock \emph{arXiv preprint arXiv:2606.08816}, 2026.
\newblock URL \url{https://arxiv.org/abs/2606.08816v1}.

\bibitem[{Gemma Team}(2026)]{gemma2026gemma4}
{Gemma Team}.
\newblock {Gemma 4 technical report}.
\newblock \emph{arXiv preprint arXiv:2607.02770}, 2026.

\bibitem[Han et~al.(2018)Han, Cho, Lee, Yun, Kim, Bae, Yang, Kim, et~al.]{han2018trrust}
Heonjong Han, Jae-Won Cho, Sangyoung Lee, Ayoung Yun, Hyojin Kim, Dasom Bae, Sunmo Yang, Chan~Yeong Kim, et~al.
\newblock {{TRRUST v2}: an expanded reference database of human and mouse transcriptional regulatory interactions}.
\newblock \emph{Nucleic Acids Research}, 46\penalty0 (D1):\penalty0 D380--D386, 2018.
\newblock \doi{10.1093/nar/gkx1013}.

\bibitem[Hoerl \& Kennard(1970)Hoerl and Kennard]{hoerl1970ridge}
Arthur~E. Hoerl and Robert~W. Kennard.
\newblock {Ridge regression: biased estimation for nonorthogonal problems}.
\newblock \emph{Technometrics}, 12\penalty0 (1):\penalty0 55--67, 1970.
\newblock \doi{10.1080/00401706.1970.10488634}.

\bibitem[Huttlin et~al.(2021)Huttlin, Bruckner, Navarrete-Perea, et~al.]{huttlin2021bioplex}
Edward~L. Huttlin, Raphael~J. Bruckner, Jose Navarrete-Perea, et~al.
\newblock {Dual proteome-scale networks reveal cell-specific remodeling of the human interactome}.
\newblock \emph{Cell}, 184\penalty0 (11):\penalty0 3022--3040.e28, 2021.
\newblock \doi{10.1016/j.cell.2021.04.011}.

\bibitem[Kedzierska et~al.(2025)Kedzierska, Crawford, Amini, and Lu]{kedzierska2025zeroshot}
Kasia~Z. Kedzierska, Lorin Crawford, Ava~P. Amini, and Alex~X. Lu.
\newblock {Zero-shot evaluation reveals limitations of single-cell foundation models}.
\newblock \emph{Genome Biology}, 26:\penalty0 101, 2025.
\newblock \doi{10.1186/s13059-025-03574-x}.

\bibitem[Liu et~al.(2024)Liu, Lin, Hewitt, Paranjape, Bevilacqua, Petroni, and Liang]{liu2024lost}
Nelson~F. Liu, Kevin Lin, John Hewitt, Ashwin Paranjape, Michele Bevilacqua, Fabio Petroni, and Percy Liang.
\newblock {Lost in the middle: How language models use long contexts}.
\newblock \emph{Transactions of the Association for Computational Linguistics}, 12:\penalty0 157--173, 2024.
\newblock \doi{10.1162/tacl_a_00638}.

\bibitem[Long et~al.(2026)Long, Liu, Sza{\l}ata, Theis, Xue, and Zhao]{long2026pertomni}
Wenxin Long, Tianyu Liu, Artur Sza{\l}ata, Fabian~J. Theis, Lingzhou Xue, and Hongyu Zhao.
\newblock {Learning perturbation effects through contrastive alignment of multimodal biological embeddings}.
\newblock \emph{bioRxiv}, 2026.
\newblock \doi{10.64898/2026.06.23.734145}.

\bibitem[Lopez et~al.(2018)Lopez, Regier, Cole, Jordan, and Yosef]{lopez2018scvi}
Romain Lopez, Jeffrey Regier, Michael~B. Cole, Michael~I. Jordan, and Nir Yosef.
\newblock {Deep generative modeling for single-cell transcriptomics}.
\newblock \emph{Nature Methods}, 15\penalty0 (12):\penalty0 1053--1058, 2018.
\newblock \doi{10.1038/s41592-018-0229-2}.

\bibitem[Milacic et~al.(2024)Milacic, Beavers, Conley, et~al.]{milacic2024reactome}
Marija Milacic, Deidre Beavers, Patrick Conley, et~al.
\newblock {The {Reactome} pathway knowledgebase 2024}.
\newblock \emph{Nucleic Acids Research}, 52\penalty0 (D1):\penalty0 D672--D678, 2024.
\newblock \doi{10.1093/nar/gkad1025}.

\bibitem[M{\"u}ller-Dott et~al.(2023)M{\"u}ller-Dott, Tsirvouli, Vazquez, Ramirez~Flores, {Badia-i-Mompel}, Fallegger, T{\"u}rei, L{\ae}greid, and Saez-Rodriguez]{mullerdott2023collectri}
Sophia M{\"u}ller-Dott, Eirini Tsirvouli, Miguel Vazquez, Ricardo~O. Ramirez~Flores, Pau {Badia-i-Mompel}, Robin Fallegger, D{\'e}nes T{\"u}rei, Astrid L{\ae}greid, and Julio Saez-Rodriguez.
\newblock {Expanding the coverage of regulons from high-confidence prior knowledge for accurate estimation of transcription factor activities}.
\newblock \emph{Nucleic Acids Research}, 51\penalty0 (20):\penalty0 10934--10949, 2023.
\newblock \doi{10.1093/nar/gkad841}.

\bibitem[Nadig(2025)]{nadig2025statistics}
Ajay Nadig.
\newblock {Differential expression summary statistics for {Perturb-Seq} datasets analyzed in Nadig et al., 2025 ({Nature Genetics})}.
\newblock Figshare, 2025.
\newblock URL \url{https://doi.org/10.6084/m9.figshare.29498366.v1}.
\newblock Version 1.

\bibitem[Nadig et~al.(2025)Nadig, Replogle, Pogson, Murthy, McCarroll, Weissman, Robinson, and O'Connor]{nadig2025trade}
Ajay Nadig, Joseph~M. Replogle, Angela~N. Pogson, Mukundh Murthy, Steven~A. McCarroll, Jonathan~S. Weissman, Elise~B. Robinson, and Luke~J. O'Connor.
\newblock {Transcriptome-wide analysis of differential expression in perturbation atlases}.
\newblock \emph{Nature Genetics}, 57:\penalty0 1228--1237, 2025.
\newblock \doi{10.1038/s41588-025-02169-3}.

\bibitem[Phillips et~al.(2026)Phillips, Martell, Misra, Stoisser, Prada-Medina, Donovan-Maiye, and M{\"a}rtens]{phillips2026synthpert}
Lawrence Phillips, Marc~Boubnovski Martell, Aditya Misra, Josefa~Lia Stoisser, Cesar~A. Prada-Medina, Rory Donovan-Maiye, and Kaspar M{\"a}rtens.
\newblock {{SynthPert}: Enhancing {LLM} biological reasoning via synthetic reasoning traces for cellular perturbation prediction}.
\newblock In \emph{Machine Learning for Genomics Explorations Workshop at ICLR}, 2026.
\newblock URL \url{https://arxiv.org/abs/2509.25346v2}.
\newblock arXiv:2509.25346v2.

\bibitem[Replogle et~al.(2022)Replogle, Saunders, Pogson, Hussmann, Lenail, Guna, Mascibroda, Wagner, Adelman, Lithwick-Yanai, Iremadze, Oberstrass, Lipson, Bonnar, Jost, Norman, and Weissman]{replogle2022perturbseq}
Joseph~M. Replogle, Reuben~A. Saunders, Angela~N. Pogson, Jeffrey~A. Hussmann, Alexander Lenail, Alina Guna, Lauren Mascibroda, Eric~J. Wagner, Karen Adelman, Gila Lithwick-Yanai, Nika Iremadze, Florian Oberstrass, Doron Lipson, Jessica~L. Bonnar, Marco Jost, Thomas~M. Norman, and Jonathan~S. Weissman.
\newblock {Mapping information-rich genotype-phenotype landscapes with genome-scale Perturb-seq}.
\newblock \emph{Cell}, 185\penalty0 (14):\penalty0 2559--2575.e28, 2022.
\newblock \doi{10.1016/j.cell.2022.05.013}.

\bibitem[Roohani et~al.(2024)Roohani, Huang, and Leskovec]{roohani2024gears}
Yusuf Roohani, Kexin Huang, and Jure Leskovec.
\newblock {Predicting transcriptional outcomes of novel multigene perturbations with {GEARS}}.
\newblock \emph{Nature Biotechnology}, 42:\penalty0 927--935, 2024.
\newblock \doi{10.1038/s41587-023-01905-6}.
\newblock Published online 17 August 2023.

\bibitem[Saunders et~al.(1998)Saunders, Gammerman, and Vovk]{saunders1998krr}
Craig Saunders, Alexander Gammerman, and Volodya Vovk.
\newblock {Ridge regression learning algorithm in dual variables}.
\newblock In \emph{Proceedings of the Fifteenth International Conference on Machine Learning}, pp.\  515--521, 1998.

\bibitem[Shi et~al.(2023)Shi, Chen, Misra, Scales, Dohan, Chi, Sch{\"a}arli, and Zhou]{shi2023large}
Freda Shi, Xinyun Chen, Kanishka Misra, Nathan Scales, David Dohan, Ed~H. Chi, Nathanael Sch{\"a}arli, and Denny Zhou.
\newblock {Large language models can be easily distracted by irrelevant context}.
\newblock In \emph{Proceedings of the 40th International Conference on Machine Learning}, pp.\  31210--31227, 2023.

\bibitem[{Sp\"arck Jones}(1972)]{sparckjones1972idf}
Karen {Sp\"arck Jones}.
\newblock {A statistical interpretation of term specificity and its application in retrieval}.
\newblock \emph{Journal of Documentation}, 28\penalty0 (1):\penalty0 11--21, 1972.
\newblock \doi{10.1108/eb026526}.

\bibitem[Szklarczyk et~al.(2023)Szklarczyk, Kirsch, Koutrouli, et~al.]{szklarczyk2023string}
Damian Szklarczyk, Rebecca Kirsch, Mikaela Koutrouli, et~al.
\newblock {The {STRING} database in 2023: protein--protein association networks and functional enrichment analyses for any sequenced genome of interest}.
\newblock \emph{Nucleic Acids Research}, 51\penalty0 (D1):\penalty0 D638--D646, 2023.
\newblock \doi{10.1093/nar/gkac1000}.

\bibitem[Szklarczyk et~al.(2025)Szklarczyk, Nastou, Koutrouli, Kirsch, Mehryary, Hachilif, Hu, Peluso, Huang, Fang, Doncheva, Pyysalo, Bork, Jensen, and {von Mering}]{szklarczyk2025string}
Damian Szklarczyk, Katerina Nastou, Mikaela Koutrouli, Rebecca Kirsch, Farrokh Mehryary, Radja Hachilif, Dewei Hu, Matteo~E. Peluso, Qingyao Huang, Tao Fang, Nadezhda~T. Doncheva, Sampo Pyysalo, Peer Bork, Lars~J. Jensen, and Christian {von Mering}.
\newblock {The {STRING} database in 2025: protein networks with directionality of regulation}.
\newblock \emph{Nucleic Acids Research}, 53\penalty0 (D1):\penalty0 D730--D737, 2025.
\newblock \doi{10.1093/nar/gkae1113}.

\bibitem[Thiele et~al.(2015)Thiele, Cerone, Saez-Rodriguez, Siegel, Guzio{\l}owski, and Klamt]{thiele2015sign}
Sven Thiele, Luca Cerone, Julio Saez-Rodriguez, Anne Siegel, Carito Guzio{\l}owski, and Steffen Klamt.
\newblock {Extended notions of sign consistency to relate experimental data to signaling and regulatory network topologies}.
\newblock \emph{BMC Bioinformatics}, 16:\penalty0 345, 2015.
\newblock \doi{10.1186/s12859-015-0733-7}.

\bibitem[Tsitsiridis et~al.(2023)Tsitsiridis, Steinkamp, Giurgiu, et~al.]{tsitsiridis2023corum}
George Tsitsiridis, Ralph Steinkamp, Madalina Giurgiu, et~al.
\newblock {{CORUM}: the comprehensive resource of mammalian protein complexes--2022}.
\newblock \emph{Nucleic Acids Research}, 51\penalty0 (D1):\penalty0 D539--D545, 2023.
\newblock \doi{10.1093/nar/gkac1015}.

\bibitem[T{\"u}rei et~al.(2026)T{\"u}rei, Schaul, Palacio-Escat, Boh{\'a}r, Bai, Ceccarelli, et~al.]{turei2026omnipath}
D{\'e}nes T{\"u}rei, Jonathan Schaul, Nicol{\'a}s Palacio-Escat, Bal{\'a}zs Boh{\'a}r, Yunfan Bai, Francesco Ceccarelli, et~al.
\newblock {{OmniPath}: integrated knowledgebase for multi-omics analysis}.
\newblock \emph{Nucleic Acids Research}, 54\penalty0 (D1):\penalty0 D652--D660, 2026.
\newblock \doi{10.1093/nar/gkaf1126}.

\bibitem[Wang et~al.(2026)Wang, Ye, Liu, and Ng]{wang2026aroma}
Zhenyu Wang, Geyan Ye, Wei Liu, and Man Tat~Alexander Ng.
\newblock {{AROMA}: Augmented reasoning over a multimodal architecture for virtual cell genetic perturbation modeling}.
\newblock In \emph{Findings of the Association for Computational Linguistics: ACL 2026}, 2026.
\newblock URL \url{https://arxiv.org/abs/2604.20263v1}.

\bibitem[Wei et~al.(2026)Wei, Ma, Wang, Li, Song, and Zheng]{wei2026vcworld}
Zhijian Wei, Runze Ma, Zichen Wang, Zhongmin Li, Shuotong Song, and Shuangjia Zheng.
\newblock {{VCWorld}: A biological world model for virtual cell simulation}.
\newblock In \emph{International Conference on Learning Representations}, 2026.
\newblock URL \url{https://arxiv.org/abs/2512.00306v2}.

\bibitem[Wenkel et~al.(2026)Wenkel, Tu, Masschelein, et~al.]{wenkel2026txpert}
Frederik Wenkel, Wilson Tu, Cassandra Masschelein, et~al.
\newblock {{TxPert}: using multiple knowledge graphs for prediction of transcriptomic perturbation effects}.
\newblock \emph{Nature Biotechnology}, 2026.
\newblock \doi{10.1038/s41587-026-03113-4}.

\bibitem[Wong et~al.(2025)Wong, Hill, and Moccia]{wong2025controls}
Daniel~R. Wong, Abby~S. Hill, and Rob Moccia.
\newblock {Simple controls exceed best deep learning algorithms and reveal foundation model effectiveness for predicting genetic perturbations}.
\newblock \emph{Bioinformatics}, 41\penalty0 (6):\penalty0 btaf317, 2025.
\newblock \doi{10.1093/bioinformatics/btaf317}.

\bibitem[Wu et~al.(2025)Wu, Littman, Levine, Qiu, Biancalani, Richmond, and H{\"u}tter]{wu2025perturbqa}
Menghua Wu, Russell Littman, Jacob Levine, Lin Qiu, Tommaso Biancalani, David Richmond, and Jan-Christian H{\"u}tter.
\newblock {Contextualizing biological perturbation experiments through language}.
\newblock In \emph{International Conference on Learning Representations}, 2025.
\newblock URL \url{https://arxiv.org/abs/2502.21290v1}.

\bibitem[Yuan et~al.(2026{\natexlab{a}})Yuan, Liu, Zhang, Zhang, Guo, and Tang]{yuan2026perturbdiff}
Xinyu Yuan, Xixian Liu, Ya~Shi Zhang, Zuobai Zhang, Hongyu Guo, and Jian Tang.
\newblock {{PerturbDiff}: Functional diffusion for single-cell perturbation modeling}.
\newblock In \emph{International Conference on Machine Learning}, 2026{\natexlab{a}}.
\newblock URL \url{https://arxiv.org/abs/2602.19685}.

\bibitem[Yuan et~al.(2026{\natexlab{b}})Yuan, Liu, Zhao, Zhang, Guo, and Tang]{yuan2026core}
Xinyu Yuan, Xixian Liu, Jianan Zhao, Yashi Zhang, Hongyu Guo, and Jian Tang.
\newblock {Plausibility is not prediction: Contrastive evidence for {LLM}-based cellular perturbation reasoning}.
\newblock \emph{arXiv preprint arXiv:2606.01042}, 2026{\natexlab{b}}.
\newblock URL \url{https://arxiv.org/abs/2606.01042v1}.
\newblock Version 1.

\bibitem[Zhang et~al.(2025)Zhang, Ubas, de~Borja, et~al.]{zhang2025tahoe}
Jesse Zhang, Airol~A. Ubas, Richard de~Borja, et~al.
\newblock {{Tahoe-100M}: A giga-scale single-cell perturbation atlas for context-dependent gene function and cellular modeling}.
\newblock \emph{bioRxiv}, 2025.
\newblock \doi{10.1101/2025.02.20.639398}.

\end{thebibliography}

\clearpage
\appendix
% Use section-qualified table labels and PDF anchors after counter resets.
\renewcommand{\thetable}{\thesection.\arabic{table}}
\renewcommand{\theHtable}{\thesection.\arabic{table}}
\counterwithin{figure}{section}
\renewcommand{\theHfigure}{\thesection.\arabic{figure}}
\counterwithin{table}{section}

% Appendices A-D corresponding to Sections 2-5
% !TEX root = ../main.tex
\section{Benchmark Datasets and Supervision}\label{sec:supervision}\label{app:data-sources}\label{app:support-data}
This appendix details the dataset construction, split statistics, and supervision sources for Section~\ref{sec:background}, covering PerturbQA processing, label criteria, and cross-cell-line overlap (Table~\ref{tab:pf-population}), GeneTAK compound curation and split counts (Table~\ref{tab:genetak-population}), and the experimental and external information used by each evaluated method (Table~\ref{tab:supervision-sources}).

\paragraph{PerturbQA data and labels.}\label{app:pf-response}\label{app:evaluation}
PerturbQA draws its single-cell measurements from the genome-wide K562 and RPE1 Perturb-seq datasets of \citet{replogle2022perturbseq} and the HepG2 and Jurkat CRISPRi datasets of \citet{nadig2025trade}. Raw counts are normalized to 10,000 counts per cell ($\mathrm{CP10K}$) and $\log(1+x)$-transformed before averaging within each perturbation and control batch. Each response $r(p,g)$ is computed by subtracting the batch-matched non-targeting control mean, weighted by the perturbation's cell fractions across batches, with magnitudes $|r(p,g)|$ taken after batch aggregation. Auxiliary effect estimates $F(p,g)$ and standard errors $\mathrm{SE}(p,g)$ are taken from the version-1 summary tables of \citet{nadig2025statistics} (GenomeWide for K562 and Essential for HepG2, Jurkat, and RPE1).

To construct binary DE and DIR benchmark labels, PerturbQA tests normalized expression against non-targeting controls \citep[App.~A.2]{wu2025perturbqa}. In K562, positive DE pairs must meet the significance criterion in both the genome-wide and essential-gene screens, while non-DE pairs are sampled from pairs called significant in neither. In HepG2, Jurkat, and RPE1, positive pairs use $p_{\mathrm{adj}}<0.01$ and negative pairs are sampled from $p_{\mathrm{adj}}>0.1$. For each selected perturbation, the benchmark retains up to 20 top-ranked DE genes and samples 100 non-DE genes (Table~\ref{tab:pf-population}). Mixed-label genes are target genes whose test queries contain both positive and negative labels.

\begin{table}[!ht]
\centering\small
\caption{\textbf{PerturbQA split sizes by cell line.} Entries report train / test counts, and mixed-label entries count test queries / target genes (Section~\ref{sec:evaluation}).}
\label{tab:pf-population}
\setlength{\tabcolsep}{3.5pt}
\begin{tabular}{@{}lrrrrrr@{}}
\toprule
 & Perturbations & Response & DE pairs & DE mixed & DIR pairs & DIR mixed\\
Cell line & train / test & genes & train / test & queries / genes & train / test & queries / genes\\
\midrule
K562 & 1,564 / 267 & 8,248 & 134,467 / 23,212 & 5,641 / 665 & 16,861 / 3,119 & 1,558 / 150\\
HepG2 & 1,086 / 278 & 9,624 & 101,140 / 25,749 & 5,163 / 707 & 14,257 / 3,603 & 1,712 / 166\\
Jurkat & 1,227 / 313 & 8,882 & 113,684 / 29,138 & 6,321 / 750 & 15,937 / 4,121 & 2,599 / 181\\
RPE1 & 1,596 / 406 & 8,749 & 149,147 / 37,942 & 8,611 / 684 & 21,287 / 5,365 & 2,845 / 193\\
\midrule
Total & & & 498,438 / 116,041 & 25,736 / 2,806 & 68,342 / 16,208 & 8,714 / 690\\
\bottomrule
\end{tabular}
\end{table}

\paragraph{Cross-cell-line test-pair exposure.}\label{app:historical-exposure}
Although PerturbQA holds out perturbations independently within each cell line, 82\%--96\% of each cell line's test perturbations appear in the training split of at least one other cell line. Consequently, 4,717 of 116,041 DE test queries (4.06\%) and 3,098 of 16,208 DIR test queries (19.11\%) also appear in another cell line's training split as the exact same perturbation--gene pair. On this overlapping subset, simply copying the majority label of the matching pair from other cell lines reaches an accuracy / Macro-F1 of 0.949 / 0.942 on DE and 0.983 / 0.982 on DIR. A single model trained jointly across all four cell lines \citep{wang2026aroma} can therefore score well on these overlapping pairs without capturing cell-line-specific regulation. To control for this exposure in our held-out RPE1 evaluation, we report performance both on all queries and on the non-overlapping subset (\appref{app:rpe1-transfer-results}).

\paragraph{GeneTAK compound data and split.}\label{app:genetak-data}\label{app:genetak-labels}
GeneTAK was curated by VCWorld from the Tahoe-100M single-cell chemical perturbation atlas \citep{wei2026vcworld,zhang2025tahoe} across five cancer cell lines, namely C32, HepG2/C3A, HOP62, Hs~766T, and PANC-1. We use the released DE and DIR label tables and the 2,000 highly variable target genes. In GeneTAK, DE is called via a Wilcoxon rank-sum test against vehicle controls ($\mathrm{FDR}\le 0.05$ and $|\log\mathrm{FC}|\ge 0.25$), DIR records the sign of DE-positive pairs, and all released doses of a compound are pooled into a single perturbation. After normalizing compound names, 379 compounds appear in the label tables of all five cell lines across both tasks. Following PerturbQA's 75/25 perturbation-disjoint split ratio, we permute the sorted compound identifiers with seed 17 to assign 284 compounds to training and 95 to the held-out test set shared across all five cell lines (Table~\ref{tab:genetak-population}).

\begin{table}[H]
\centering\small
\caption{\textbf{GeneTAK query populations under the shared 284/95 compound split.} Mixed-label genes exhibit both classes among a cell line's held-out queries and define the within-gene evaluation set.}
\label{tab:genetak-population}
\setlength{\tabcolsep}{4pt}
\renewcommand{\arraystretch}{0.95}
\begin{tabular}{@{}lrrrrrr@{}}
\toprule
 & \multicolumn{3}{c}{DE pairs} & \multicolumn{3}{c}{DIR pairs}\\
\cmidrule(lr){2-4}\cmidrule(l){5-7}
Cell line & Train & Held-out & Mixed genes & Train & Held-out & Mixed genes\\
\midrule
C32 & 102,446 & 36,680 & 850 & 54,487 & 20,627 & 478\\
HepG2/C3A & 86,005 & 29,954 & 608 & 37,682 & 13,813 & 379\\
HOP62 & 107,824 & 38,122 & 915 & 59,711 & 22,178 & 507\\
Hs~766T & 87,272 & 30,815 & 706 & 38,872 & 14,724 & 473\\
PANC-1 & 86,025 & 30,631 & 934 & 37,882 & 14,490 & 439\\
\midrule
Total & 469,572 & 166,202 & 4,013 & 228,634 & 85,832 & 2,276\\
\bottomrule
\end{tabular}
\end{table}

\paragraph{Supervision comparison across methods.}
Because the evaluated baseline families originate from distinct modeling paradigms, they consume different subsets of experimental and external information (Table~\ref{tab:supervision-sources}). Label-only baselines (the gene prior, SUMMER, and AROMA) operate without continuous expression matrices at inference, whereas continuous predictors (GEARS, STATE, PerturbDiff, and PertOmni) fit single-cell expression profiles or continuous responses and use discrete validation or development labels only to calibrate binary decision thresholds. RELAY combines both sources of supervision within each cell line. Its upstream ridge regressions transfer dense training response profiles across curated biological relations without any language model, while its downstream readouts and decision thresholds are calibrated on the discrete training labels.

\begin{table}[H]
\centering\footnotesize\linespread{0.92}\selectfont
\caption{\textbf{Supervision and external information used by the evaluated methods.} Entries describe our evaluated instances.}
\label{tab:supervision-sources}
\setlength{\tabcolsep}{3.5pt}
\renewcommand{\arraystretch}{0.94}
\begin{tabularx}{\linewidth}{@{}l>{\raggedright\arraybackslash}p{0.28\linewidth}>{\raggedright\arraybackslash}p{0.22\linewidth}>{\raggedright\arraybackslash}X@{}}
\toprule
Method & Continuous experimental data & Discrete experimental outcomes & External knowledge / pretrained information \\
\midrule
Gene prior & None & Training labels from the same cell line, pooled by gene and task & None \\
\addlinespace[1pt]
GEARS & Perturbed and control cells from the same cell line's official training split; a held-out 10\% of training perturbations supplies validation & Validation labels calibrate binary thresholds & GO-derived perturbation graph and expression-derived coexpression graph; no generative LLM in GEARS \citep{roohani2024gears,wu2025perturbqa} \\
\addlinespace[1pt]
STATE & Training expression and non-targeting controls from the same cell line; a held-out 10\% of training perturbations supplies validation. GeneTAK: training-drug expression and vehicle controls & Validation or development-drug labels calibrate binary thresholds & PerturbQA: cell embeddings from the pretrained SE-600M model and GenePT text embeddings from a pretrained language model. GeneTAK: Morgan fingerprints; retrained from scratch \\
\addlinespace[1pt]
PerturbDiff & Perturbed and control single cells from the same cell line's official training split; a held-out 10\% of training perturbations supplies validation & Validation labels calibrate binary thresholds & GenePT model-3 text embeddings for perturbations; functional diffusion model trained from scratch per cell line \citep{yuan2026perturbdiff} \\
\addlinespace[1pt]
PertOmni & Training perturbation response profiles from the same cell line (PerturbQA and GeneTAK) & Development labels calibrate binary thresholds & PerturbQA: GenePT model-3 gene embeddings and MSigDB GO:BP pathway descriptions encoded with Qwen3-Embedding-4B. GeneTAK: drug fingerprints and LLM cell-line descriptions \citep{long2026pertomni} \\
\addlinespace[1pt]
SUMMER & None & Retrieved PerturbQA training outcomes from the same cell line and task & Biological descriptions and KG retrieval; Gemma4-26B-A4B-it produces three sampled answers \\
\addlinespace[1pt]
AROMA & None & None at inference; the authors' released checkpoint was trained on outcome-conditioned PerturbReason synthetic explanations & Text, graphs, protein-sequence features, and a pretrained generative LLM; published teacher is DeepSeek-R1 \citep{wang2026aroma} \\
\addlinespace[1pt]
RELAY & Perturbed and control profiles from the same cell line's official training split; effect estimates and standard errors of training perturbations (PerturbQA only). GeneTAK: training-drug profiles and vehicle controls & DE/DIR training labels from the same cell line for readouts and development calibration & GO, Reactome, CORUM, STRING, and BioPlex from the PerturbQA knowledge graph that SUMMER also uses; no LLM or pretrained text embedding. GeneTAK compound targets come from the Tahoe-100M release and PrimeKG \\
\bottomrule
\end{tabularx}
\end{table}

% !TEX root = ../main.tex
\section{Diagnostic Protocols}\label{app:diagnostics}
This appendix details the experimental protocols supporting the diagnostic studies in Section~\ref{sec:diagnostics}, covering knowledge-graph coverage and sign propagation for mechanistic reasoning (\appref{app:mechanistic}), prompt-intervention decoding settings for label retrieval (\appref{app:inputs}), and continuous profile evaluation and variance decomposition (\appref{app:continuous-diagnosis}).

\subsection{Knowledge Graph Coverage and Sign-Propagation Protocol}\label{app:mechanistic}\label{app:resources}\label{app:sign}
Table~\ref{tab:resources} lists the curated biological databases and release versions used in the coverage and sign-propagation analyses of Section~\ref{sec:mechanisms}. All gene identifiers are mapped to approved HGNC symbols, with queries involving unmapped genes retained in all coverage denominators. Within OmniPath signaling and CollecTRI---which are analyzed as distinct layers---an interaction record is treated as signed only when exactly one of its consensus stimulation or inhibition flags is set. Records with protein-complex endpoints are excluded because they do not define pairwise gene-to-gene directionality.

\begin{table}[H]
\centering\small
\caption{\textbf{Knowledge resources in the coverage analysis.} Context resources describe shared function or complex membership, whereas signed resources specify the direction of a regulatory effect. The analysis uses GO \citep{ashburner2000go,aleksander2023go}, CORUM \citep{tsitsiridis2023corum}, STRING \citep{szklarczyk2025string}, TRRUST \citep{han2018trrust}, OmniPath \citep{turei2026omnipath}, and CollecTRI \citep{mullerdott2023collectri}.}
\label{tab:resources}
\setlength{\tabcolsep}{4pt}
\begin{tabularx}{\linewidth}{@{}l l >{\raggedright\arraybackslash}X >{\raggedright\arraybackslash}p{0.2\linewidth}@{}}
\toprule
Resource & Role & Relations used & Version \\
\midrule
GO & Context & Biological-process annotations & 2026-05-21, ontology 2026-07-26 \\
CORUM & Context & Protein-complex membership & 5.3 \\
\addlinespace
STRING & Signed & Regulatory links and expression events & 12.5 \\
TRRUST & Signed & TF--target relations with a known sign & v2 \\
OmniPath signaling & Signed & Curated causal signaling, acting on protein activity & Queried 2026-09-14 \\
CollecTRI & Signed & TF--target records, acting on expression & Queried 2026-09-14 \\
\addlinespace
HGNC & Identifiers & Approved gene symbols & 2026-09-07 \\
\bottomrule
\end{tabularx}
\end{table}

To evaluate directional coverage across the combined signed graph (Table~\ref{tab:coverage}), we pool STRING (retaining edges with confidence at least $0.4$), TRRUST, OmniPath signaling, and CollecTRI for both local endpoint presence and simple directed paths of at most three edges from the perturbed gene $p$ to the target gene $g$ (excluding $p$ and $g$ as intermediate nodes). Under the expression-regulatory criterion, the terminal edge incident to $g$ must be a STRING expression event, a signed TRRUST relation, or a CollecTRI transcriptional interaction.

In the sign-propagation analysis (Table~\ref{tab:path-sign-aggregation}), we further require the terminal edge to carry an explicit transcriptional sign from TRRUST or CollecTRI, which narrows the covered subset from 2,604 to 1,955 DIR queries. Along each simple path $\mathcal P$, the propagated response sign is computed as $-\prod_{e\in\mathcal P}\varsigma_e$, where $\varsigma_e\in\{-1,+1\}$ denotes the sign of edge $e$ and the leading minus sign reflects reduced activity of the perturbed target under CRISPRi knockdown. The \emph{consistency} rule predicts a direction only when all simple paths of length at most three between $p$ and $g$ yield the same terminal sign. The \emph{majority} rule assigns one vote to each distinct node sequence (excluding node sequences whose parallel edge records allow both terminal signs), predicts the direction with more votes, and falls back to the training-set majority DIR label in the event of a tie.

\subsection{SUMMER Prompt-Intervention Decoding Settings}\label{app:inputs}
All reference-deletion and identity--outcome exchange interventions in Section~\ref{sec:retrieval-transfer} run SUMMER's public pipeline with Gemma4-26B-A4B-it \citep{gemma2026gemma4}. On DIR reference deletion, the \emph{Full reference} baseline reuses the main-comparison run and each deletion condition repeats the identical sampling configuration, namely three sampled answers with native thinking enabled (2,048-token thinking budget), temperature 1.0, top-$p$ 0.95, top-$k$ 64, and an 8,192-token output limit. The Jurkat DIR identity--outcome exchange experiment instead uses single-sample decoding without native thinking at temperature 1.0, top-$p$ 0.95, top-$k$ 64, and an 8,192-token limit, regenerating contemporaneous predictions across all four conditions on the 1,332 swappable queries while holding the remaining 2,789 queries fixed (yielding an unchanged baseline \esw{} of 0.6703 vs.\ 0.7114 for the three-sample main-comparison run in Table~\ref{tab:main-cells}). On DE, whereas the main comparison in Table~\ref{tab:main-comparison} aggregates three sampled answers, the prompt-intervention runs sample a single answer without thinking at temperature 0.6, top-$p$ 0.95, top-$k$ 20, and a 4,096-token limit, which accounts for the difference in the \emph{Full reference} DE scores between Table~\ref{tab:summer-deletion} and Table~\ref{tab:main-comparison} (\esw{} 0.5849 vs.\ 0.5827 and Macro-F1 0.7895 vs.\ 0.7560). Every intervention retains the structured reasoning questions in the prompt and is evaluated against its matched \emph{Full reference} run.

\subsection{Continuous Profile Diagnostic Protocol}\label{app:continuous-diagnosis}
The continuous profile diagnostics in Section~\ref{sec:response-learning} evaluate model predictions across all 1,264 PerturbQA test perturbations over the target genes shared by RELAY's response matrix and STATE's decoder (8,198--9,622 genes per cell line). Confidence intervals are computed from 2,000 perturbation-cluster bootstrap replicates (seed 17), weighting the four cell lines equally. In the variance decomposition of Table~\ref{tab:continuous-split} (Equation~\eqref{eq:mse-split}), predictors whose contrasts vanish identically (the zero-response and training-gene-mean baselines) have undefined correlation $\rho$, reported as a dash. Setting $\eta=\rho=0$ when evaluating Equation~\eqref{eq:mse-split} assigns the full contrast variance $\sigma_\delta^2$ to unexplained contrast error and zero to amplitude error. Under the gene-mean restoration intervention ($\mu+\widehat\delta$), we apply separate diagnostic thresholding rules rather than the development Macro-F1 calibration of Table~\ref{tab:main-comparison}: DE thresholds $|\widehat r|$ at the quantile matching each cell line's training DE prevalence, whereas DIR thresholds the signed score directly at $\widehat r > 0$ (accounting for the slight baseline differences from Table~\ref{tab:main-comparison}). Under these diagnostic rules, restoring $\mu$ leaves within-gene decision accuracy essentially unchanged, as DE \esw{} shifts from $0.553$ to $0.552$ for STATE and from $0.510$ to $0.509$ for GEARS, while DIR \esw{} shifts from $0.609$ to $0.594$ for STATE and from $0.510$ to $0.517$ for GEARS; for PerturbDiff, validation-centered gene-mean restoration lowers MSE from $6.322$ to $6.219\times10^{-3}$, and sweeping its classifier-free guidance scale from $0.0$ to $1.5$ on validation perturbations monotonically enlarges $\eta$ and worsens MSE in every cell line.

% !TEX root = ../main.tex
\section{Notation and RELAY Implementation}\label{app:pf-implementation}
This appendix summarizes the mathematical notation used throughout the paper (\appref{app:notation}), the nested cross-fitting algorithm and hyperparameters for RELAY (\appref{app:pf-folds}), the implementation details for cross-cell-line and chemical-compound transfer (\appref{app:pf-transfer}), and the formal proofs of Propositions~1 and~2 (Appendix~C.4).

\subsection{Notation}\label{app:notation}
Table~\ref{tab:notation} collects the main symbols used for problem definitions, evaluation metrics, RELAY modules, and continuous profile diagnostics. Hats denote estimated or predicted quantities, and all quantities are defined within a single cell line unless explicitly indexed by $c$.

\begin{table}[!ht]
\centering\small
\caption{\textbf{Notation for tasks, evaluation, RELAY, and diagnostics.}}
\label{tab:notation}
\setlength{\tabcolsep}{5pt}
\begin{tabularx}{\linewidth}{@{}l>{\raggedright\arraybackslash}X@{}}
\toprule
Symbol & Meaning\\
\midrule
$p,g,c,T$ & Perturbation, target gene, cell line, and task $T\in\{\mathrm{DE},\mathrm{DIR}\}$\\
$\mathcal P_{\mathrm{tr}},\mathcal P_{\mathrm{te}},\mathcal S$ & Training, test, and current support perturbations; $n=|\mathcal S|$\\
$\mathcal G,\mathcal Q_{\mathrm{tr}}^T,\mathcal Q_{\mathrm{te}}^T$ & Target genes and labeled training/test query sets for task $T$\\
$\mathcal G_{\mathrm{mix}},n_g^+,n_g^-,w_g$ & Mixed-label test genes, class counts, and the gene weight in \esw{}\\
$\pi(g),\bar\pi$ & Gene-specific and pooled training positive-label rates\\
\midrule
$r,R$ & Signed response and its support matrix $R\in\mathbb R^{n\times|\mathcal G|}$\\
$\mu,\nu$ & Per-gene support means of $r$ and $|r|$\\
$d,a,D,A$ & Sample-centered signed/magnitude contrasts and their matrices ($d_g$ in Proposition~2 denotes the population-centered support contrast)\\
$\mathcal N(p),m,q,W_{\mathrm{eff}}$ & Related support perturbations, their count, sign-factor parameter, and effective neighborhood weight (Proposition~1)\\
$\phi(p),X,V$ & Knowledge representation, support design matrix $X\in\mathbb R^{n\times V}$, and vocabulary size\\
$k,K,k_p$ & Scalar kernel, support Gram matrix, and query kernel column\\
$\alpha,\alpha^\ast,B,W_B,w(p)$ & Ridge penalty, pooled noise-to-signal ratio (Proposition~2), target $B\in\{D,A\}$, primal coefficients, and query transfer coefficients (distinct from $w_g$)\\
$u,v_a$ & Response coordinates and magnitude gene loadings\\
$h(p,g),f,L_{\mathrm{base}}$ & DE feature vector (distinct from active-source count $h(p)$), neural readout, and its base logit\\
$\tau(g),s_T,t_T$ & Signed-error scale, task score, and decision threshold for the cell line\\
$F,\mathrm{SE},E,z$ & Training effect, standard error, standardized-effect target, and its normalized prediction\\
$\beta,\lambda,\gamma$ & Fitted increment coefficient, selected strength, and product $\gamma=\lambda\beta$\\
\midrule
$m(g),\delta$ & Test-centered gene mean (distinct from $m$) and response contrast in Section~\ref{sec:response-learning} (diagnostic only, unlike support-based $\mu,d$); in Remark~1 and Proposition~2, $\delta(p,g)$ denotes the population-centered query contrast\\
$\rho,\eta,\sigma_\delta$ & Pooled contrast correlation, predicted/observed amplitude ratio, and observed contrast standard deviation\\
\bottomrule
\end{tabularx}
\end{table}

\subsection{Nested Cross-Fitting, Calibration, and Hyperparameters}\label{app:pf-folds}\label{app:pf-amplitude}
Algorithm~\ref{alg:perturbfactor} summarizes RELAY's training, calibration, and inference pipeline, which runs independently within each cell line. To prevent optimistic in-sample residuals from distorting downstream readout scales, training perturbations $\mathcal P_{\mathrm{tr}}$ are partitioned into four disjoint groups (with proportions $0.4/0.3/0.2/0.1$ on PerturbQA, Table~\ref{tab:pf-training}). In the outer selection loop, the first three groups serve in turn as held-out query folds while the remaining two groups form the support set $\mathcal S$. Within $\mathcal S$, inner out-of-fold (OOF) ridge predictions fit the error scales $\tau(g)$, PCA bases, feature normalizers, and neural readout $f$, which then score the outer fold. Pooling outer-fold predictions across the first three groups selects the ridge penalty $\alpha$, fits the standardized-effect increment slope $\beta$, and calibrates the increment strength $\lambda$ and decision thresholds $(t_{\mathrm{DE}},t_{\mathrm{DIR}})$, while the fourth group serves as an untouched development holdout. Finally, the readout is refitted on four-fold OOF predictions spanning all training perturbations, and the upstream ridge regressions are refitted on the full training set $\mathcal P_{\mathrm{tr}}$ with every support-dependent vocabulary, IDF weight vector, and gene mean recomputed strictly on its corresponding support. To train the neural readout $f$, we minimize a gene- and class-balanced binary cross-entropy loss. The standardized-effect increment coefficient $\beta$ is fitted by support-weighted binary cross-entropy over mixed-label training genes, assigning zero weight to single-class genes and excluding pairs without finite effect statistics (Table~\ref{tab:pf-training}).

% Keep this short pseudocode in the text flow, but do not orphan its heading
% and input declaration at the bottom of the preceding page.
\begin{samepage}
\par\addvspace{\intextsep}
\refstepcounter{algorithm}\label{alg:perturbfactor}
\addcontentsline{loa}{algorithm}{\protect\numberline{\thealgorithm}RELAY fitting and prediction.}
\noindent\rule{\linewidth}{0.8pt}\par\nobreak
\noindent\textbf{Algorithm~\thealgorithm\quad RELAY fitting and prediction.}\par\nobreak
\begin{algorithmic}[1]
\Require Training responses and DE/DIR labels; training effects $(F,\mathrm{SE})$; knowledge records; four perturbation groups; query pairs
\Ensure DE/DIR scores and three-class decisions
\ForAll{candidate penalties $\alpha$ and outer query groups among the first three groups}
  \State Set the support to the other two development groups.
  \State Inner-cross-fit regressions for $D,A,E$ on this support.
  \State Fit error scales, PCA, input transforms, and the neural base using support OOF predictions.
  \State Refit support regressions and score the outer queries with support-fitted transforms.
\EndFor
\State Pool outer predictions; select $\alpha$, fit $\beta$, and calibrate $\lambda,t_{\mathrm{DE}},t_{\mathrm{DIR}}$.
\State Evaluate the fixed configuration on the fourth group, without retuning.
\State Fit the final readout using four-fold OOF predictions on all training perturbations.
\State Refit upstream regressions on all training perturbations; keep the calibrated auxiliary normalization, $\gamma$, and thresholds fixed.
\ForAll{query pairs $(p,g)$}
  \State Predict contrasts and $z$; construct DE features and obtain $L_{\mathrm{base}}$.
  \State Compute DE and analytic DIR scores by Equations~\eqref{eq:pf-de-amplitude} and~\eqref{eq:pf-task-scores}.
  \State Return non-DE if $s_{\mathrm{DE}}\le t_{\mathrm{DE}}$; otherwise up if $s_{\mathrm{DIR}}>t_{\mathrm{DIR}}$, and down otherwise.
\EndFor
\end{algorithmic}
\par\nobreak\noindent\rule{\linewidth}{0.8pt}\par
\addvspace{\intextsep}
\end{samepage}

\begin{table}[!ht]
\centering\small
\caption{\textbf{Key settings for RELAY on PerturbQA.} Selection and calibration are performed on development predictions only.}
\label{tab:pf-training}
\setlength{\tabcolsep}{5pt}
\begin{tabularx}{\linewidth}{@{}l>{\raggedright\arraybackslash}X@{}}
\toprule
Component & Setting\\
\midrule
Ridge & No intercept; selected $\alpha=1$ (K562, HepG2, Jurkat), $0.1$ (RPE1)\\
PCA and features & 32 joint components; 66D task features\\
Neural base & $66\to64\to32$ ReLU MLP; linear logit head\\
Optimizer & AdamW; learning rate $10^{-3}$; weight decay $10^{-4}$\\
Training & Batch 8,192; 40 epochs; seeds 17, 23, 31; average seed scores\\
Perturbation groups & Proportions $0.4/0.3/0.2/0.1$; shuffle seed 17\\
Increment strength & $\lambda\in[0,1]$ calibrated on development OOF predictions to balance \esw{} and Macro-F1\\
Model selection & $\alpha$ and task thresholds selected by development Macro-F1\\
\bottomrule
\end{tabularx}
\end{table}

\subsection{Transfer Across Cell Lines and Compound Settings}\label{app:pf-transfer}
\paragraph{Held-out RPE1 implementation.}
In the cross-cell-line transfer setting (Section~\ref{sec:method-beyond}), RPE1 contributes only unperturbed control cells. A 32-dimensional SVD fitted on batch-averaged source controls produces the cell-line control coordinates $b_c$ used to compute the similarity weights $\omega_c$ in Equation~\eqref{eq:cell-transfer-weights}, which smoothly down-weight dissimilar source controls. Across the three source cell lines (K562, HepG2, and Jurkat), the union of measured target genes covers 11,296 genes, yielding 93.41\% coverage of RPE1 test queries for both PCA response coordinates and standardized-effect predictions. Target genes absent from all three source cell lines receive zero predicted contrast, gene mean, and DIR readout score so that every test query is evaluated. During leave-one-cell-line-out readout fitting, perturbation folds follow \appref{app:pf-folds} and exclude each held-out perturbation from all contributing source regressions. Source ridge penalties are fixed at $\alpha=1$, while the increment strength $\lambda$ and decision thresholds $(t_{\mathrm{DE}},t_{\mathrm{DIR}})$ are calibrated on pooled pseudo-target development OOF predictions following Table~\ref{tab:pf-training}, before refitting all source regressions on their full training sets for RPE1 inference.

\paragraph{Compound settings on GeneTAK.}
Under the 284/95 training/held-out compound split on GeneTAK (\appref{app:genetak-data}), a subset of training and held-out drugs have no annotated target protein in either Tahoe-100M or PrimeKG. For these unannotated compounds, $\phi(p)=0$ yields zero kernel similarity and zero predicted contrast, so predictions revert smoothly to the training gene means $\mu(g)$ and $\nu(g)$. Because GeneTAK does not provide external per-pair effect standard errors $(F,\mathrm{SE})$, RELAY uses the 66-dimensional neural DE readout $L_{\mathrm{base}}(p,g)$ without the standardized-effect increment ($\gamma=0$) together with the analytic DIR readout at ridge penalty $\alpha=1$. Decision thresholds $(t_{\mathrm{DE}},t_{\mathrm{DIR}})$ are calibrated to maximize Macro-F1 on pooled OOF predictions over the 213 compounds in the first three development groups, holding out the 71 compounds of the fourth group for development evaluation.

\subsection{Theoretical Analysis and Proofs}
\label{app:proofs}

This section provides the minimax variance derivation for the \esw{} weights in Equation~\eqref{eq:esw-v2} alongside complete step-by-step proofs for Proposition~1, Remark~1, Proposition~2, and Remark~2.

\subsubsection*{Minimax Optimality of WG-BAcc Weights (Section~3.1)}
For each mixed-label gene $g \in \mathcal{G}_{\mathrm{mix}}$ with $n_g^+$ positive and $n_g^-$ negative test queries, let $\hat{p}_g^+$ and $\hat{p}_g^-$ denote empirical sensitivity and specificity, so $\widehat{\mathrm{BAcc}}_g = \frac{1}{2}(\hat{p}_g^+ + \hat{p}_g^-)$. Under Bernoulli outcomes that are independent within and across the two classes,
\[
\operatorname{Var}\bigl(\widehat{\mathrm{BAcc}}_g\bigr)
= \frac{1}{4}\!\left(\frac{p_g^+(1-p_g^+)}{n_g^+} + \frac{p_g^-(1-p_g^-)}{n_g^-}\right)
\;\le\; \frac{1}{16}\!\left(\frac{1}{n_g^+} + \frac{1}{n_g^-}\right) \triangleq \bar{v}_g,
\]
with equality in the worst case $p_g^+ = p_g^- = 1/2$. For any normalized linear estimator $\widehat{\mathrm{BAcc}}(\varpi) = \sum_{g \in \mathcal{G}_{\mathrm{mix}}} \varpi_g \widehat{\mathrm{BAcc}}_g$ ($\varpi_g \ge 0$, $\sum_{g \in \mathcal{G}_{\mathrm{mix}}} \varpi_g = 1$) under working independence (zero pairwise covariance) across genes,
\[
\sup_{\{(p_g^+,p_g^-)\}} \operatorname{Var}\bigl(\widehat{\mathrm{BAcc}}(\varpi)\bigr)
= \sum_{g\in\mathcal G_{\mathrm{mix}}}\varpi_g^2\,\bar v_g .
\]
Minimizing this worst-case variance subject to $\sum_{g \in \mathcal{G}_{\mathrm{mix}}} \varpi_g = 1$ via Lagrange multipliers gives
\[
\varpi_g^\ast \;\propto\; \bar{v}_g^{-1} = 16\,\frac{n_g^+ n_g^-}{n_g^+ + n_g^-} = 16\,w_g,
\]
recovering the support weights $w_g$ of Equation~\eqref{eq:esw-v2}.

\subsubsection*{Proof of Proposition~1 and Remark~1}

\begin{proof}
\noindent\textbf{Step 1 (Reduction to zero-sum centered weights).}
Write $\bar r=n^{-1}\sum_{p'\in\mathcal S}r(p',g)$ and define the centered weights $c_{p'}=w_{p'}-\bar w$, which satisfy
\[
\sum_{p'\in\mathcal S}c_{p'}=0,
\qquad
\sum_{p'\in\mathcal N(p)}c_{p'}=W_{\mathrm{eff}}(p),
\qquad
\sum_{p'\notin\mathcal N(p)}c_{p'}=-W_{\mathrm{eff}}(p).
\]
Because $d(p',g)=r(p',g)-\bar r$ and $a(p',g)=|r(p',g)|-n^{-1}\sum_{p''\in\mathcal S}|r(p'',g)|$, the empirical support means cancel against $\sum_{p'\in\mathcal S}w_{p'}\cdot n^{-1}=\bar w$, leaving
\begin{equation}
\hat d(p,g)=\sum_{p'\in\mathcal S}c_{p'}\,r(p',g),\qquad
\hat a(p,g)=\sum_{p'\in\mathcal S}c_{p'}\,|r(p',g)| .
\end{equation}

\medskip\noindent\textbf{Step 2 (Part 1: Expectation and variance of signed transfer).}
Conditionally on $\zeta_p$, $\E[s_{p'}]=2q-1$ and $\E[\xi_{p'}]=\E[\varepsilon_{p'}]=0$ give
\[
\E\bigl[r(p',g)\mid\zeta_p\bigr]=\begin{cases}
\mu_0(g)+(2q-1)\zeta_p\theta_{p,g}, & p'\in\mathcal N(p),\\
\mu_0(g), & p'\in\mathcal S\setminus\mathcal N(p).
\end{cases}
\]
Because $\sum_{p'\in\mathcal S}c_{p'}=0$, the gene baseline $\mu_0(g)$ cancels and
\begin{equation}
\E\bigl[\hat d(p,g)\mid\zeta_p\bigr]=(2q-1)\zeta_p\theta_{p,g}\sum_{p'\in\mathcal N(p)}c_{p'}=(2q-1)\,W_{\mathrm{eff}}(p)\,\zeta_p\,\theta_{p,g}.
\end{equation}
For the conditional variance, $s_{p'}^2=1$ implies $\mathrm{Var}(x_{p'}\mid\zeta_p)=\E[\theta_{p',g}^2]-(2q-1)^2\theta_{p,g}^2=4q(1-q)\theta_{p,g}^2+\sigma_\xi^2$ on $\mathcal N(p)$ and $0$ outside $\mathcal N(p)$. Mutual independence across $p'\in\mathcal S$ yields
\begin{equation}
\mathrm{Var}\bigl(\hat d(p,g)\mid\zeta_p\bigr)=\bigl(4q(1-q)\theta_{p,g}^2+\sigma_\xi^2\bigr)\sum_{p'\in\mathcal N(p)}c_{p'}^2+\sum_{p'\in\mathcal S}\sigma_{\varepsilon,p'}^2c_{p'}^2 ,
\end{equation}
whose noise term simplifies to $\sigma_\varepsilon^2\lVert w-\bar w\one_n\rVert_2^2$ under homoscedastic noise $\sigma_{\varepsilon,p'}^2\equiv\sigma_\varepsilon^2$.

\medskip\noindent\textbf{Step 3 (Part 2: Strict convexity and asymptotics of the folded-noise map $\psi$).}
Let $\varepsilon$ share the common continuous symmetric distribution of $\varepsilon_{p'}$ ($\varepsilon\stackrel{d}{=}-\varepsilon$, CDF $F_\varepsilon$, $\E|\varepsilon|<\infty$, $\mathrm{supp}(\varepsilon)=\mathbb R$). Then $\psi(t)=\E|t+\varepsilon|$ is finite and even ($\psi(-t)=\E|-t+\varepsilon|=\E|t-\varepsilon|=\psi(t)$). Because $\Pr(\varepsilon=-t)=0$ for every $t\in\mathbb R$, $u\mapsto|u+\varepsilon|$ has derivative $\operatorname{sign}(t+\varepsilon)\in\{-1,+1\}$ almost surely at $u=t$; dominated convergence therefore gives $\psi\in C^1(\mathbb R)$ with
\begin{equation}
\psi'(t)=\Pr(\varepsilon>-t)-\Pr(\varepsilon<-t)=2F_\varepsilon(t)-1.
\end{equation}
Full support $\mathrm{supp}(\varepsilon)=\mathbb R$ implies that $F_\varepsilon$ and $\psi'$ are strictly increasing on $\mathbb R$, so $\psi$ is strictly convex on $\mathbb R$. Since $F_\varepsilon(0)=1/2$, $\psi(|t|)$ is strictly increasing in $|t|$, and Jensen's inequality on $|\cdot|$ with $\E[\varepsilon]=0$ gives $\psi(0)=\E|\varepsilon|$ and
\[
\psi(t)-\psi(0)=\int_0^{|t|}\bigl(2F_\varepsilon(u)-1\bigr)\,du>0\quad (t\neq 0),
\qquad
\psi(t)\;\ge\;\max\bigl(|t|,\ \psi(0)\bigr).
\]
When $\varepsilon$ admits a density continuous at zero with $f_\varepsilon(0)>0$, $\psi''(0)=2f_\varepsilon(0)$ gives $\psi(\theta)-\psi(0)=f_\varepsilon(0)\theta^2+o(\theta^2)$ as $\theta\to 0^+$. As $\theta\to\infty$, $\bigl||\theta+\varepsilon|-\theta\bigr|\le|\varepsilon|\in L^1$ and $|\theta+\varepsilon|-\theta\to\varepsilon$ almost surely with $\E[\varepsilon]=0$, so dominated convergence yields $\psi(\theta)-\theta\to 0$, or $\psi(\theta)-\psi(0)=\theta-\E|\varepsilon|+o(1)$.

\medskip\noindent\textbf{Step 4 (Part 2: Excess magnitude in Regimes (a) and (b)).}
Identical distribution of $(s_{p'},\xi_{p'})$ across $\mathcal N(p)$ and of $\varepsilon_{p'}$ across $\mathcal S$ gives $\E[|r(p',g)|\mid\zeta_p]=G$ for $p'\in\mathcal N(p)$ and $\psi(\mu_0(g))$ for $p'\notin\mathcal N(p)$, whence
\begin{equation}
\E\bigl[\hat a(p,g)\mid\zeta_p\bigr]=G\!\!\sum_{p'\in\mathcal N(p)}\!\!c_{p'}+\psi(\mu_0(g))\!\!\sum_{p'\notin\mathcal N(p)}\!\!c_{p'}=W_{\mathrm{eff}}(p)\bigl(G-\psi(\mu_0(g))\bigr).
\end{equation}
Conditioning on $s_{p'}$ and applying Jensen's inequality to the convex function $\xi\mapsto\psi\bigl(\mu_0(g)+s_{p'}\zeta_p(\theta_{p,g}+\xi)\bigr)$ with $\E[\xi_{p'}]=0$ gives, for all $\mu_0(g)\in\mathbb R$ and $q\in[0,1]$,
\[
G\;\ge\; H(q)\triangleq q\,\psi\bigl(\mu_0(g)+\zeta_p\theta_{p,g}\bigr)+(1-q)\,\psi\bigl(\mu_0(g)-\zeta_p\theta_{p,g}\bigr).
\]
In \emph{Regime (a)}, $(2q-1)\zeta_p\mu_0(g)\ge 0$. Writing $q=\tfrac12+(q-\tfrac12)$ decomposes
\[
H(q)=\tfrac12\bigl[\psi(\mu_0(g)+\theta_{p,g})+\psi(\mu_0(g)-\theta_{p,g})\bigr]
+\bigl(q-\tfrac12\bigr)\bigl[\psi(\mu_0(g)+\zeta_p\theta_{p,g})-\psi(\mu_0(g)-\zeta_p\theta_{p,g})\bigr].
\]
Because $|\mu_0(g)+\zeta_p\theta_{p,g}|^2-|\mu_0(g)-\zeta_p\theta_{p,g}|^2=4\zeta_p\mu_0(g)\theta_{p,g}$ and $\psi(|u|)$ is strictly increasing in $|u|$, the difference in brackets shares the sign of $\zeta_p\mu_0(g)$, making the second term nonnegative whenever $(2q-1)\zeta_p\mu_0(g)\ge 0$. Strict convexity of $\psi$ at $\theta_{p,g}>0$ and $\psi(u)\ge|u|$ then give
\begin{align*}
G&\;\ge\;\tfrac12\bigl[\psi(\mu_0(g)+\theta_{p,g})+\psi(\mu_0(g)-\theta_{p,g})\bigr]\;>\;\psi(\mu_0(g)),\\
G&\;\ge\;\tfrac12\bigl(|\mu_0(g)+\theta_{p,g}|+|\mu_0(g)-\theta_{p,g}|\bigr)\;\ge\;\theta_{p,g}.
\end{align*}
In \emph{Regime (b)}, $\theta_{p,g}>2|\mu_0(g)|$. Both endpoints satisfy $|\mu_0(g)\pm\theta_{p,g}|\ge\theta_{p,g}-|\mu_0(g)|>|\mu_0(g)|$, so strict monotonicity of $\psi(|u|)$ yields
\[
G\;\ge\;H(q)\;\ge\;\psi\bigl(\theta_{p,g}-|\mu_0(g)|\bigr)\;>\;\psi(\mu_0(g))
\qquad\text{for all }q\in[0,1].
\]
\end{proof}

\paragraph{Regime sharpness and KRR threshold $\alpha_0$ (Remark~1).}
For any fixed tuple $(q,\mu_0(g),\zeta_p,\theta_{p,g})$, $H(q)>\psi(\mu_0(g))$ is the exact necessary and sufficient condition for $G>\psi(\mu_0(g))$ uniformly over admissible zero-mean $\xi_{p'}$ (attained at $\xi_{p'}\equiv 0$). When $q\in[0,1]$ is unrestricted, the threshold $\theta_{p,g}>2|\mu_0(g)|$ in Regime~(b) is sharp: for $\mu_0(g)\neq 0$ and $0<\theta_{p,g}\le 2|\mu_0(g)|$, choosing $q=1$, $\zeta_p=-\operatorname{sign}(\mu_0(g))$, and $\xi_{p'}\equiv 0$ gives
\[
G=\psi\bigl(|\mu_0(g)|-\theta_{p,g}\bigr)\le\psi(\mu_0(g))
\]
(strictly below $\psi(\mu_0(g))$ on $(0,2|\mu_0(g)|)$ and equal at $\theta_{p,g}=2|\mu_0(g)|$). Finally, for the non-proportional cross-moment claim in Remark~1 when $k_p^\top c_{a,p}>0$, substituting the resolvent identity $(K+\alpha I_n)^{-1}=\alpha^{-1}\bigl(I_n-(K+\alpha I_n)^{-1}K\bigr)$ and $\lVert(K+\alpha I_n)^{-1}K\rVert_{\mathrm{op}}=\lVert K\rVert_{\mathrm{op}}/(\lVert K\rVert_{\mathrm{op}}+\alpha)$ ($K\succeq 0$) yields
\begin{align*}
k_p^\top(K+\alpha I_n)^{-1}c_{a,p}
&\;\ge\;\frac{1}{\alpha}\!\left(k_p^\top c_{a,p}-\frac{\lVert K\rVert_{\mathrm{op}}}{\lVert K\rVert_{\mathrm{op}}+\alpha}\,\lVert k_p\rVert_2\lVert c_{a,p}\rVert_2\right)\;>\;0\\
&\quad\text{for all }\alpha>\alpha_0\triangleq\lVert K\rVert_{\mathrm{op}}\!\left(\frac{\lVert k_p\rVert_2\lVert c_{a,p}\rVert_2}{k_p^\top c_{a,p}}-1\right).
\end{align*}

\subsubsection*{Proof of Proposition~2 and Remark~2}

\begin{proof}
Throughout Part~1, contrasts are defined around the population gene mean. Write $w_\alpha(p)=M_\alpha k_p=(K+\alpha I_n)^{-1}k_p$ and $y_g=d_g+\varepsilon_g$. Replacing the population mean by the empirical support mean of Equation~\eqref{eq:pf-decomposition} shifts each contrast by the sampling error $n^{-1}\one_n^\top y_g$. Its variance is $n^{-2}\one_n^\top\operatorname{Cov}(y_g)\one_n$, which is $O(n^{-1})$ provided $\one_n^\top\operatorname{Cov}(y_g)\one_n=O(n)$ on the full support, including unannotated perturbations. This additional condition is not implied by the covariance assumptions on the annotated block alone.

\medskip\noindent\textbf{Step 1 (Part 1 \& Remark~2: Annotated block reduction and pooled amplitude calibration).}
Partition $\mathcal S=\mathcal S_+\cup\mathcal S_0$ into annotated perturbations $\mathcal S_+=\{p'\in\mathcal S:h(p')>0\}$ and unannotated perturbations $\mathcal S_0=\{p'\in\mathcal S:h(p')=0\}$. By Equation~\eqref{eq:abl-kernel}, $K_{p',:}=K_{:,p'}=0$ and $(k_p)_{p'}=0$ for all $p'\in\mathcal S_0$, so $(w_\alpha(p))_{p'}=0$ on $\mathcal S_0$ (and $w_\alpha(p)=0$ whenever $h(p)=0$). Only the $\mathcal S_+$ block therefore enters the moments, where $\E[d_g\varepsilon_g^\top]=0$ and $\E[\varepsilon_g\delta(p,g)]=0$ give $\E[y_gy_g^\top\mid k_p]_{\mathcal S_+,\mathcal S_+}=(\sigma_{\delta,g}^2K+\sigma_{\varepsilon,g}^2I_n)_{\mathcal S_+,\mathcal S_+}$ and $\E[y_g\delta(p,g)\mid k_p]_{\mathcal S_+}=\sigma_{\delta,g}^2(k_p)_{\mathcal S_+}$. Hence, for every $g\in\mathcal G$ and every query $p$,
\begin{equation}\label{eq:app-second-moment}
\mathbb{E}_{d_g, \varepsilon_g}\big[\hat{d}_\alpha(p,g)^2 \,\big|\, k_p\big] = w_\alpha(p)^\top \mathbb{E}[y_g y_g^\top\mid k_p] w_\alpha(p) = k_p^\top M_\alpha \bigl(\sigma_{\delta,g}^2 K + \sigma_{\varepsilon,g}^2 I_n\bigr) M_\alpha k_p,
\end{equation}
\begin{equation}\label{eq:app-cross-moment}
\mathbb{E}_{d_g, \varepsilon_g, \delta}\big[\hat{d}_\alpha(p,g) \delta(p,g) \,\big|\, k_p\big] = w_\alpha(p)^\top \mathbb{E}[y_g \delta(p,g)\mid k_p] = \sigma_{\delta,g}^2 k_p^\top M_\alpha k_p.
\end{equation}
Averaging Equations~\eqref{eq:app-second-moment} and~\eqref{eq:app-cross-moment} across $g\in\mathcal G$ and test perturbations $p$ by linearity of expectation (requiring no cross-gene independence), recalling $\alpha^\ast=\bar\sigma_\varepsilon^2/\bar\sigma_\delta^2$, and applying $M_\alpha(K+\alpha^\ast I_n)M_\alpha=M_\alpha+(\alpha^\ast-\alpha)M_\alpha^2$ give
\begin{align}
\sigma_{\hat\delta}^2(\alpha)&=\bar\sigma_\delta^2\Bigl(\E_p\bigl[k_p^\top M_\alpha k_p\bigr]+(\alpha^\ast-\alpha)\,\E_p\bigl[k_p^\top M_\alpha^{2}k_p\bigr]\Bigr),\\
\overline{\hat\delta\delta}(\alpha)&=\bar\sigma_\delta^2\,\E_p\bigl[k_p^\top M_\alpha k_p\bigr].
\end{align}
For the population-pooled quantities $\eta=\sigma_{\hat\delta}/\bar\sigma_\delta$ and $\rho=\overline{\hat\delta\delta}/(\sigma_{\hat\delta}\bar\sigma_\delta)$ of Section~\ref{sec:response-learning}, $\eta/\rho=\sigma_{\hat\delta}^2/\overline{\hat\delta\delta}$ yields
\begin{equation}
\frac{\eta(\alpha)}{\rho(\alpha)}=1+(\alpha^\ast-\alpha)\,\frac{\E_p\bigl[k_p^\top M_\alpha^{2}k_p\bigr]}{\E_p\bigl[k_p^\top M_\alpha k_p\bigr]} .
\end{equation}
Because $M_\alpha\succ0$, both expectations are strictly positive whenever $\E_p[k_p^\top M_\alpha k_p]>0$, giving $\operatorname{sign}(\eta-\rho)=\operatorname{sign}(\alpha^\ast-\alpha)$ and $\eta(\alpha^\ast)^2=\rho(\alpha^\ast)^2=\E_p[k_p^\top M_{\alpha^\ast}k_p]$. For arbitrary conditional moments $(\overline{\Sigma}_y,\bar c_p)$ in Remark~2, the same identity $\eta/\rho=\sigma_{\hat\delta}^2/\overline{\hat\delta\delta}$ gives $\eta(\alpha)/\rho(\alpha)=\E_p[k_p^\top M_\alpha\overline{\Sigma}_yM_\alpha k_p]/\E_p[k_p^\top M_\alpha\bar c_p]$, and substituting $\alpha M_\alpha=I_n-M_\alpha K\to I_n$ as $\alpha\to\infty$ yields the $\alpha^{-1}$ asymptotic whenever $\E_p[k_p^\top\bar c_p]>0$.

\medskip\noindent\textbf{Step 2 (Part 2: Sherman--Morrison decomposition under background similarity).}
Write $\tilde K+\alpha I_n=A_\alpha+b\one_n\one_n^\top$ with $A_\alpha=(1-b)K+\alpha I_n\succ0$, and set $v_\alpha=A_\alpha^{-1}\one_n$, $u=\one_n^\top A_\alpha^{-1}y=v_\alpha^\top y$, $s=\one_n^\top v_\alpha$, and $t=k_p^\top v_\alpha$. By the Sherman--Morrison formula,
\begin{equation}
(\tilde K+\alpha I_n)^{-1}y=A_\alpha^{-1}y-\frac{bu}{1+bs}\,A_\alpha^{-1}\one_n .
\end{equation}
Multiplying on the left by $\tilde k_p^\top=(1-b)k_p^\top+b\one_n^\top$ and using $(1-b)A_\alpha^{-1}=\bigl(K+\tfrac{\alpha}{1-b}I_n\bigr)^{-1}$ give
\begin{equation}
\begin{split}
\tilde k_p^\top(\tilde K+\alpha I_n)^{-1}y
&=(1-b)k_p^\top A_\alpha^{-1}y+bu-\frac{bu\bigl((1-b)t+bs\bigr)}{1+bs}\\
&=k_p^\top\Bigl(K+\tfrac{\alpha}{1-b}I_n\Bigr)^{-1}y+\frac{b\bigl(1-(1-b)t\bigr)}{1+bs}\,u .
\end{split}
\end{equation}

\medskip\noindent\textbf{Step 3 (Part 2: Degree-dispersion and joint-PSD bounds on centered contrasts).}
When $\one_n^\top y=0$, $u=v_\alpha^\top y=v_{\alpha,\perp}^\top y$ with $v_{\alpha,\perp}=(I_n-n^{-1}\one_n\one_n^\top)v_\alpha$. Setting $c=(\alpha+(1-b)\bar\kappa)^{-1}$ for $\bar\kappa=n^{-1}\one_n^\top K\one_n\ge 0$ gives
\[
A_\alpha(v_\alpha-c\one_n)=\one_n-c\bigl((1-b)K\one_n+\alpha\one_n\bigr)=-c(1-b)(K\one_n-\bar\kappa\one_n).
\]
Since $\lVert A_\alpha^{-1}\rVert_{\mathrm{op}}\le\alpha^{-1}$ and orthogonal projection onto $\one_n^\perp$ is non-expansive,
\[
\lVert v_{\alpha,\perp}\rVert_2\;\le\;\lVert v_\alpha-c\one_n\rVert_2\;\le\;\frac{1-b}{\alpha\bigl(\alpha+(1-b)\bar\kappa\bigr)}\,\lVert K\one_n-\bar\kappa\one_n\rVert_2 ,
\]
which vanishes identically whenever $K\one_n=\kappa_{\one}\one_n$ for the constant-direction eigenvalue $\kappa_{\one}=\bar\kappa$ of $K$ (distinct from the within-source cosines $\kappa_j$). Moreover, when the joint kernel on $\mathcal S\cup\{p\}$ is positive semidefinite with $k(p,p)\le 1$, diagonalizing $K=\sum_{i=1}^n\lambda_iu_iu_i^\top$ and expanding $k_p=\sum_{i=1}^n\sqrt{\lambda_i}\gamma_iu_i$ ($\sum_i\gamma_i^2\le k(p,p)\le 1$) and $\one_n=\sum_{i=1}^n\beta_iu_i$ yield
\begin{align*}
\bigl|(1-b)t\bigr|
&=\Bigl|\sum_{i=1}^n\frac{(1-b)\sqrt{\lambda_i}\,\gamma_i\beta_i}{(1-b)\lambda_i+\alpha}\Bigr|
\;\le\;\Bigl(\sum_{i=1}^n\gamma_i^2\,\frac{(1-b)^2\lambda_i}{(1-b)\lambda_i+\alpha}\Bigr)^{1/2}s^{1/2}
\;\le\;\sqrt{(1-b)s},\\
\frac{b\,\bigl|1-(1-b)t\bigr|}{1+bs}
&\le\;\frac{b\bigl(1+\sqrt{(1-b)s}\bigr)}{1+bs}
\;\le\;\sqrt{\frac{b}{1+bs}},
\end{align*}
where the final inequality follows from $b(1+bs)-b^2\bigl(1+\sqrt{(1-b)s}\bigr)^2=b\bigl(\sqrt{1-b}-b\sqrt{s}\bigr)^2\ge 0$.

\medskip\noindent\textbf{Step 4 (Part 2: Signal dilution of Nadaraya--Watson averaging).}
For centered $y$ ($\one_n^\top y=0$), $k_p=0$ gives $\hat d_{\mathrm{NW}}(p,g)=0$ by definition; when $k_p^\top\one_n>0$, $\tilde k_p^\top y=(1-b)k_p^\top y$ and $\tilde k_p^\top\one_n=(1-b)k_p^\top\one_n+nb>0$, yielding
\[
\hat d_{\mathrm{NW}}(p,g)=\frac{(1-b)k_p^\top y}{(1-b)k_p^\top\one_n+nb}=(1-\lambda_p)\,\frac{k_p^\top y}{k_p^\top\one_n},
\qquad
\lambda_p=\frac{nb}{(1-b)k_p^\top\one_n+nb}\in(0,1).
\]
\end{proof}
% !TEX root = ../main.tex
\section{Experimental Protocols, Per-Cell-Line Results, and Supplementary Ablations}\label{app:main-cell-results}\label{app:genetak-results}
This appendix details the baseline scoring, thresholding, and bootstrap uncertainty protocols (\appref{app:baseline-data}), reports the paired bootstrap differences and per-cell-line breakdowns for PerturbQA, held-out RPE1, and GeneTAK (\appref{app:rpe1-transfer-results}), and provides the supplementary knowledge-source and transfer-operator ablation tables (\appref{app:ablations}) accompanying Section~\ref{sec:experiments}.

\subsection{Baseline Scoring, Thresholding, and Uncertainty Protocols}\label{app:baseline-data}\label{app:main-protocol}\label{app:baseline-details}\label{app:main-uncertainty}
\paragraph{PerturbQA baseline scoring and threshold calibration.}
Table~\ref{tab:baseline-reproduction} summarizes the reproduction strategies and implementation settings for the seven PerturbQA baselines. For each target gene, cell line, and task, the empirical gene prior computes the positive training fraction $\pi(g)$ and predicts the positive class when $\pi(g)>1/2$. Exact ties ($\pi(g)=1/2$, occurring in 235 of 116,041 DE test queries) and unanswered baseline queries fall back to the majority label of the pooled cell-line training rate $\bar\pi$ (for comparison, predicting non-DE on every query yields an equal-cell mean DE Macro-F1 of 0.4626). For the continuous predictors GEARS, STATE, PerturbDiff, and PertOmni, DE is scored by absolute predicted response and DIR by signed predicted response, with binary decision thresholds calibrated on held-out training perturbations (a 10\% validation split for GEARS, STATE, and PerturbDiff, and pooled four-fold out-of-fold development predictions for PertOmni) to maximize Macro-F1. Specifically, for GEARS and PerturbDiff, subtracting the control profile from predicted $\log(1+\mathrm{CP10K})$ profiles yields the predicted responses used both for binary task scoring and for the continuous profile diagnostics in Table~\ref{tab:profile-comparison}, with the 10\% validation split used for both epoch selection and threshold calibration; PertOmni predicts response profiles directly via neighbor averaging over training responses. For STATE, raw scores are mapped to empirical quantiles of the validation-score distribution, and query pairs outside STATE's decoder vocabulary or lacking a perturbation embedding are treated as unanswered and receive the fallback $\bar\pi$.

\begin{table}[H]
\centering\small
\caption{\textbf{Reproduction strategies for the main PerturbQA baselines.}}
\label{tab:baseline-reproduction}
\setlength{\tabcolsep}{4pt}
\renewcommand{\arraystretch}{1.12}
\begin{tabularx}{\linewidth}{@{}>{\raggedright\arraybackslash}p{0.13\linewidth} >{\raggedright\arraybackslash}p{0.20\linewidth} >{\raggedright\arraybackslash}X@{}}
\toprule
Method & Strategy & Implementation and settings \\
\midrule
GEARS & Retrain the authors' package &
\texttt{cell-gears} 0.1.2 with default hyperparameters (20 epochs, hidden size 64, batch size 32, learning rate $10^{-3}$). Official training perturbations for each cell line, without highly-variable-gene pre-filtering. K562 uses the genome-wide screen because the essential-gene screen covers only about half of its training perturbations. \\
\addlinespace[4pt]
STATE & Train the transition model and decoder from scratch &
PerturbQA-aligned split with validation seed 42 and a negative-binomial gene decoder. Inputs are GenePT model-3 text embeddings for perturbations and cell embeddings from the frozen, pretrained SE-600M State Embedding model. \\
\addlinespace[4pt]
PerturbDiff & Train the latent diffusion model from scratch &
Cell-line-isolated training on each PerturbQA training split (10\% perturbation-level validation split for validation-MSE early stopping), using official GenePT perturbation embeddings and blind test sampling (zero test-expression access). The released pretrained checkpoint is excluded because its pretraining corpus covers 94.7\% (1,197/1,264) of PerturbQA test perturbations. \\
\addlinespace[4pt]
PertOmni & Train the official architecture from scratch &
Same-cell training with GenePT model-3 gene embeddings and official MSigDB \texttt{c5.go.bp.v2025.1.Hs} GO:BP pathway embeddings encoded with Qwen3-Embedding-4B, averaged over five seeds using the authors' bidirectional accuracy early stopping and development-MSE selection of $K$ for \texttt{kNN-text} transfer. Alternative pathway sources and multi-cell joint training perform similarly. \\
\addlinespace[4pt]
AROMA & Use the public checkpoint without additional training &
The authors' checkpoint \citep{wang2026aroma} was trained on all four cell lines (\appref{app:historical-exposure}). A single parsed ternary answer is projected to DE or DIR; a non-DE answer to a DIR query provides no direction. \\
\addlinespace[4pt]
SUMMER & Run the public code with a different backbone &
Gemma4-26B-A4B-it \citep{gemma2026gemma4} with the three-answer aggregate. The labeled retrieval index uses the queried cell line's official training split for the queried task. \\
\addlinespace[4pt]
CORE-Voting & Use published results &
The official implementation and ReasonKG resource were unavailable to us. We use the reported DE results \citep{yuan2026core} and reconstruct Macro-F1 from accuracy and positive-class F1. \\
\bottomrule
\end{tabularx}
\end{table}

\paragraph{GeneTAK baseline protocols.}
To evaluate the LLM reasoning baseline VCWorld \citep{wei2026vcworld} on the 95 held-out compounds across all five GeneTAK cell lines, we run the authors' public repository with Gemma4-26B-A4B-it in a single pass without repeats. We retain the released prompt templates, compound-similarity and gene-neighbor graphs, and entity descriptions, while replacing randomly sampled prompt demonstration outcomes with their ground-truth training labels. Unanswered queries receive the pooled training-rate fallback $\bar\pi$. Likewise, for the within-cell-line gene prior on GeneTAK, the 1,078 DIR test queries whose target gene has no positive DE training pair in that cell line fall back to the majority label of $\bar\pi$. For STATE on GeneTAK, we retrain the model from scratch on the 284 training drugs within each cell line using the released Tahoe ST-HVG configuration and Morgan fingerprint representations (excluding the released Tahoe checkpoint, whose training vocabulary includes the held-out test compounds). Following the benchmark label construction, we pool doses per compound, select checkpoints by validation loss, and calibrate binary thresholds by development-drug Macro-F1. For PertOmni on GeneTAK, we train the official architecture from scratch per cell line on the 284 training drugs (five seeds, official bidirectional accuracy early stopping) using Morgan-fingerprint drug embeddings and LLM-encoded cell-line context embeddings, selecting $K$ for \texttt{kNN-text} transfer by development MSE and calibrating DE and DIR thresholds by Macro-F1 on pooled out-of-fold development predictions; multi-cell joint training across all five cell lines performs similarly (mean DE Macro-F1 .8661, \esw{} .5011, \pga{} .5352; DIR .7789, .5037, .5400). Because GEARS and the released genetic PerturbDiff architecture take gene-level representations rather than compound descriptors, neither is evaluated on GeneTAK.

\paragraph{Paired bootstrap confidence intervals.}
All 95\% confidence intervals (2.5th to 97.5th percentiles) are computed from 2,000 paired bootstrap replicates \citep{efron1994bootstrap} with random draws shared across compared methods, conditional on fixed test predictions. In the main PerturbQA comparison, each replicate resamples mixed-label target genes within each cell line (carrying each gene's metric value and class-support weight $w_g$) for within-gene metrics (\pga{}, pgBAcc, and \esw{}). For global metrics (accuracy and Macro-F1), each replicate resamples query perturbations (carrying all query rows of a perturbation together), averaging equally across the four cell lines before computing paired differences. In the SUMMER prompt-intervention experiments (Section~\ref{sec:retrieval-transfer}), replicates resample query perturbations within each cell line, carrying all query rows together and recomputing gene eligibility and support weights within each draw. On GeneTAK, each replicate (seed 17) resamples the 95 held-out compounds with weights shared across all five cell lines, recomputes within-gene eligibility and support weights, and averages across cell lines equally.

\subsection{Paired Differences, Per-Cell-Line Results, and Held-Out RPE1 Breakdown}\label{app:rpe1-transfer-results}\label{app:generalization}
Table~\ref{tab:main-deltas} reports the paired differences between RELAY and each PerturbQA baseline along with 95\% bootstrap confidence intervals, while Tables~\ref{tab:main-cells} and~\ref{tab:genetak-cells} provide the complete per-cell-line breakdowns for PerturbQA (Section~\ref{sec:main-response-results}) and GeneTAK (Section~\ref{sec:main-drug}), respectively.

\begin{table}[H]
\centering\small
\caption{\textbf{Paired differences, RELAY minus each baseline (percentage points).} Means across cell lines, weighted equally, on the PerturbQA test set, with 95\% bootstrap intervals from the protocol of \appref{app:main-uncertainty}. CORE-Voting lacks query-level predictions and is omitted.}
\label{tab:main-deltas}
\setlength{\tabcolsep}{4pt}
\begin{tabular}{@{}llrrr@{}}
\toprule
Task & Baseline & $\Delta$Macro-F1 & $\Delta$\esw{} & $\Delta$pgAUROC \\
\midrule
DE & Gene prior & 1.27 [0.82, 1.73] & 15.74 [15.01, 16.49] & 27.32 [26.33, 28.31] \\
 & GEARS & 17.38 [16.73, 18.10] & 14.55 [13.64, 15.48] & 25.88 [24.35, 27.35] \\
 & STATE & 21.46 [20.63, 22.36] & 10.16 [9.29, 11.01] & 22.80 [21.35, 24.27] \\
 & PerturbDiff & 15.70 [14.99, 16.48] & 11.41 [10.48, 12.36] & 23.96 [22.49, 25.34] \\
 & PertOmni & 9.20 [8.63, 9.80] & 4.65 [3.86, 5.46] & 12.52 [11.29, 13.81] \\
 & AROMA & 3.23 [2.73, 3.73] & 14.05 [13.25, 14.87] & 25.80 [24.76, 26.86] \\
 & SUMMER & 13.80 [13.32, 14.26] & 7.48 [6.58, 8.39] & 18.41 [17.18, 19.64] \\
\midrule
DIR & Gene prior & 7.49 [6.16, 8.88] & 26.72 [25.33, 27.85] & 36.11 [34.38, 37.77] \\
 & GEARS & 16.37 [14.89, 17.78] & 25.63 [23.92, 27.11] & 31.49 [28.62, 34.42] \\
 & STATE & 13.08 [11.48, 14.74] & 15.00 [13.44, 16.46] & 20.27 [17.81, 22.82] \\
 & PerturbDiff & 9.30 [7.81, 10.79] & 12.27 [10.82, 13.72] & 18.23 [15.66, 20.91] \\
 & PertOmni & 2.81 [1.91, 3.78] & 4.41 [3.30, 5.46] & 5.91 [4.02, 7.89] \\
 & AROMA & 14.57 [13.06, 16.17] & 16.43 [14.76, 18.14] & 30.85 [28.85, 32.97] \\
 & SUMMER & 5.07 [4.01, 6.11] & 6.53 [5.28, 7.85] & 19.74 [17.37, 22.05] \\
\bottomrule
\end{tabular}
\end{table}

\begin{table}[H]
\centering\small
\caption{\textbf{Main-experiment results by cell line.} All entries include the common fallback, and DIR uses the official DE-positive query population. Unanswered LLM queries receive the pooled training-rate score $\bar\pi\in(0,1)$ and its majority-class decision (\appref{app:baseline-data}), introducing an intermediate score level that can slightly separate \pga{} from pgBAcc on single-answer outputs (e.g., AROMA on DE). Mean rows weight the four cell lines equally and are computed from unrounded values.}
\label{tab:main-cells}
\setlength{\tabcolsep}{3pt}
\begin{tabular}{@{}llrrrrrrrr@{}}
\toprule
 & & \multicolumn{4}{c}{DE} & \multicolumn{4}{c}{DIR}\\
\cmidrule(lr){3-6}\cmidrule(l){7-10}
Cell line & Method & Macro-F1 & \eswhead{} & pgBAcc & pgAUROC & Macro-F1 & \eswhead{} & pgBAcc & pgAUROC \\
\midrule
K562 & Gene prior & .8458 & .5000 & .5000 & .5000 & .7299 & .5000 & .5000 & .5000 \\
 & GEARS & .6295 & .5001 & .4990 & .5008 & .6323 & .5293 & .5359 & .5160 \\
 & STATE & .6942 & .6053 & .5904 & .5859 & .7549 & .7082 & .6739 & .7365 \\
 & PerturbDiff & .6881 & .5487 & .5384 & .5488 & .7353 & .6788 & .6425 & .7340 \\
 & PertOmni & .7711 & .6297 & .6237 & .6788 & .8301 & .7793 & .7388 & .8836 \\
 & AROMA & .8307 & .5173 & .5169 & .5165 & .6665 & .5948 & .5628 & .5530 \\
 & SUMMER & .7394 & .6004 & .5886 & .6086 & .7952 & .7591 & .7021 & .7342 \\
 & RELAY & .8890 & .6937 & .6855 & .8086 & .8842 & .8498 & .8062 & .9375 \\
\midrule
HepG2 & Gene prior & .8831 & .5000 & .5000 & .5000 & .8245 & .5000 & .5000 & .5000 \\
 & GEARS & .7397 & .5392 & .5308 & .5260 & .7485 & .5188 & .5145 & .5807 \\
 & STATE & .6880 & .5753 & .5584 & .5605 & .7667 & .6683 & .6316 & .7076 \\
 & PerturbDiff & .7497 & .5643 & .5462 & .5414 & .8128 & .6722 & .6052 & .6893 \\
 & PertOmni & .8152 & .6205 & .6103 & .6598 & .8636 & .7322 & .6914 & .7888 \\
 & AROMA & .8652 & .5186 & .5155 & .5159 & .7545 & .6180 & .5747 & .5641 \\
 & SUMMER & .7679 & .5893 & .5872 & .5996 & .8408 & .6853 & .6207 & .6532 \\
 & RELAY & .8911 & .6615 & .6474 & .7858 & .8886 & .7671 & .7029 & .8518 \\
\midrule
Jurkat & Gene prior & .8915 & .5000 & .5000 & .5000 & .7753 & .5000 & .5000 & .5000 \\
 & GEARS & .7161 & .5026 & .5025 & .5056 & .6528 & .5020 & .4898 & .5195 \\
 & STATE & .6254 & .4908 & .4864 & .4872 & .6572 & .5125 & .4782 & .5078 \\
 & PerturbDiff & .7443 & .5220 & .5152 & .5239 & .7411 & .6547 & .6108 & .6707 \\
 & PertOmni & .7979 & .5858 & .5724 & .6030 & .8185 & .7277 & .6565 & .7723 \\
 & AROMA & .8809 & .5160 & .5147 & .5155 & .7265 & .6305 & .5572 & .5568 \\
 & SUMMER & .7616 & .5625 & .5522 & .5661 & .8031 & .7114 & .6381 & .6547 \\
 & RELAY & .8948 & .6262 & .6215 & .7478 & .8396 & .7592 & .6924 & .8480 \\
\midrule
RPE1 & Gene prior & .9050 & .5000 & .5000 & .5000 & .8595 & .5000 & .5000 & .5000 \\
 & GEARS & .7956 & .5059 & .5042 & .5249 & .8002 & .4934 & .4899 & .5685 \\
 & STATE & .7104 & .5520 & .5445 & .5471 & .7869 & .5795 & .5736 & .6819 \\
 & PerturbDiff & .7661 & .5382 & .5257 & .5203 & .8276 & .5725 & .5687 & .6214 \\
 & PertOmni & .8242 & .6075 & .5941 & .6502 & .8643 & .6530 & .6378 & .7633 \\
 & AROMA & .8702 & .5157 & .5129 & .5129 & .7585 & .5682 & .5417 & .5367 \\
 & SUMMER & .7553 & .5784 & .5681 & .5821 & .8470 & .6519 & .5939 & .6126 \\
 & RELAY & .9014 & .6482 & .6354 & .7505 & .8765 & .6926 & .6620 & .8071 \\
\midrule
Mean & Gene prior & .8814 & .5000 & .5000 & .5000 & .7973 & .5000 & .5000 & .5000 \\
 & GEARS & .7202 & .5119 & .5091 & .5143 & .7085 & .5109 & .5075 & .5462 \\
 & STATE & .6795 & .5559 & .5449 & .5452 & .7414 & .6171 & .5893 & .6584 \\
 & PerturbDiff & .7370 & .5433 & .5314 & .5336 & .7792 & .6445 & .6068 & .6788 \\
 & PertOmni & .8021 & .6109 & .6001 & .6479 & .8441 & .7231 & .6811 & .8020 \\
 & AROMA & .8618 & .5169 & .5150 & .5152 & .7265 & .6029 & .5591 & .5526 \\
 & SUMMER & .7560 & .5827 & .5740 & .5891 & .8215 & .7019 & .6387 & .6637 \\
 & RELAY & .8941 & .6574 & .6475 & .7732 & .8722 & .7672 & .7159 & .8611 \\
\bottomrule
\end{tabular}
\end{table}

In the held-out RPE1 transfer evaluation (Section~\ref{sec:main-ood}), a test query is classified as overlapping if its perturbation--gene pair appears in the official DE training split of at least one source cell line (K562, HepG2, or Jurkat). Excluding these 1,511 overlapping pairs from the 37,942 RPE1 test queries leaves 36,431 non-overlapping DE queries, including 4,286 DIR queries (Table~\ref{tab:rpe1-transfer-results}). To isolate the effect of exact pair overlap, we compare RELAY against both the \emph{source gene prior} (which assigns each gene its majority training label pooled across the three source cell lines) and an \emph{exact-pair prior}. The exact-pair prior copies the majority training label of the matching perturbation--gene pair from the source cell lines when available and falls back to the source gene prior otherwise. On the 1,511 overlapping queries, the exact-pair prior attains a DE Macro-F1 of $0.9402$ compared with $0.9006$ for RELAY, whereas on the 36,431 non-overlapping queries it reduces identically to the source gene prior. On the three-class task of Table~\ref{tab:ood}, the exact-pair prior reaches a Macro-F1 of $.7246$ overall and $.6552$ on non-overlapping queries, whereas RELAY achieves $.7597$ overall and $.7176$ on the non-overlapping subset.

\begin{table}[H]
\centering\small
\caption{\textbf{Binary tasks after transfer to held-out RPE1.} RELAY's DE and DIR results under the main metric definitions, where non-overlap excludes the 1,511 of 37,942 queries whose pair appears in training data from source cell lines. Table~\ref{tab:ood} gives the three-class results, with AROMA values as published for its model trained on the three source cell lines \citep[Table~2]{wang2026aroma}, not the four-cell-line checkpoint of Table~\ref{tab:baseline-reproduction}.}
\label{tab:rpe1-transfer-results}\label{tab:rpe1-readout-details}
\setlength{\tabcolsep}{4pt}
\begin{tabular}{@{}lrrrrr@{}}
\toprule
Task / population & Macro-F1 & Accuracy & pgBAcc & \esw{} & pgAUROC \\
\midrule
DE / all & .8459 & .9254 & .6018 & .5933 & .7091 \\
DIR / all & .8507 & .8557 & .7114 & .7333 & .8125 \\
DE / non-overlap & .8191 & .9251 & .5908 & .5807 & .6852 \\
DIR / non-overlap & .8264 & .8325 & .6702 & .6576 & .7686 \\
\bottomrule
\end{tabular}
\end{table}

\begin{table}[H]
\centering\small
\caption{\textbf{GeneTAK results by cell line on the 95 held-out drugs.} VCWorld is evaluated using the authors' public repository (\appref{app:baseline-data}). Mean rows weight cell lines equally.}
\label{tab:genetak-cells}\label{tab:genetak-de-cells}\label{tab:genetak-dir-cells}
\setlength{\tabcolsep}{3pt}
\renewcommand{\arraystretch}{0.95}
\begin{tabular}{@{}llrrrrrrrr@{}}
\toprule
 & & \multicolumn{4}{c}{DE} & \multicolumn{4}{c}{DIR}\\
\cmidrule(lr){3-6}\cmidrule(l){7-10}
Cell line & Method & Macro-F1 & \eswhead{} & pgBAcc & pgAUROC & Macro-F1 & \eswhead{} & pgBAcc & pgAUROC \\
\midrule
C32 & Gene prior & .9240 & .5000 & .5000 & .5000 & .8062 & .5000 & .5000 & .5000 \\
 & STATE & .8295 & .5027 & .5065 & .5160 & .6443 & .5340 & .5341 & .5530 \\
 & PertOmni & .8728 & .5036 & .5014 & .5354 & .8003 & .5083 & .5068 & .5461 \\
 & VCWorld & .7094 & .5236 & .5253 & .5253 & .6008 & .5136 & .5154 & .5157 \\
 & RELAY & .9239 & .5481 & .5557 & .6210 & .7990 & .5640 & .5667 & .6401 \\
\midrule
HepG2/C3A & Gene prior & .9386 & .5000 & .5000 & .5000 & .7855 & .5000 & .5000 & .5000 \\
 & STATE & .8556 & .5115 & .5187 & .5208 & .6605 & .5265 & .5066 & .5375 \\
 & PertOmni & .8866 & .4983 & .5009 & .5252 & .7587 & .5008 & .4998 & .5075 \\
 & VCWorld & .7135 & .5324 & .5379 & .5379 & .6068 & .5324 & .5239 & .5235 \\
 & RELAY & .9365 & .5395 & .5568 & .6005 & .7586 & .5445 & .5497 & .6062 \\
\midrule
HOP62 & Gene prior & .9104 & .5000 & .5000 & .5000 & .7907 & .5000 & .5000 & .5000 \\
 & STATE & .8270 & .5079 & .5047 & .5137 & .6755 & .5391 & .5421 & .5671 \\
 & PertOmni & .8529 & .5033 & .5020 & .5568 & .7912 & .5099 & .5083 & .5619 \\
 & VCWorld & .7015 & .5260 & .5170 & .5170 & .5901 & .5075 & .5045 & .5044 \\
 & RELAY & .9048 & .5376 & .5390 & .5848 & .7979 & .5654 & .5754 & .6399 \\
\midrule
Hs~766T & Gene prior & .9192 & .5000 & .5000 & .5000 & .7597 & .5000 & .5000 & .5000 \\
 & STATE & .8228 & .5229 & .5263 & .5558 & .6228 & .5422 & .5415 & .5765 \\
 & PertOmni & .8657 & .4979 & .4989 & .5322 & .7490 & .5026 & .5017 & .5167 \\
 & VCWorld & .6874 & .5340 & .5426 & .5426 & .5879 & .5261 & .5257 & .5255 \\
 & RELAY & .9203 & .5452 & .5596 & .6116 & .7629 & .5664 & .5845 & .6604 \\
\midrule
PANC-1 & Gene prior & .9058 & .5000 & .5000 & .5000 & .8051 & .5000 & .5000 & .5000 \\
 & STATE & .8142 & .5113 & .5070 & .5207 & .6346 & .5267 & .5179 & .5466 \\
 & PertOmni & .8577 & .5034 & .5040 & .5615 & .7959 & .5028 & .5033 & .5353 \\
 & VCWorld & .6824 & .5250 & .5258 & .5258 & .5891 & .5100 & .5012 & .5009 \\
 & RELAY & .9038 & .5613 & .5745 & .6513 & .7955 & .5469 & .5419 & .5954 \\
\midrule
Mean & Gene prior & .9196 & .5000 & .5000 & .5000 & .7894 & .5000 & .5000 & .5000 \\
 & STATE & .8298 & .5113 & .5127 & .5254 & .6475 & .5337 & .5284 & .5561 \\
 & PertOmni & .8671 & .5013 & .5014 & .5422 & .7790 & .5049 & .5040 & .5335 \\
 & VCWorld & .6988 & .5282 & .5297 & .5297 & .5949 & .5179 & .5142 & .5140 \\
 & RELAY & .9179 & .5463 & .5571 & .6138 & .7828 & .5574 & .5637 & .6284 \\
\bottomrule
\end{tabular}
\end{table}

\subsection{Supplementary Source and Operator Ablations}\label{app:ablations}\label{app:effect-tradeoff}\label{app:abl-what}\label{app:abl-encoding}\label{app:representation}\label{app:abl-operator}\label{app:operator-details}\label{app:abl-components}\label{app:abl-tradeoffs}

\paragraph{Direct-score transfer variants.}
In Block~(a) of Table~\ref{tab:buildup-ablation} (Section~\ref{sec:main-hypothesis}), all variants transfer target quantities through the knowledge kernel via kernel ridge regression and evaluate direct task scores without a learned readout. Ridge penalties are selected on pooled development predictions by \esw{}, and binary decision thresholds are calibrated on development folds. For each target gene $g$, let $L_g\subseteq\mathcal P_{\mathrm{tr}}$ denote the subset of training perturbations with a benchmark label for the queried task. \emph{Labels ($L_g$)} transfers discrete binary labels $y$ over $L_g$ centered around the gene's label mean, while \emph{Contrasts ($L_g$)} and \emph{Contrasts $|\widehat d+\mu|$} transfer signed contrasts $d=r-\mu$ restricted to $L_g$ or across all training perturbations $\mathcal P_{\mathrm{tr}}$ and score DE by $|\widehat d+\mu|$. \emph{Magnitude $\widehat a+\nu$} instead scores DE directly by the transferred magnitude view $\widehat a+\nu$. Finally, \emph{Binarized responses} transfers $\mathrm{sign}(r)$ for DIR and the indicator $\mathbb{I}[|r|>t]$ for DE across all training perturbations $\mathcal P_{\mathrm{tr}}$. Here $t\in[0.10,0.14]$ is selected within each support fold to maximize balanced accuracy against training DE labels, achieving 95.5--97.2\% balanced accuracy on training pairs while $\mathrm{sign}(r)$ matches 100\% of training DIR labels. At equal support ($L_g$), discrete labels slightly outperform raw contrasts on DIR ($+1.85$ \esw{} points), whereas expanding support from $L_g$ to $\mathcal P_{\mathrm{tr}}$ accounts for the primary within-gene gain.

\paragraph{Knowledge-source ablations.}
Table~\ref{tab:source-ablation} reports development-fold contrast prediction MSE when individual curated databases are removed from or used in isolation within the knowledge representation $\phi(p)$ (Section~\ref{sec:method-encoding}). Every source contributes positively to contrast recovery in both the signed and magnitude views. STRING provides the largest single-source reduction in MSE over GenePT embeddings, and combining all five databases achieves the lowest error overall. Learning per-source scalar weights alters development MSE by less than 0.7\% and fluctuates across folds, motivating our unweighted source-normalized kernel (Equation~\eqref{eq:abl-kernel}).

\begin{table}[H]
\vspace{-4pt}
\centering\small
\caption{\textbf{Contrast prediction error by encoding.} Development folds only, reporting mean squared error over all genes and query perturbations ($\times10^{3}$), averaged equally over folds and then over cell lines. All encodings share the support partitions, the frozen penalties ($\alpha=1$, or $0.1$ for RPE1), and the no-intercept solver. $\Delta$ is the relative change from the full knowledge representation.}
\label{tab:source-ablation}\label{tab:representation-mse}
\setlength{\tabcolsep}{5pt}
\renewcommand{\arraystretch}{0.88}
\begin{tabular}{@{}lrrrr@{}}
\toprule
 & \multicolumn{2}{c}{Signed contrast} & \multicolumn{2}{c}{Magnitude contrast}\\
\cmidrule(lr){2-3}\cmidrule(l){4-5}
Encoding & MSE $\downarrow$ & $\Delta$ & MSE $\downarrow$ & $\Delta$\\
\midrule
Knowledge representation (RELAY) & \textbf{4.886} & --- & \textbf{2.924} & ---\\
\midrule
Without CORUM   & 4.894 & $+0.2\%$ & 2.927 & $+0.1\%$\\
Without Reactome & 4.912 & $+0.5\%$ & 2.933 & $+0.3\%$\\
Without GO      & 4.938 & $+1.1\%$ & 2.960 & $+1.2\%$\\
Without BioPlex & 4.948 & $+1.3\%$ & 2.964 & $+1.4\%$\\
Without STRING  & 4.999 & $+2.3\%$ & 2.996 & $+2.4\%$\\
\midrule
STRING only   & 5.092 & $+4.2\%$  & 3.030 & $+3.6\%$\\
Reactome only & 5.304 & $+8.6\%$  & 3.171 & $+8.4\%$\\
BioPlex only  & 5.443 & $+11.4\%$ & 3.224 & $+10.3\%$\\
CORUM only    & 5.474 & $+12.0\%$ & 3.265 & $+11.6\%$\\
GO only       & 5.488 & $+12.3\%$ & 3.280 & $+12.2\%$\\
\midrule
GenePT substitution & 5.600 & $+14.6\%$ & 3.289 & $+12.5\%$\\
\bottomrule
\end{tabular}
\vspace{-6pt}
\end{table}

\paragraph{Response-transfer operator and representation protocols.}
In Block~(c) of Table~\ref{tab:buildup-ablation} (Section~\ref{sec:main-hypothesis}), \emph{GenePT in RELAY} replaces only $\phi(p)$ with $L_2$-normalized GenePT model-3 text embeddings \citep{chen2025genept} under the frozen penalties, whereas \emph{PertOmni kNN in RELAY} substitutes PertOmni's learned \texttt{kNN-text} neighbor-averaged signed and magnitude responses (out-of-fold on development folds and on test) for RELAY's ridge regressions while keeping all downstream readouts fixed. In Block~(d) of Table~\ref{tab:buildup-ablation}, we compare kernel ridge regression against top-$k$ neighbor-averaging operators (reporting equal-cell point estimates). In the first three rows, we hold the knowledge kernel, fold partitions, $(F,\mathrm{SE})$ statistics, and downstream readout retraining fixed, and replace only the ridge regressions for $D$ and $A$ with equal-weight or kernel-weighted averages over the top-$k$ training perturbations with positive similarity $k(p,p')>0$ (assigning zero contrast when no positive neighbor exists). For each cell line, $k$ is selected on pooled development predictions to maximize the mean of DE and DIR Macro-F1, yielding $k=5,5,5,10$ (equal-weight) and $k=10,20,20,20$ (kernel-weighted) on K562, HepG2, Jurkat, and RPE1, respectively. The fourth row (\emph{Public graph-neighbor average}) averages measured training responses over each query's top 20 STRING and GO neighbors \citep{fawkes2026kg}, without a learned readout; thresholds are development-calibrated, with training gene means as the no-neighbor fallback.

\end{document}